\documentclass[10pt,twocolumn,letterpaper]{article}

\usepackage[pagenumbers]{wacv} 

\usepackage{placeins}
\usepackage{algorithm}
\usepackage{algpseudocode}
\usepackage{multirow}

\newcommand{\AN}[1]{{\color{black}#1}}
\newcommand{\ANedit}[1]{{\color{black}#1}}
\newcommand{\TK}[1]{{\color{black}#1}}

\definecolor{wacvblue}{rgb}{0.21,0.49,0.74}
\usepackage[pagebackref,breaklinks,colorlinks,allcolors=wacvblue]{hyperref}

\def\wacvPaperID{465} 
\def\confName{WACV}
\def\confYear{2027}

\usepackage{makecell}

\makeatletter
\apptocmd{\@maketitle}{\@thanks}{}{}
\makeatother

\title{Preserving Subject Clarity in Image Outpainting with Multiscale Wavelet Supervision}
\author{
\textbf{Abhilash Neog$^{1,2}$}\thanks{Work was done during an internship at Microsoft.} \quad
\textbf{Taewan Kim$^{1}$} \quad
\textbf{Yi Wu$^{1}$} \quad
\textbf{Xu Chen$^{1}$} \quad
\textbf{Jian Jiao$^{1}$}
\\[0.5em]
$^{1}$Microsoft \quad $^{2}$Virginia Tech
}

\begin{document}
\maketitle

\begin{abstract}


\TK{
Commercial and advertising images are frequently affected by poor framing, partially cropped subjects, truncated text or logos, and insufficient context, all of which can reduce \textbf{subject clarity}, i.e., the ability of an image to clearly communicate its primary subject. Image outpainting offers a scalable solution by extending image boundaries and recovering missing content and context. However, existing diffusion-based outpainting methods often produce visually plausible completions while degrading subject fidelity through structural inconsistencies, semantic drift, or loss of fine-grained detail. \ANedit{To address this limitation, we propose} a subject clarity outpainting framework that combines vision-language model (VLM)-guided semantic conditioning with multiscale wavelet supervision for subject-localized detail preservation. To support training, we develop a subject-centric data curation pipeline that constructs subject-intersecting outpainting pairs from advertising and natural images. The resulting objective introduces no additional inference cost and \ANedit{is designed to be compatible with diffusion-based backbones}. \ANedit{Across four advertising and natural-image benchmarks, our method improves subject clarity, reducing subject-centered DreamSim error and FID on average by 3.0\% and 2.4\% over matched supervised fine-tuning, and by 10.8\% and 7.7\% over the strongest state-of-the-art approach per dataset, respectively.}
}
\end{abstract}    
\section{Introduction}
\label{sec:intro}
\begin{figure}[!htbp]
  \centering
  \setlength{\tabcolsep}{1pt}
  \begin{tabular}{@{}cccc@{}}
    \small\textbf{Input / GT} &
    \small\textbf{PowerPaint} &
    \small\textbf{FLUX.1 Fill} &
    \small\textbf{Ours} \\[-1pt]

    \includegraphics[width=0.245\linewidth]{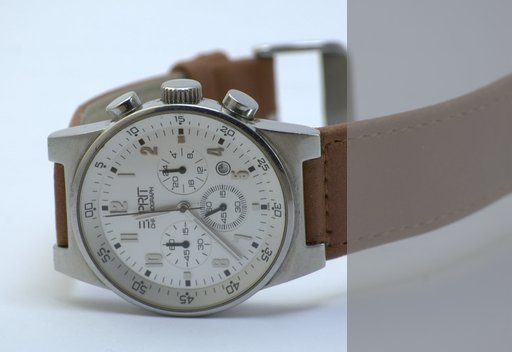} &
    \includegraphics[width=0.245\linewidth]{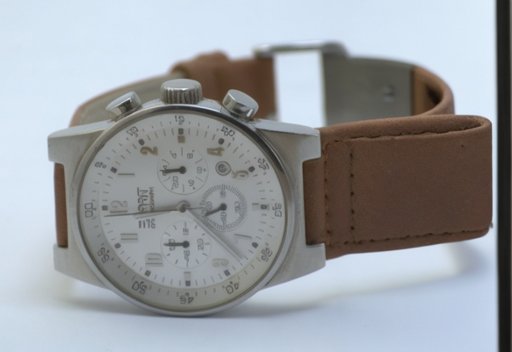} &
    \includegraphics[width=0.245\linewidth]{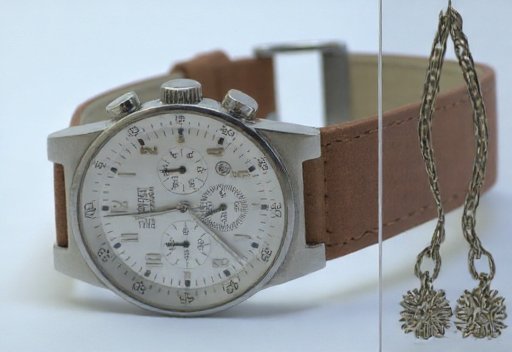} &
    \includegraphics[width=0.245\linewidth]{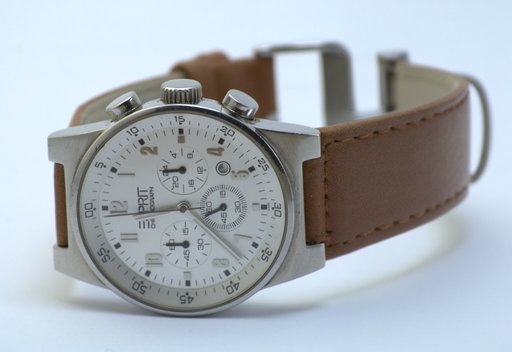} \\[1pt]

    \includegraphics[width=0.245\linewidth]{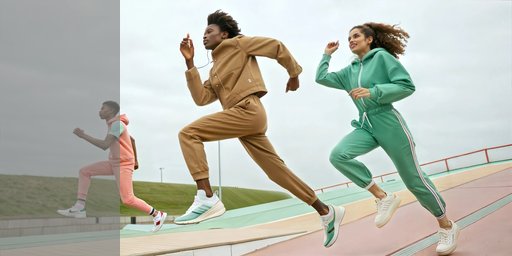} &
    \includegraphics[width=0.245\linewidth]{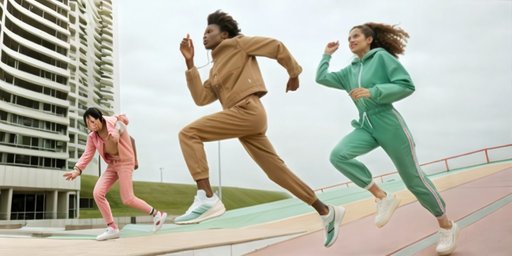} &
    \includegraphics[width=0.245\linewidth]{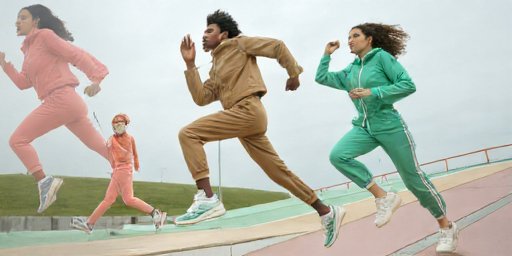} &
    \includegraphics[width=0.245\linewidth]{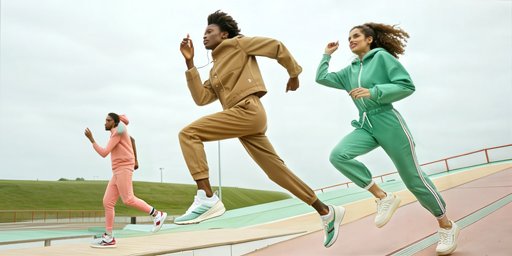} \\[1pt]

    \includegraphics[width=0.245\linewidth]{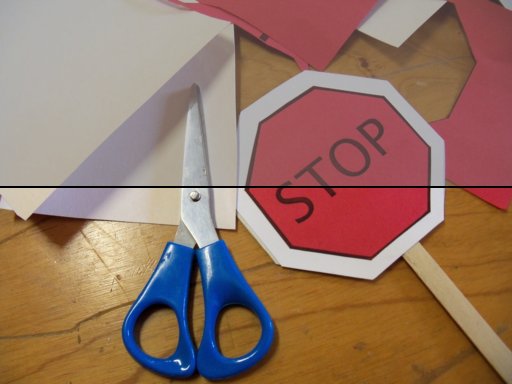} &
    \includegraphics[width=0.245\linewidth]{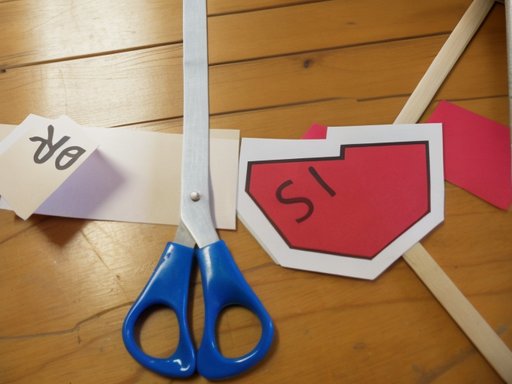} &
    \includegraphics[width=0.245\linewidth]{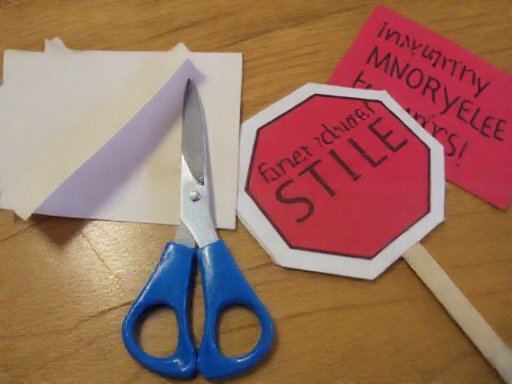} &
    \includegraphics[width=0.245\linewidth]{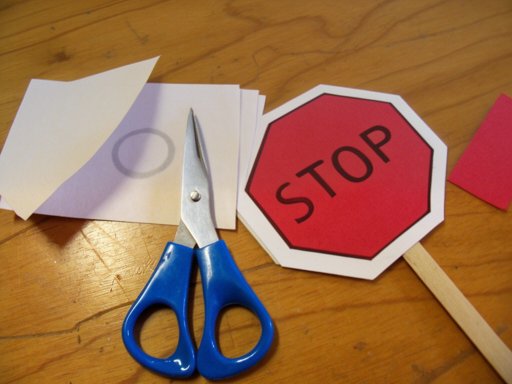}

  \end{tabular}
  \caption{\AN{\textbf{What is subject clarity outpainting?}
    General outpainting extends an image beyond its boundaries to synthesize missing scene content. We focus on the regime in which the original framing partially crops the primary subject. In this setting, outpainting must reconstruct the missing subject region while preserving \emph{subject fidelity}---identity, structure, fine-grained appearance, and continuity with the visible portion. Existing methods can generate plausible surroundings yet distort the continued subject.}}
  \label{fig:motivation}
\end{figure}


\TK{
Images are often used to communicate information through a visually prominent primary subject. However, partially cropped subjects, incomplete framing, truncated logos or text, and insufficient surrounding context can make the intended subject difficult to recognize despite being present in the image. Such issues are particularly impactful in advertising, where clear communication of the primary subject is essential. We refer to this challenge as \emph{subject clarity}, defined as the ability of an image to clearly convey its primary subject through visibility, recognizability, contextual support, and fine-grained appearance detail. While subject clarity may be degraded by a variety of factors, including occlusion and weak foreground-background separation, we focus on failures arising from partial subject cropping, incomplete framing, and insufficient surrounding context, which are naturally addressed through image outpainting. By extending image boundaries and recovering missing content and context, outpainting provides a scalable mechanism for improving subject clarity without requiring additional image capture.
}


\TK{
Improving subject clarity through outpainting introduces a complementary challenge: preserving the fidelity of the original subject. An outpainted image is useful only if the main subject remains recognizable and visually intact after completion. This requires preserving subject-defining characteristics such as structure, texture, typography, branding cues, and other fine-grained appearance details while synthesizing coherent surrounding content. We therefore focus on the practically important \emph{subject-intersecting} setting of image outpainting, where the generated region overlaps the primary subject and its continuation must remain faithful to the visible portion.
}

\AN{
Modern diffusion-based inpainting and outpainting models produce globally convincing completions and incorporate mechanisms for structural, semantic, or contextual consistency. Despite these advances, preserving the unseen continuation of a subject intersected by the outpainting boundary remains challenging. We observe that representative fill models such as FLUX.1 Fill~\cite{flux2024}, BrushNet~\cite{ju2024brushnet}, and PowerPaint~\cite{zhuang2024task}, as well as training-free guidance methods~\cite{moufad2026efficient} built on rectified-flow backbones, can soften subject boundaries, drift textures, distort text, or hallucinate spurious structure where the generated region continues a partially visible subject. Such outputs may remain globally plausible while being locally inconsistent with the visible subject (\cref{fig:motivation}). These observations suggest that global realism and subject fidelity are related but distinct qualities, motivating explicit supervision of subject-localized fidelity in the subject-intersecting outpainting setting.
}

\AN{A natural remedy is supervised fine-tuning (SFT) on in-domain, subject-centric outpainting pairs. We find that SFT improves subject fidelity over off-the-shelf models, but a residual gap remains. We hypothesize that standard SFT does not sufficiently encourage the model to deploy subject-defining information selectively within the generated continuation. Such information is often carried by localized high-frequency structures, including boundaries, textures, text, whereas much of the outpainted image consists of lower-frequency scene context. \ANedit{Hence, we ask, \textit{can we emphasize subject-localized, high-frequency detail during training to improve subject fidelity and, in turn, subject clarity?}}
}
\TK{
A Fourier-domain representation provides a straightforward way to expose frequency discrepancies. However, Fourier coefficients are inherently global and do not directly localize where those discrepancies occur. A multilevel discrete wavelet transform (DWT), by contrast, preserves spatial support while decomposing detail by scale and orientation, making it well suited to subject-intersecting outpainting. We therefore adopt DWT-based supervision for subject-localized detail preservation.
}

Guided by this hypothesis, we develop an automated curation pipeline and construct a 12.6k subject-centric outpainting dataset spanning synthetic advertising images (Synth-Ads), real advertising images (Real-Ads), and the instance-annotated LVIS~\cite{gupta2019lvis} and Open Images~\cite{kuznetsova2020open} datasets. We further introduce a multilevel DWT objective that supervises pseudo-clean predictions in wavelet-detail space over the subject-intersecting outpainting region through subject-adaptive subband weighting and noise-aligned scale scheduling. This objective complements standard SFT, adds no model parameters, and leaves inference unchanged.


\TK{
Our experiments show that explicit supervision of subject-localized high-frequency detail consistently improves subject clarity across advertising and natural-image benchmarks. To summarize, our main contributions are:
\begin{itemize} 
\item We formulate \emph{subject-clarity outpainting} as a distinct outpainting setting in which \emph{subject fidelity} is a primary requirement complementary to global completion quality. 
\item We develop a subject-centric data curation pipeline that combines multimodal subject understanding and segmentation-based localization to construct subject-intersecting pairs, yielding a 12.6k subject clarity outpainting dataset spanning advertising and natural images. 
\item We propose a subject-clarity outpainting framework that integrates VLM-guided semantic conditioning with subject-localized, multilevel DWT supervision. The proposed objective incorporates adaptive subband weighting and noise-aligned scale scheduling to explicitly preserve subject-relevant, high-frequency details without introducing additional inference-time cost.

\end{itemize}
}
\section{Related Work}
\label{sec:related_works}

\paragraph{Diffusion- and flow-based image outpainting.}
\TK{
Image completion is commonly formulated as \emph{inpainting}, where a generative model synthesizes missing content conditioned on visible context. Recent diffusion-based methods built upon latent diffusion models~\cite{rombach2022high} have achieved strong performance through masked-image conditioning, including BrushNet~\cite{ju2024brushnet}, PowerPaint~\cite{zhuang2024task}, and HD-Painter~\cite{manukyan2025hd}. More recent image generation and editing systems have shifted toward multimodal diffusion transformers~\cite{esser2024scaling}, together with flow-matching objectives~\cite{lipman2022flow,liu2022flow}, which underpin modern models such as FLUX Fill~\cite{flux2024}. These models can also be specialized for outpainting through lightweight adaptation, as exemplified by the fal/FLUX.2 outpainting adapter~\cite{fal2024outpaint}. A complementary line of work improves image completion without additional training. FlowChef~\cite{patel2024steering} steers rectified-flow trajectories toward image constraints, while DING~\cite{moufad2026efficient} performs efficient zero-shot image completion through decoupled diffusion guidance. Our work instead focuses on the subject-intersecting outpainting setting, where successful completion depends not only on global realism but also on preserving subject fidelity across visible and generated subject regions.
}
\vspace{-1em}
\paragraph{Frequency-aware generation and detail preservation.}
\TK{

Frequency-domain representations have been used to improve fidelity, efficiency, and detail preservation in image restoration and generation. Prior work includes Fourier-based diffusion inpainting~\cite{hu2022diffusion}, wavelet-based inpainting and generation~\cite{li2021detail,phung2023wavelet}, wavelet-enhanced diffusion for high-resolution image synthesis and video super-resolution~\cite{sigillo2025latent,chenwevsr2026}, and frequency-aware flow matching~\cite{pavasovic2026wait}. HiFi-Inpaint~\cite{liu2026hifi} uses high-frequency priors from a reference image for inpainting, whereas our outpainting setting relies only on the visible image context. Latent Wavelet Diffusion (LWD)~\cite{sigillo2025latent} emphasizes detail-rich regions identified by wavelet energy. In contrast, we retain the scene-level outpainting objective and apply multilevel wavelet supervision specifically within the subject-intersecting region.
}

\section{Method}
\label{sec:method}
\begin{figure*}[t]
  \centering
  \includegraphics[width=0.8\textwidth]{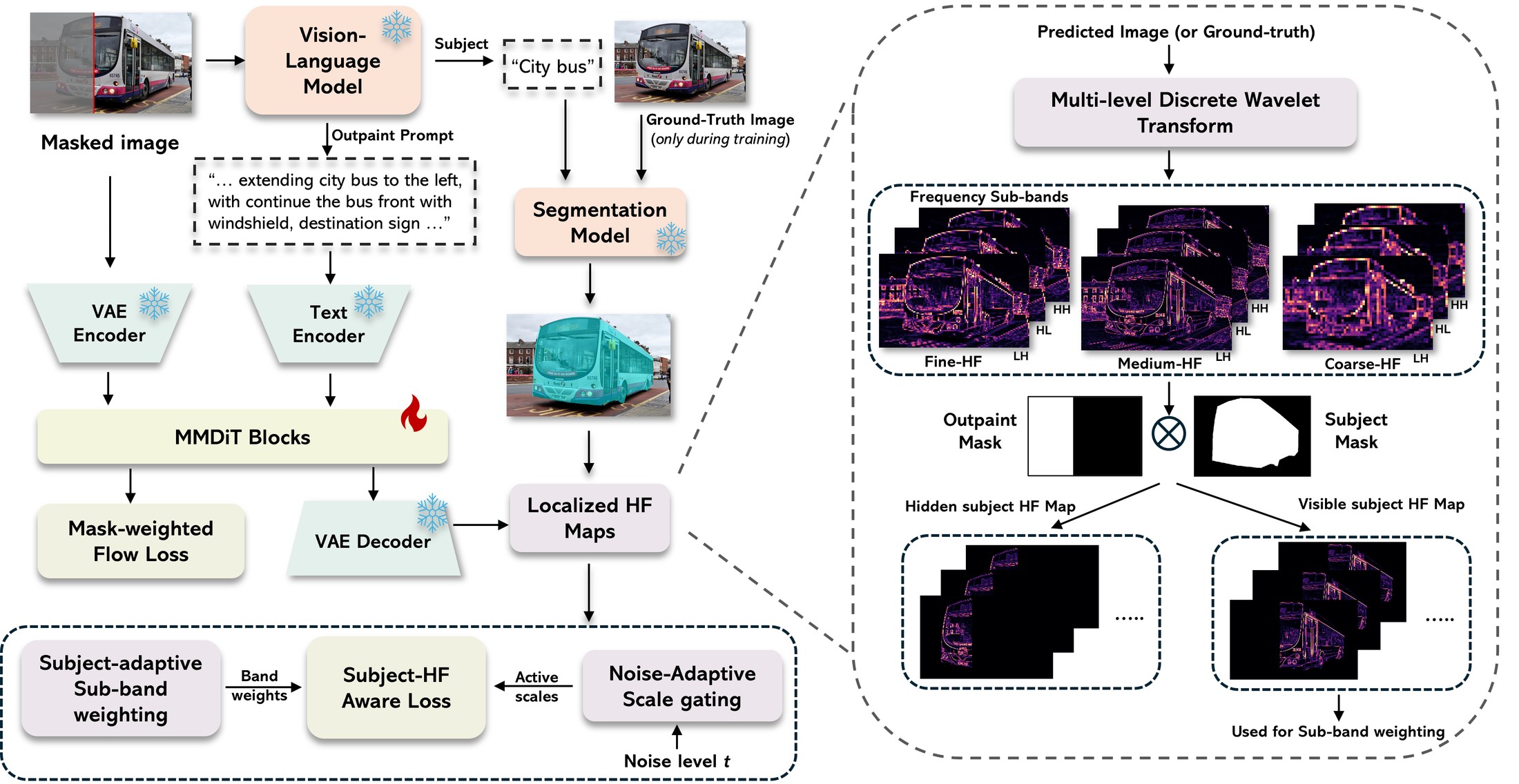}
  \caption{\AN{\textbf{Overview of the proposed training framework.}
    vision--language model provides scene guidance to the outpainting
    backbone. During training, the predicted and ground-truth images are
    compared in a multilevel wavelet space within the affected subject region.
    Subject-frequency statistics determine the relative importance of wavelet
    subbands, while the noise level controls the active scales. The resulting
    subject-aware wavelet loss complements the standard flow objective and is
    not used during inference.}}
  \label{fig:method-overview}
\end{figure*}

\TK{
Our approach combines subject-centric data curation with subject-localized multilevel wavelet supervision. We first present the rectified-flow formulation for image outpainting, formalize subject-clarity outpainting, and introduce our mask-weighted flow objective (\cref{sec:formulation}). We then describe our subject-centric data-curation pipeline (\cref{sec:dataset}) and the proposed wavelet-based training objective (\cref{sec:hfloss}). \Cref{fig:method-overview} summarizes the proposed training framework.
}

\subsection{Outpainting Formulation}
\label{sec:formulation}

\TK{
\paragraph{Outpainting with rectified flow.}
Let $I\in\mathbb{R}^{H\times W\times 3}$ denote a complete image and $M\in\{0,1\}^{H\times W}$ a binary \emph{fill mask}, where $M{=}1$ marks the region to be synthesized and $M{=}0$ the visible region. Given the observation $I_{\mathrm{obs}}=(1-M)\odot I$ and the fill mask $M$, image outpainting aims to synthesize the missing content within $M$ so that the completed image is consistent with the visible content. 

We adopt a rectified-flow formulation in the latent space of a frozen VAE with encoder $\mathcal{E}$ and decoder $\mathcal{D}$, and write $x_0=\mathcal{E}(I)$ for the clean latent. The model receives text-based scene guidance $c$ and a conditioning representation $z_c$ derived from $(I_{\mathrm{obs}},M)$. We independently sample a Gaussian noise latent $x_1\sim\mathcal{N}(0,\mathbf{I})$ and a continuous training time $t\in(0,1)$ from a logit-normal distribution~\cite{labs2025flux}. The intermediate latent is constructed along the linear path 
\begin{equation}
    x_t=(1-t)x_0+t x_1.
    \label{eq:interp} 
\end{equation} 
The corresponding conditional velocity target and model prediction are 
\begin{equation} 
    u=\frac{d x_t}{dt}=x_1-x_0, 
    \qquad 
    \hat{v}=v_\theta(x_t,t,c,z_c), 
    \label{eq:velocity} 
\end{equation} 
where $v_\theta$ denotes the learned rectified-flow velocity field. While $u$ follows the data-to-noise path, inference integrates the learned velocity field from noise to data.

\paragraph{Subject-clarity outpainting.} 
Let $S\in\{0,1\}^{H\times W}$ denote the subject mask. We define the affected and visible subject regions as 
\begin{equation}
    \Omega=M\odot S, 
    \qquad 
    V=(1-M)\odot S, 
    \label{eq:subject-regions} 
\end{equation} 
respectively. We focus on the subject-intersecting setting, where the fill region overlaps the primary subject, i.e., $\Omega\neq\emptyset$. We refer to this setting as \emph{subject-clarity outpainting}. Its objective is to synthesize a coherent continuation of $V$ within $\Omega$ while preserving subject fidelity, including identity, structure, fine-grained appearance, and continuity across the outpaint boundary. Because full-image objectives can be dominated by unchanged visible content and surrounding background, they may underemphasize localized subject errors in $\Omega$, motivating our subject-centric data curation and supervision.

\paragraph{Mask-weighted flow training.} A standard flow-matching objective weights all latent tokens uniformly, although the fill region is the primary region to be synthesized. We therefore assign greater weight to fill-region tokens while retaining supervision over the visible region. Let $m_i\in[0,1]$ denote the fill-mask value for packed latent token $i$, obtained by resizing $M$ to the latent patch-token grid. We define the mask-weighted flow loss as
\begin{equation} 
    \mathcal{L}_{\mathrm{flow}} = \frac{\sum_i w_i \left\lVert \hat{v}_i-u_i \right\rVert_2^2} {\sum_i w_i}, 
    \qquad 
    w_i=1+\lambda_m m_i. 
    \label{eq:flowloss} 
\end{equation} 
Here, $\lambda_m$ controls the additional weight assigned to fill-region tokens; we use $\lambda_m=4$, selected empirically.
}

\subsection{Subject-Centric Outpainting Data Curation}
\label{sec:dataset}


\TK{
We curate subject-centric outpainting pairs from our synthetic and real advertising datasets, Synth-Ads and Real-Ads, to provide target-domain coverage. We further incorporate the instance-annotated LVIS~\cite{gupta2019lvis} and Open Images~\cite{kuznetsova2020open} datasets to increase object and scene diversity. Our curation pipeline consists of four stages (see \cref{sec:appendix_data_curation} for full details):
}

\begin{enumerate}
  \item \textbf{Raw image screening.} 
    \TK{
    We remove low-resolution images and use a VLM to reject advertising images without a clear primary subject, with purely ambient content, or with unsupported plain backgrounds; LVIS and Open Images rely on their instance annotations instead.
    }

  \item \textbf{Subject-mask generation.} \ANedit{For maskless advertising
    images, Sa2VA~\cite{yuan2025sa2va} segments the VLM-identified primary
    subject. For LVIS and Open Images, we union qualifying ground-truth
    instances.}

  \item \textbf{Mask-dependent filtering.} \ANedit{We retain sufficiently
    large, visually structured subjects (subject area $\geq5\%$). Internal-edge
    filtering excludes smooth or textureless subjects, while external-edge
    filtering for instance-masked sources excludes plain backgrounds.}

  \item \textbf{Subject-adaptive pair generation.} 
    \TK{
    We generate outpainting pairs by sampling a crop from one canvas side, with the crop extent adapted to subject size so that small subjects remain sufficiently visible. We partition the resulting 12,648 pairs into training, validation, and test sets; see \cref{tab:dataset_splits} for the detailed breakdown.}
\end{enumerate}

\subsection{\ANedit{Subject-localized Multilevel Wavelet Supervision}}
\label{sec:hfloss}

\TK{
The latent-space flow objective in \cref{eq:flowloss} does not explicitly
distinguish the localized high-frequency details important for subject
fidelity. We therefore introduce an auxiliary pixel-space wavelet loss
restricted to the subject-intersecting region.
}

\paragraph{Pseudo-clean recovery.}
\TK{ 
Given the predicted velocity, we estimate the clean latent by inverting \cref{eq:interp} under $\hat{v}\approx u$: 
\begin{equation} 
    \hat{x}_0=x_t-t\hat{v}, 
    \qquad 
    \hat{I}=\mathcal{D}(\hat{x}_0). 
    \label{eq:pseudox0} 
\end{equation} 
The decoded pseudo-clean image $\hat{I}$ enables direct supervision in pixel-space wavelet coefficients.
}

\paragraph{Multilevel wavelet detail space.}
\TK{
We use a fixed Haar DWT to obtain a parameter-free, interpretable decomposition of image detail across spatial scales and orientations. Unlike global frequency representations, wavelet coefficients retain spatial support, allowing the loss to localize detail discrepancies to the affected subject region. We apply a $J$-level 2D DWT to the pseudo-clean prediction $\hat{I}$ and complete image $I$, obtaining directional detail subbands $\mathcal{W}_{j,o}(\cdot)$ at level $j\in\{1,\ldots,J\}$ and orientation $o\in\{\mathrm{LH},\mathrm{HL},\mathrm{HH}\}$. These subbands jointly capture horizontal, vertical, and diagonal detail at multiple spatial resolutions, enabling supervision to vary by scale and orientation. We use $J=3$ levels, indexed from fine to coarse, to capture fine edges, intermediate textures, and coarser structural detail.
}

\paragraph{Subject region.}
\TK{
We restrict wavelet supervision to the generated subject region $\Omega=M\odot S$, preventing background detail from dominating the auxiliary objective. To avoid interpolation artifacts near the outpaint boundary, we use the slightly eroded support $\widetilde{\Omega}=\operatorname{erode}(M)\odot S$.
}

\paragraph{Subject-frequency adaptive weighting.}
\AN{Not all wavelet bands are equally important for a given subject: text and
sharp boundaries emphasize fine directional bands, whereas smoother shapes
place more energy at coarser scales. We estimate these statistics from the
\emph{visible} subject region, which provides subject-specific evidence without
using the hidden completion, 
\TK{and use them to distribute supervision across scale-orientation bands. 
We define the per-sample subband weights $\alpha_{j,o}$ as:}} 
\begin{equation}
  \alpha_{j,o} = (1-\lambda)\,\alpha^{\mathrm{uni}}_{j,o} + \lambda\,\alpha^{\mathrm{adapt}}_{j,o},
  \qquad
  \sum_{j,o}\alpha_{j,o}=1.
  \label{eq:wavelet-weights}
\end{equation}
\AN{
\ANedit{
Here, $\alpha^{\mathrm{uni}}_{j,o}=1/(3J)$ assigns equal weight to
every scale-orientation band, whereas $\alpha^{\mathrm{adapt}}_{j,o}$ is 
\TK{
the normalized mean absolute wavelet response within the visible subject region for band $(j,o)$.
} 
} 
The uniform component prevents a single noisy or highly textured band from 
\TK{
dominating the supervision. We use $\lambda=0.75$, selected empirically, to favor adaptive weighting while retaining a uniform prior.
}
}

\paragraph{Noise-aligned scale scheduling.}
\TK{
The fidelity of the pseudo-clean estimate $\hat{x}_0$ generally improves as $t$ approaches the clean endpoint. Coarse structural information can be supervised at moderately noisy timesteps, whereas fine-scale detail becomes more reliable at lower noise levels. We therefore apply level-wise gates
}
\begin{equation}
  g_j(t)=\sigma\!\bigl(\kappa\,(\tau_j-t)\bigr),
  \label{eq:gate}
\end{equation}
\TK{
where $\sigma$ is the logistic function, $\kappa$ controls the gate sharpness, and $\tau_j$ is the activation threshold for level $j$. We use $(\tau_1,\tau_2,\tau_3)=(0.35,0.50,0.65)$ for levels ordered from fine to coarse, allowing coarse-scale supervision to remain active at higher noise levels while concentrating fine-scale supervision near the clean endpoint.
}

\paragraph{Wavelet subject clarity loss.}
\AN{The final auxiliary term is a gated, masked, and weighted $L_1$ loss over
wavelet subbands in the subject-overlap region:}
\begin{equation}
  \mathcal{L}_{\mathrm{hf}}
  = \sum_{j=1}^{J}\sum_{o}
      g_j(t)\,\alpha_{j,o}\,
      \frac{\sum_{p\in\widetilde{\Omega}_j}
      \bigl\lVert \mathcal{W}_{j,o}(\hat{I})_p
      - \mathcal{W}_{j,o}(I)_p \bigr\rVert_1}
      {C\,\lvert\widetilde{\Omega}_j\rvert},
  \label{eq:hfloss}
\end{equation}
\AN{where $C$ is the number of channels, $\widetilde{\Omega}_j$ denotes the
eroded subject-overlap mask downsampled to level $j$, and $p$ indexes \TK{spatial locations}.
This loss preserves subject-relevant edges and textures while staying
compatible with the base flow objective. We use an $L_1$ distance because
wavelet coefficients are sparse and heavy-tailed, \TK{making it less sensitive to } outliers than an $L_2$ penalty.}
\begin{algorithm}[t]
  \caption{\AN{Subject clarity-aware outpainting training}}
  \label{alg:training}
  \small
  \begin{algorithmic}[1]
    \Require Target $I$, fill mask $M$, subject mask $S$, prompt $c$
    \Require Trainable parameters $\theta$; frozen $\mathcal{E},\mathcal{D}$
    \State $x_0\gets\mathcal{E}(I)$; construct conditioning latent
      $z_c$ from $\bigl(I\odot(1-M),M\bigr)$
    \State Sample $x_1\sim\mathcal{N}(0,\mathbf{I})$ and
      $t=\sigma(\varepsilon)$, $\varepsilon\sim\mathcal{N}(0,1)$
    \State $x_t\gets(1-t)x_0+t x_1$;\quad $u\gets x_1-x_0$
    \State $\hat v\gets v_\theta(x_t,t,c,z_c)$
    \State Compute $\mathcal{L}_{\mathrm{flow}}$ using the
      mask-weighted token loss in \cref{eq:flowloss}
    \State $\hat I\gets\mathcal{D}(x_t-t\hat v)$;\quad
      $\widetilde{\Omega}\gets\operatorname{erode}(M)\odot S$
    \State Compute adaptive weights $\alpha_{j,o}$ from normalized DWT
      energy in the visible subject region $(1-M)\odot S$
    \State Compute timestep gates
      $g_j(t)\gets\sigma\!\left(\kappa(\tau_j-t)\right)$, with
      $\tau_j\in\{0.35,0.50,0.65\}$ from fine to coarse
    \For{$j=1,\ldots,J$ and $o\in\{\mathrm{LH},\mathrm{HL},\mathrm{HH}\}$}
      \State Downsample $\widetilde{\Omega}$ to level $j$ and accumulate
      $g_j(t)\alpha_{j,o}
      \lVert\mathcal{W}_{j,o}(\hat I)-\mathcal{W}_{j,o}(I)\rVert_1$
      over the selected coefficients
    \EndFor
    \State Normalize the accumulated terms to obtain
      $\mathcal{L}_{\mathrm{hf}}$ in \cref{eq:hfloss}
    \State $\mathcal{L}\gets\mathcal{L}_{\mathrm{flow}}+
      \lambda_{\mathrm{hf}}\mathcal{L}_{\mathrm{hf}}$
    \State Update only $\theta$ by backpropagating $\mathcal{L}$
  \end{algorithmic}
\end{algorithm}

\paragraph{\ANedit{Full objective.}}
\label{sec:fullobj}

\AN{The total training loss combines the two terms:}
\begin{equation}
  \mathcal{L} = \mathcal{L}_{\mathrm{flow}}
    + \lambda_{\mathrm{hf}}\,\mathcal{L}_{\mathrm{hf}}.
  \label{eq:total}
\end{equation}
\TK{
We use $\lambda_{\mathrm{hf}}=0.1$, selected empirically, to control the contribution of the auxiliary wavelet loss, and optimize the low-rank adapter parameters $\theta$. The wavelet loss requires only subject masks as additional training supervision, introduces no model parameters, and leaves inference unchanged. The complete training procedure is summarized in \cref{alg:training}.
}

\vspace{-1em}
\section{Experiments}
\label{sec:experiments}

\begin{table*}[!hbpt]
  \centering
  \caption{\AN{Outpainting results on the Synth-Ads and Real-Ads test sets.
  \TK{Per-dataset best and second best are \textbf{bold} and \underline{underlined}, respectively.}
  }}
  \vspace{-1.5em}
  \label{tab:ad_results}
  \footnotesize
  \setlength{\tabcolsep}{4pt}
  \resizebox{\textwidth}{!}{%
  \begin{tabular}{llccccc|ccccc}
    \toprule
    \textbf{Method} & \textbf{Backbone}
      & \multicolumn{5}{c|}{\textbf{Synth-Ads}}
      & \multicolumn{5}{c}{\textbf{Real-Ads}} \\
    \cmidrule(lr){3-7}\cmidrule(lr){8-12}
    & & \multicolumn{2}{c}{\textbf{Overall}}
      & \multicolumn{3}{c|}{\textbf{Subject Clarity}}
      & \multicolumn{2}{c}{\textbf{Overall}}
      & \multicolumn{3}{c}{\textbf{Subject Clarity}} \\
    \cmidrule(lr){3-4}\cmidrule(lr){5-7}
    \cmidrule(lr){8-9}\cmidrule(lr){10-12}
    & & \textbf{FID}\,$\downarrow$ & \textbf{LPIPS}$_{\mathrm{full}}$\,$\downarrow$
      & \textbf{LPIPS}$_\Omega$\,$\downarrow$
      & \textbf{DINOv2}$_\Omega$\,$\downarrow$
      & \textbf{DreamSim}$_{\mathrm{ctx}}$\,$\downarrow$
      & \textbf{FID}\,$\downarrow$ & \textbf{LPIPS}$_{\mathrm{full}}$\,$\downarrow$
      & \textbf{LPIPS}$_\Omega$\,$\downarrow$
      & \textbf{DINOv2}$_\Omega$\,$\downarrow$
      & \textbf{DreamSim}$_{\mathrm{ctx}}$\,$\downarrow$ \\
    \midrule
    BrushNet~\cite{ju2024brushnet} & SDXL
      & 31.41 & 0.2590 & 0.5015 & 0.4586 & 0.0743
      & 74.90 & 0.2839 & 0.5528 & 0.5124 & 0.1096 \\
    PowerPaint~\cite{zhuang2024task} & SD\,1.5
      & 31.79 & 0.2590 & 0.4152 & 0.3789 & 0.0815
      & 72.77 & 0.2741 & 0.4799 & 0.4453 & 0.1184 \\
    VIP~\cite{yang2024vip} & SD\,1.5
      & 25.26 & 0.2118 & 0.3957 & 0.3490 & 0.0514
      & 55.20 & 0.2113 & 0.4064 & 0.3593 & 0.0783 \\
    FLUX.1 Fill~\cite{flux2024} & FLUX\,1.0
      & 29.56 & 0.2327 & 0.3542 & 0.3157 & 0.0644
      & 86.92 & 0.3148 & 0.4545 & 0.4083 & 0.1555 \\
    FLUX2+DING~\cite{moufad2026efficient} & FLUX\,2.0
      & 34.65 & 0.2940 & 0.5590 & 0.4895 & 0.0966
      & 71.95 & 0.2806 & 0.5567 & 0.4808 & 0.1268 \\
    FLUX2+FlowChef~\cite{patel2024steering} & FLUX\,2.0
      & 37.79 & 0.3165 & 0.6309 & 0.5561 & 0.1061
      & 73.19 & 0.2905 & 0.5953 & 0.5210 & 0.1275 \\
    fal/FLUX2-lora-outpaint~\cite{fal2024outpaint} & FLUX\,2.0
      & 24.89 & 0.2195 & 0.3837 & 0.3376 & 0.0543
      & 52.51 & 0.2120 & 0.4066 & 0.3517 & 0.0787 \\
    FLUX.2 SFT & FLUX\,2.0
      & \underline{23.61} & \underline{0.2099} & \textbf{0.3527}
      & \underline{0.3151} & \underline{0.0477}
      & \underline{50.13} & \underline{0.2019} & \underline{0.3735}
      & \underline{0.3218} & \underline{0.0663} \\
    \midrule
    \textit{FLUX.2 SFT-HF-Aware (ours)} & FLUX\,2.0
      & \textbf{23.42} & \textbf{0.2089} & \underline{0.3539}
      & \textbf{0.3150} & \textbf{0.0472}
      & \textbf{48.26} & \textbf{0.2009} & \textbf{0.3662}
      & \textbf{0.3169} & \textbf{0.0631} \\
    \bottomrule
  \end{tabular}}
\end{table*}

\begin{table*}[!hbpt]
  \centering
  \caption{\AN{Outpainting results on the LVIS and Open Images test sets.
  \TK{Per-dataset best and second best are \textbf{bold} and \underline{underlined}, respectively.}
  }}
  \vspace{-1.5em}
  \label{tab:lvis_oi_results}
  \footnotesize
  \setlength{\tabcolsep}{4pt}
  \resizebox{\textwidth}{!}{%
  \begin{tabular}{llccccc|ccccc}
    \toprule
    \textbf{Method} & \textbf{Backbone}
      & \multicolumn{5}{c|}{\textbf{LVIS}}
      & \multicolumn{5}{c}{\textbf{Open Images}} \\
    \cmidrule(lr){3-7}\cmidrule(lr){8-12}
    & & \multicolumn{2}{c}{\textbf{Overall}}
      & \multicolumn{3}{c|}{\textbf{Subject Clarity}}
      & \multicolumn{2}{c}{\textbf{Overall}}
      & \multicolumn{3}{c}{\textbf{Subject Clarity}} \\
    \cmidrule(lr){3-4}\cmidrule(lr){5-7}
    \cmidrule(lr){8-9}\cmidrule(lr){10-12}
    & & \textbf{FID}\,$\downarrow$ & \textbf{LPIPS}$_{\mathrm{full}}$\,$\downarrow$
      & \textbf{LPIPS}$_\Omega$\,$\downarrow$
      & \textbf{DINOv2}$_\Omega$\,$\downarrow$
      & \textbf{DreamSim}$_{\mathrm{ctx}}$\,$\downarrow$
      & \textbf{FID}\,$\downarrow$ & \textbf{LPIPS}$_{\mathrm{full}}$\,$\downarrow$
      & \textbf{LPIPS}$_\Omega$\,$\downarrow$
      & \textbf{DINOv2}$_\Omega$\,$\downarrow$
      & \textbf{DreamSim}$_{\mathrm{ctx}}$\,$\downarrow$ \\
    \midrule
    BrushNet~\cite{ju2024brushnet} & SDXL
      & 19.91 & 0.2587 & 0.5173 & 0.4910 & 0.0991
      & 60.41 & 0.2674 & 0.5159 & 0.5100 & 0.1102 \\
    PowerPaint~\cite{zhuang2024task} & SD\,1.5
      & 22.49 & 0.2788 & 0.4530 & 0.4262 & 0.1065
      & 66.27 & 0.2853 & 0.4588 & 0.4520 & 0.1193 \\
    VIP~\cite{yang2024vip} & SD\,1.5
      & 15.82 & \textbf{0.2082} & 0.4107 & 0.3615 & 0.0708
      & 48.18 & \underline{0.2116} & 0.4062 & 0.3628 & 0.0756 \\
    FLUX.1 Fill~\cite{flux2024} & FLUX\,1.0
      & 22.76 & 0.2594 & 0.3953 & 0.3510 & 0.0951
      & 63.10 & 0.2716 & 0.3929 & 0.3686 & 0.1020 \\
    FLUX2+DING~\cite{moufad2026efficient} & FLUX\,2.0
      & 23.10 & 0.2925 & 0.5731 & 0.4999 & 0.1279
      & 64.86 & 0.2981 & 0.5528 & 0.5028 & 0.1383 \\
    FLUX2+FlowChef~\cite{patel2024steering} & FLUX\,2.0
      & 24.51 & 0.3098 & 0.6252 & 0.5553 & 0.1354
      & 71.02 & 0.3214 & 0.6293 & 0.5784 & 0.1472 \\
    fal/FLUX2-lora-outpaint~\cite{fal2024outpaint} & FLUX\,2.0
      & 16.67 & 0.2184 & 0.3942 & 0.3459 & 0.0735
      & 48.25 & 0.2201 & 0.3900 & 0.3537 & 0.0786 \\
    FLUX.2 SFT & FLUX\,2.0
      & \underline{15.61} & 0.2131 & \underline{0.3744}
      & \underline{0.3272} & \underline{0.0677}
      & \underline{43.00} & 0.2119 & \underline{0.3755}
      & \underline{0.3295} & \underline{0.0715} \\
    \midrule
    \textit{FLUX.2 SFT-HF-Aware (ours)} & FLUX\,2.0
      & \textbf{15.36} & \underline{0.2115} & \textbf{0.3714}
      & \textbf{0.3249} & \textbf{0.0658}
      & \textbf{41.51} & \textbf{0.2114} & \textbf{0.3728}
      & \textbf{0.3252} & \textbf{0.0690} \\
    \bottomrule
  \end{tabular}}
  \vspace{-1em}
\end{table*}

\subsection{Setup}
\label{sec:setup}

\noindent \textbf{Implementation details.}
\AN{We build our outpainting model on the FLUX.2-klein-base-4B~\cite{flux2klein2026} rectified-flow transformer and \TK{inject rank-$32$ LoRA adapters~\cite{hu2021lora} into its} \ANedit{transformer blocks}, while keeping the VAE and text encoder frozen. \ANedit{Our model
and the matched SFT baseline are trained for $8$ epochs over the pooled training split of the four datasets, using two NVIDIA RTX A6000 GPUs.} Each image is paired with a VLM-generated scene-guidance prompt. \ANedit{Apart from this prompt-generation step, inference follows the standard FLUX.2-klein-base, without any architectural changes}. \TK{The matched \emph{FLUX.2 SFT} baseline uses the same backbone LoRA configuration, and training setup, but excludes} the wavelet loss.}\\

\vspace{-0.8em}
\TK{
\noindent\textbf{Baselines.} We compare against representative inpainting and outpainting models across several backbones: BrushNet~\cite{ju2024brushnet} (SDXL), PowerPaint~\cite{zhuang2024task} (SD~1.5), VIP~\cite{yang2024vip} (SD~1.5, trained on our data), and FLUX.1 Fill~\cite{flux2024}; FLUX.2-based methods, including the training-free DING~ \cite{moufad2026efficient} and FlowChef~\cite{patel2024steering}, and the public fal/FLUX.2 outpainting adapter~\cite{fal2024outpaint}; and the proprietary GPT-Image-2 editor~\cite{openai2026gptimage2}. All methods receive the same benchmark inputs, with VLM prompts adapted to their native conditioning interfaces where supported. We evaluate native outputs without post-hoc compositing, which could conceal preservation and boundary errors. The only exception is GPT-Image-2, whose API returns a full edited canvas and therefore requires restoration of the original visible region. Further details are provided in \cref{sec:implementation_details}.
}
\\

\vspace{-0.8em}
\noindent\textbf{Evaluation metrics.} 
\TK{
We report two groups of metrics. Overall quality is measured using FID~\cite{heusel2017gans} for distribution-level realism and full-image LPIPS~\cite{zhang2018unreasonable} for perceptual fidelity. To assess subject preservation without dilution from unchanged content, we report LPIPS$_\Omega$, DINOv2$_\Omega$~\cite{oquab2023dinov2}, and DreamSim$_{\mathrm{ctx}}$~\cite{fu2023dreamsim}.
}
\TK{
LPIPS$_\Omega$ restricts the standard LPIPS to the generated subject region, while DINOv2$_\Omega$ averages the cosine distance between corresponding prediction and ground-truth DINOv2 patch embeddings over $\Omega$.
}
DreamSim$_{\mathrm{ctx}}$ evaluates human-aligned perceptual similarity on a
subject-centered crop that retains the visible and generated subject parts
along with nearby scene context. 
Together, these
metrics capture complementary low-level, feature-level, and perceptual aspects
of subject clarity. 

\subsection{Results}
\label{sec:results}

\AN{\ANedit{Our approach improves over matched SFT on both advertising datasets (\cref{tab:ad_results}): Synth-Ads and Real-Ads.
On Real-Ads, ours improves all five metrics, reducing FID by
$3.7\%$, LPIPS$_\Omega$ by $2.0\%$, DINOv2$_\Omega$ by $1.5\%$, and
DreamSim$_{\mathrm{ctx}}$ by $4.8\%$. On Synth-Ads, ours improves both overall
metrics and DreamSim$_{\mathrm{ctx}}$ while remaining comparable on the other
localized metrics. The training-free guidance methods degrade subject metrics despite sharing the same FLUX.2 backbone, indicating that 
trajectory steering alone is insufficient for subject fidelity, while the public outpainting LoRA narrows but does not close the gap to matched SFT.}} 

\begin{figure*}[htbp]
  \centering
  \begin{subfigure}[t]{0.75\textwidth}
    \scriptsize
    \makebox[\linewidth][c]{%
      \makebox[0.166\linewidth][c]{\textbf{Input / GT}}%
      \makebox[0.166\linewidth][c]{\textbf{PowerPaint}}%
      \makebox[0.166\linewidth][c]{\textbf{FLUX.1 Fill}}%
      \makebox[0.166\linewidth][c]{\textbf{fal/FLUX.2 LoRA}}%
      \makebox[0.166\linewidth][c]{\textbf{FLUX.2 SFT}}%
      \makebox[0.166\linewidth][c]{\textbf{Ours}}}
  \end{subfigure}\par\smallskip
  \begin{subfigure}[t]{0.75\textwidth}
    \includegraphics[width=\linewidth]{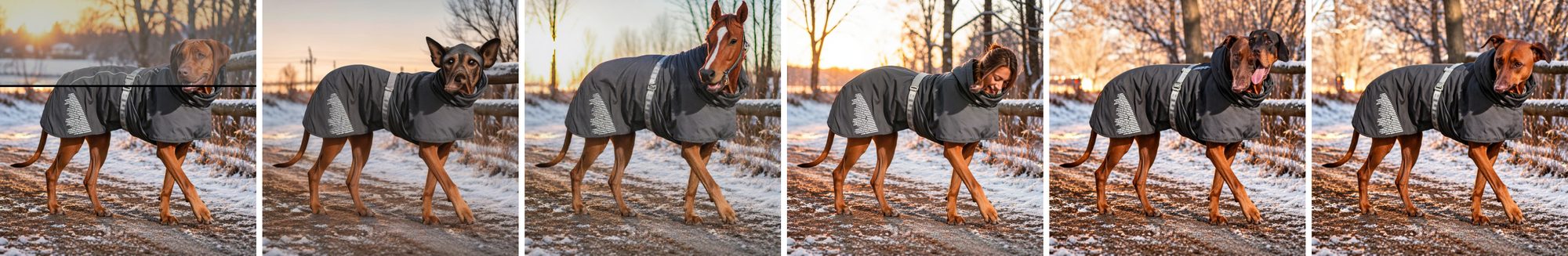}
  \end{subfigure}\par
  \begin{subfigure}[t]{0.75\textwidth}
    \centering
    \setlength{\fboxsep}{4pt}
    \fbox{\begin{minipage}{0.98\linewidth}
      \scriptsize
      \ANedit{\textbf{Example prompt} (our approach):
      \textit{``Fill the gray region by extending the dog wearing a black coat,
      with the dog's upper body, back, neck, and head continuing from the
      visible coat.''}
      Prompt wording differs slightly across baselines}
    \end{minipage}}
  \end{subfigure}\par
  \begin{subfigure}[t]{0.75\textwidth}
    \includegraphics[width=\linewidth, height=0.15\textheight]{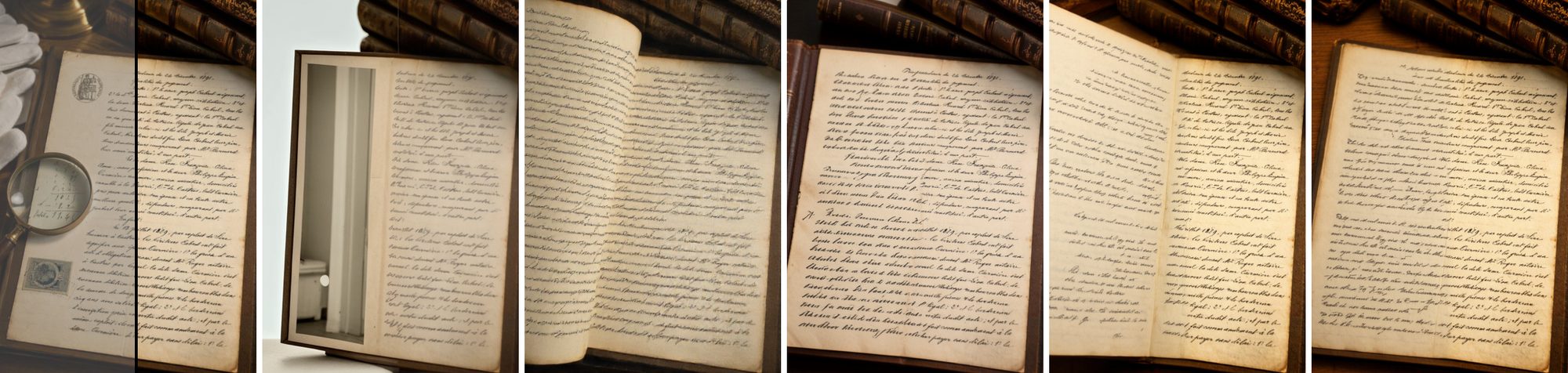}
  \end{subfigure}\par
  \begin{subfigure}[t]{0.75\textwidth}
    \includegraphics[width=\linewidth, height=0.08\textheight]{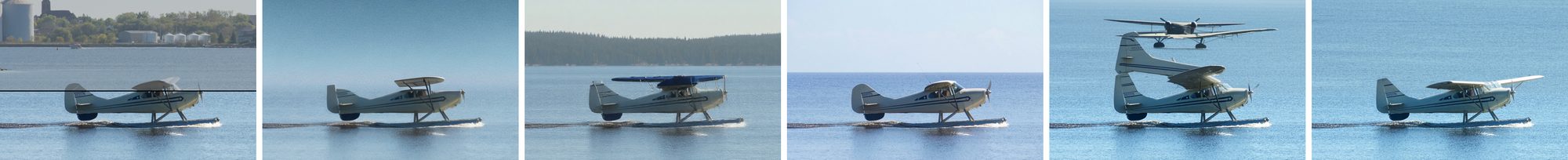}
  \end{subfigure}\par
  \begin{subfigure}[t]{0.75\textwidth}
    \includegraphics[width=\linewidth, height=0.18\textheight]{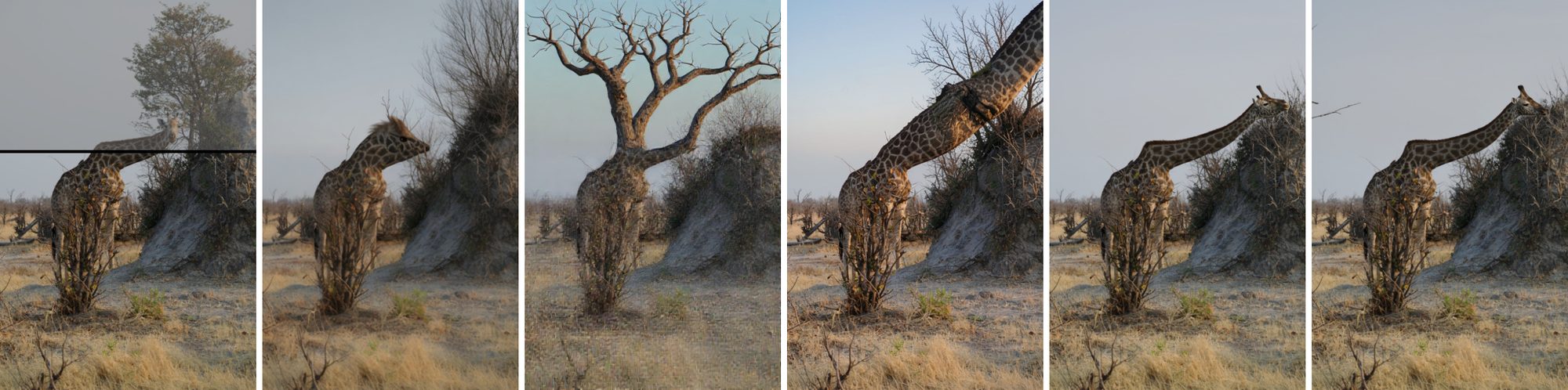}
  \end{subfigure}
  \caption{\ANedit{The Input / GT column shows the model input and ground-truth outpaint region in gray, with the black line marking the outpaint boundary. The prompts
  used for these examples are reported in \cref{tab:qualitative_prompts}.
  } Relative to
  representative external baselines and the matched SFT model, our method more
  consistently preserves subject identity and produces coherent continuations
  across the boundary.}
  \label{fig:qualitative}
\end{figure*}

\ANedit{On the cross-domain LVIS and Open Images benchmarks
(\cref{tab:lvis_oi_results}), ours improves all five metrics over matched SFT.
The largest localized gains are in DreamSim$_{\mathrm{ctx}}$, reduced by
$2.8\%$ on LVIS and $3.5\%$ on Open Images. VIP, trained on the same training split as our model, is competitive on full-image
LPIPS, but this does not translate to the localized subject metrics, for which ours is best on both
datasets.} \ANedit{We also observe that baselines often introduce unsupported
details while extending the subject and surrounding
context. This can produce hallucinated content that distorts identity-defining
structure and reduces subject clarity. Qualitative examples are shown in
\cref{fig:qualitative}.}

\begin{figure}[tb]
  \centering
  \includegraphics[width=0.75\columnwidth]{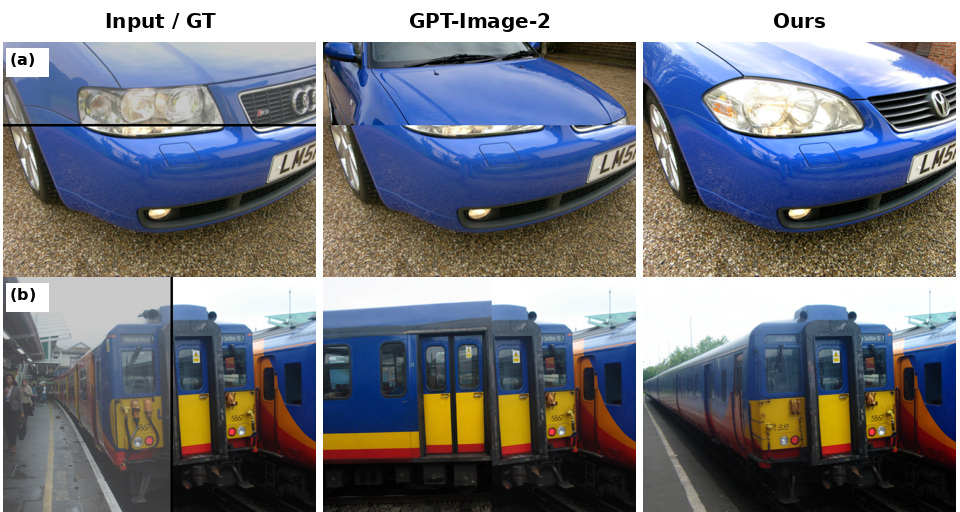}
  \caption{\AN{\textbf{Comparison with GPT-Image-2.}
  \ANedit{Representative comparisons illustrating \TK{(a)} subject-continuation
  misalignment, and \TK{(b)} failure to preserve subject structure.}}}
  \label{fig:gpt2_qualitative}
  \vspace{-1em}
\end{figure}

\TK{
We additionally evaluate GPT-Image-2 by providing the same masked inputs and prompting it to outpaint the missing regions. We report results on the 2,971 of 3,123 test images (95.1\%) for which GPT-Image-2 returns valid outputs, excluding samples rejected by API-level restrictions. Even with the original visible region restored in its outputs, ours outperforms GPT-Image-2 on all metrics for Real-Ads, LVIS, and Open Images (\cref{tab:gpt2_shared_cohort}). Results are mixed on Synth-Ads, whose targets were generated by GPT-Image-2. Qualitatively, GPT-Image-2 produces plausible completions but can distort subject structure or misalign the generated continuation with the visible subject (\cref{fig:gpt2_qualitative}).
}

\begin{table}[tb]
  \centering
  \caption{
  \TK{
  Comparison on test samples with available GPT-Image-2 predictions (95.1\% overall; 97.2\% Synth-Ads,
  96.3\% Real-Ads, 94.1\% LVIS, and 93.6\% Open Image). Best in \textbf{bold}; second best \underline{underlined}. Lower is better.
  }
  \vspace{-1em}
  }
  \label{tab:gpt2_shared_cohort}
  \scriptsize
  \setlength{\tabcolsep}{2.5pt}
  \resizebox{\columnwidth}{!}{%
  \begin{tabular}{llccccc}
    \toprule
    \textbf{Dataset} & \textbf{Method} & \textbf{FID}
      & \textbf{LPIPS}$_{\mathrm{full}}$
      & \textbf{LPIPS}$_\Omega$ & \textbf{DINOv2}$_\Omega$
      & \textbf{DreamSim}$_{\mathrm{ctx}}$ \\
    \midrule
    \multirow{3}{*}{Synth-Ads}
      & FLUX.2 SFT & 24.07 & 0.2103 & \textbf{0.3524} & \textbf{0.3134} & 0.0479 \\
      & GPT-Image-2 & \textbf{23.59} & \textbf{0.2083} & 0.4274 & 0.3816 & \textbf{0.0454} \\
      & Ours & \underline{23.88} & \underline{0.2094} & \underline{0.3539} & \underline{0.3154} & \underline{0.0475} \\
    \midrule
    \multirow{3}{*}{Real-Ads}
      & FLUX.2 SFT & \underline{49.25} & \underline{0.2010} & \underline{0.3754} & \underline{0.3200} & \underline{0.0660} \\
      & GPT-Image-2 & 50.19 & 0.2135 & 0.4828 & 0.4241 & 0.0677 \\
      & Ours & \textbf{47.31} & \textbf{0.1999} & \textbf{0.3686} & \textbf{0.3156} & \textbf{0.0629} \\
    \midrule
    \multirow{3}{*}{LVIS}
      & FLUX.2 SFT & \underline{16.35} & \underline{0.2146} & \underline{0.3766} & \underline{0.3283} & \underline{0.0680} \\
      & GPT-Image-2 & 16.70 & 0.2160 & 0.4799 & 0.4328 & 0.0714 \\
      & Ours & \textbf{16.08} & \textbf{0.2131} & \textbf{0.3741} & \textbf{0.3263} & \textbf{0.0659} \\
    \midrule
    \multirow{3}{*}{Open Images}
      & FLUX.2 SFT & \underline{41.80} & \underline{0.2121} & \underline{0.3763} & \underline{0.3290} & \underline{0.0722} \\
      & GPT-Image-2 & 45.45 & 0.2188 & 0.4827 & 0.4362 & 0.0757 \\
      & Ours & \textbf{40.43} & \textbf{0.2119} & \textbf{0.3740} & \textbf{0.3253} & \textbf{0.0699} \\
    \bottomrule
  \end{tabular}}
\end{table}

\AN{\ANedit{Overall, our approach reduces these structural failures and
unsupported details through subject-localized wavelet supervision. This
supervision acts as a targeted regularizer that emphasizes multiscale edges,
textures, and geometric cues in the generated subject region, discouraging
extraneous details while prioritizing identity-defining structure and coherent
subject completion. Meanwhile, the standard flow objective continues to
supervise the lower-frequency scene context, preserving overall image
quality.}}

\subsection{\ANedit{Multimodal-LLM Preference Evaluation}}
\label{sec:mllm_main}

\AN{\ANedit{Reference-based metrics may not fully capture perceptually
meaningful structural and semantic differences, while criterion-level human
evaluation over the full test sets is costly. We therefore complement them
with an anonymous pairwise evaluation by GPT-5.5 \TK{on the Real-Ads and LVIS test sets}. 
The judge compares our method with matched FLUX.2 SFT on subject completion, identity consistency,
context alignment, and outpainting \TK{realism, using} win-and-half-tie rates,
where $50\%$ denotes parity. 
As shown in \cref{tab:mllm_preference_main}, our method is
preferred on all four criteria on both \TK{datasets}
with the largest preference for context alignment on both datasets.
Full evaluation details, including a representative judgement, are provided in \cref{sec:mllm_judge}.}}

\begin{table}[t]
  \centering
  \caption{\AN{Pairwise preference evaluation.
  Win-and-half-tie rates (\%); $50\%$ denotes parity. SFT refers to
  FLUX.2 SFT.}}
  \vspace{-1em}
  \label{tab:mllm_preference_main}
  \scriptsize
  \setlength{\tabcolsep}{2pt}
  \resizebox{\columnwidth}{!}{%
  \begin{tabular}{llcccc}
    \toprule
    \textbf{Dataset} & \textbf{Method}
      & \shortstack{\textbf{Subj.}\\\textbf{completion (\%)}}
      & \shortstack{\textbf{Identity}\\\textbf{consist. (\%)}}
      & \shortstack{\textbf{Context}\\\textbf{align. (\%)}}
      & \shortstack{\textbf{Outpainting}\\\textbf{realism (\%)}} \\
    \midrule
    \multirow{2}{*}{Real-Ads}
      & Ours & \textbf{59.17} & \textbf{58.75}
      & \textbf{60.21} & \textbf{58.13} \\
      & FLUX.2 SFT & 40.83 & 41.25 & 39.79 & 41.88 \\
    \midrule
    \multirow{2}{*}{LVIS}
      & Ours & \textbf{53.73} & \textbf{53.19}
      & \textbf{54.92} & \textbf{54.00} \\
      & FLUX.2 SFT & 46.27 & 46.81 & 45.09 & 46.00 \\
    \bottomrule
  \end{tabular}}
\end{table}

\subsection{Ablation Studies}
\label{sec:ablations}

\begin{figure}[tb]
  \centering
  \scriptsize

  \makebox[\columnwidth][c]{%
    \makebox[0.2\columnwidth][c]{\textbf{Input / GT}}%
    \makebox[0.2\columnwidth][c]{\shortstack{\textbf{w/o subj.}\\
                                             \textbf{localization}}}%
    \makebox[0.2\columnwidth][c]{\textbf{FLUX.2 SFT}}%
    \makebox[0.2\columnwidth][c]{\shortstack{\textbf{w/o N.S.}\\
                                             \textbf{\& S.A.}}}%
    \makebox[0.2\columnwidth][c]{\textbf{Ours}}%
  }

  \vspace{2pt}

  \includegraphics[width=\columnwidth]{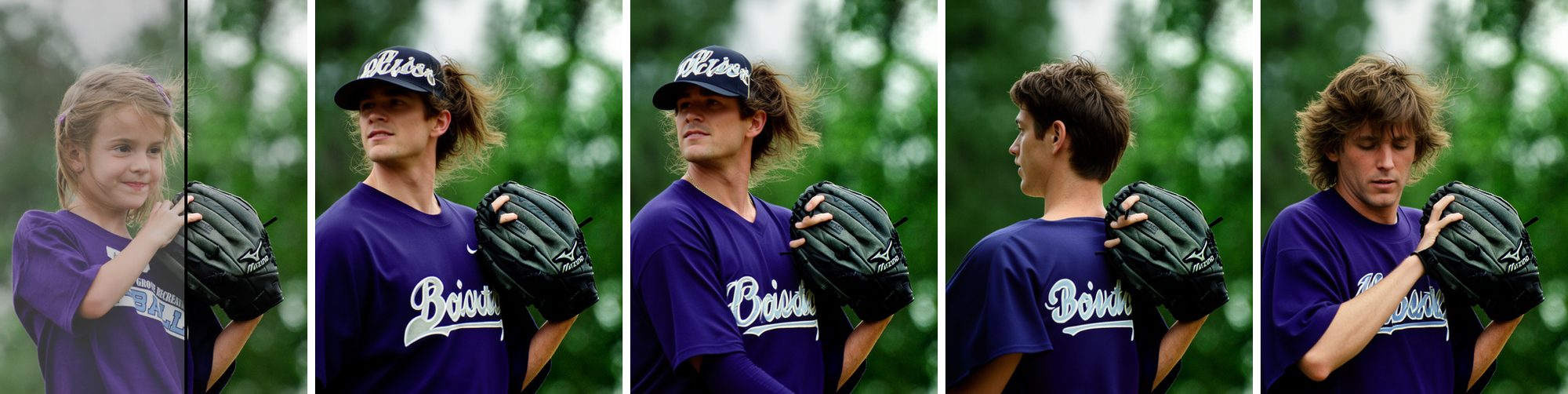}

  \caption{\AN{{Representative qualitative ablation.} \ANedit{N.S. \& S.A. denote noise
  schedule and subband adaptation, respectively.} Additional examples and analysis are
  provided in \cref{sec:qualitative_ablations}.}}

  \label{fig:qualitative_ablations}
\end{figure}

\ANedit{Among the components, VLM guidance and subject localization matter most:
relative to removing either, the full model reduces
DreamSim$_{\mathrm{ctx}}$ by $15.4\%$ and $5.1\%$, respectively. The noise
schedule and adaptive subband weights are complementary, as disabling both
degrades the localized metrics more than disabling either alone. Mask-weighted
flow contributes a smaller but consistent gain. Qualitatively
(\cref{fig:qualitative_ablations}), without high-frequency supervision the
matched SFT model often adds unsupported details that reduce subject clarity,
whereas penalizing high-frequency error across the whole outpaint region rather
than the subject alone yields less coherent, less realistic subject
completions. For additional examples and analysis, see \cref{sec:qualitative_ablations}.}


\begin{table}[tb]
  \centering
  \caption{
  \TK{Component ablation on Real-Ads. Best in \textbf{bold}; second best \underline{underlined}. Lower is better.}
  }
  \vspace{-1em}
  \label{tab:ablations}
  \scriptsize
  \setlength{\tabcolsep}{2.5pt}
  \resizebox{\columnwidth}{!}{%
  \begin{tabular}{lccccc}
    \toprule
    \textbf{Variant}
    & \multicolumn{2}{c}{\textbf{Overall}}
      & \multicolumn{3}{c}{\textbf{Subject Clarity}} \\
    \cmidrule(lr){2-3}\cmidrule(lr){4-6}
    & \textbf{FID}
      & \textbf{LPIPS}$_{\mathrm{full}}$
      & \textbf{LPIPS}$_\Omega$
      & \textbf{DINOv2}$_\Omega$
      & \textbf{DreamSim}$_{\mathrm{ctx}}$ \\
    \midrule
    w/o VLM guidance
      & 50.70 & 0.2057 & 0.3865 & 0.3326 & 0.0746 \\
    w/o mask-weighted flow
      & 48.95 & 0.2023 & \underline{0.3689}
      & \textbf{0.3169} & 0.0647 \\
    w/o subject localization
      & 50.70 & 0.2032 & 0.3767 & 0.3233 & 0.0665 \\
    w/o noise schedule
      & 49.69 & 0.2012 & 0.3701
      & \underline{0.3186} & 0.0657 \\
    w/o subband adaptation
      & \underline{48.92} & \underline{0.2011} & 0.3742
      & 0.3224 & \underline{0.0642} \\
    \makecell[l]{w/o noise schedule\\ \hspace{.5em} \& subband adaptation}
      & 49.31 & \underline{0.2011} & 0.3753
      & 0.3259 & 0.0646 \\
    \midrule
    Ours (full)
      & \textbf{48.26} & \textbf{0.2009} & \textbf{0.3662}
      & \textbf{0.3169} & \textbf{0.0631} \\
    \bottomrule
  \end{tabular}}
\end{table}

\subsection{\ANedit{Effect of Subject-HF-Aware Training}}
\label{sec:attention_main}

\begin{figure}[tb]
  \centering
  \includegraphics[width=\columnwidth]{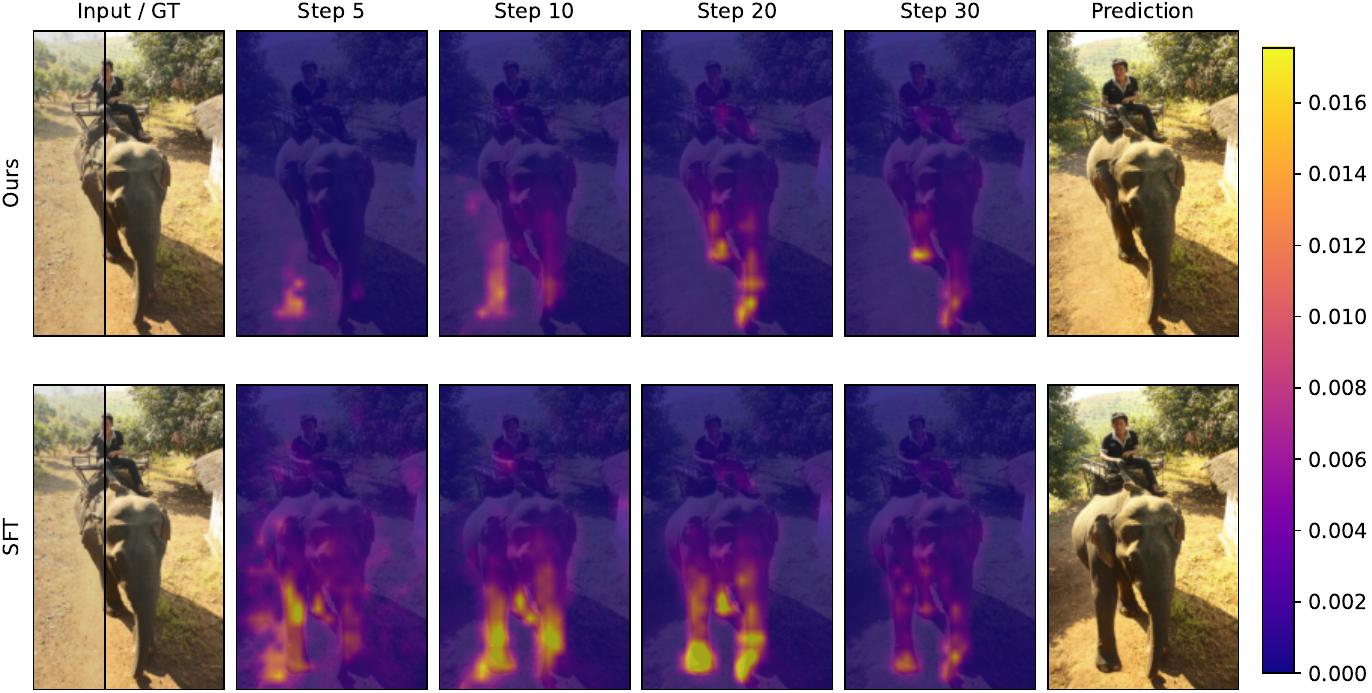}
  \caption{\AN{\textbf{Attention-routing analysis.} Direct attention from
  generated-image queries to the prompt subwords for ``foreleg.'' SFT spreads
  this cue broadly into the outpaint region, whereas ours concentrates it on
  the supported subject continuation. Maps are overlaid on dimmed predictions
  and span all image queries, whereas the reported metrics use outpaint
  queries only.
  }}
  \label{fig:attention_foreleg}
\end{figure}

\TK{
To examine how the proposed objective affects model behavior, we analyze inference-time attention in a representative case where matched SFT duplicates an elephant foreleg, while ours completes it correctly. For each generated-image query, we sum its attention over one of two conditioning key sets: visible-subject image tokens or the prompt subwords for ``foreleg.'' \ANedit{We define \emph{leakage} as the mean of these per-query sums over outpaint queries outside the ground-truth subject}, and the \emph{localization ratio} $R$ as the corresponding mean within the ground-truth subject continuation divided by leakage, with higher $R$ indicating stronger reference-aligned localization. We measure both quantities at multiple denoising steps in the final double-stream transformer block. The ground-truth subject mask is used only for this post-hoc analysis. 

With visible-subject image tokens as the key set, ours reduces leakage by $29\%$ on average and improves localization at every measured step, reaching $R=13.5$ versus $4.9$ for SFT at step $20$. We repeat the analysis using the prompt subwords corresponding to ``foreleg'' as keys (\cref{fig:attention_foreleg}). Here, ours reduces leakage by $77\%$ and reaches $R=5.4$ versus $0.9$ for SFT at step $20$. These results suggest that our model concentrates both visual and part-specific textual attention more selectively within the supported subject continuation. Because attention weights omit value vectors and subsequent nonlinear computation, we treat this as a mechanistic, rather than causal, analysis; step-level results and additional probes are provided in \cref{sec:attention_case_study}.}

\section{Conclusion}
\label{sec:conclusion}


\AN{
\TK{
We introduced subject-clarity outpainting as a subject-intersecting setting that prioritizes subject fidelity alongside global completion quality. We addressed this problem through subject-centric data curation 
}
and a lightweight wavelet objective that adapts supervision across frequency bands
and noise levels. The resulting model consistently outperforms its matched SFT
baseline across four datasets without \TK{modifying} inference. More broadly, our
findings suggest that global completion quality and subject preservation
should be evaluated separately when adapting generative models to
fidelity-sensitive applications. 
\ANedit{Extending the framework beyond 
\TK{FLUX 2 and addressing residual failures involving} 
overlapping partially visible subjects and truncated text remain \TK{directions} for future work.}
}

\clearpage
{
    \small
    \bibliographystyle{ieeenat_fullname}
    \bibliography{main}
}
\clearpage
\appendix
\section{Dataset Details}
\label{sec:dataset_details}

\AN{This appendix provides additional provenance and statistics for the four
datasets. Synth-Ads and Real-Ads are maskless
advertising-image collections for which subject masks are inferred by
Sa2VA~\cite{yuan2025sa2va}. \ANedit{Synth-Ads consists of 5,659 images
regenerated by GPT-Image-2 from real advertising sources to obtain clean,
coherent backgrounds around the original product, while Real-Ads consists of
1,356 genuine advertising images that enter the pipeline without synthesis;
\cref{sec:appendix_data_curation} details their preparation and
\cref{tab:filtering_summary} reports their filtering statistics.} For the instance-masked sources, we use images from
the LVIS v1 validation split~\cite{gupta2019lvis} and a class-balanced
subset of the Open Images V7 training split~\cite{kuznetsova2020open}.
The retained processing logs account for 19,808 LVIS inputs and 4,650 Open
Images inputs. The Open Images subset is sampled from the 350 classes for which
segmentation annotations are available. We draw from its training partition to
obtain broader object and scene coverage, and cap sampling at 15 images per
class to promote class diversity rather than allowing frequent categories to
dominate. Both collections are repartitioned into our train/validation/test
splits after filtering.}
\\

\subsection{Subject-Centric Data Curation}
\label{sec:appendix_data_curation}

\AN{We assemble training data from four sources spanning two regimes: two
maskless advertising-image collections, Synth-Ads and Real-Ads, for
which no ground-truth segmentation is available, and two
instance-masked detection datasets, LVIS and Open Images, which provide
instance segmentation maps and labels. Synth-Ads is synthesized from real
advertising images using GPT-Image-2, whereas Real-Ads contains real
advertisements. Both regimes follow a common four-stage pipeline - raw
screening, subject-mask generation, mask-dependent filtering, and pair
generation - with segmentation replaced by ground-truth instances when
available. Unless otherwise noted, thresholds are shared across datasets.}
\\

\noindent\textbf{Synth-Ads generation.}
\AN{Starting from a large collection of real advertising images, we use the
GPT-5.5 vision-language model to produce a per-image background-editing
prompt, a background-complexity label, and a flag indicating whether the image
contains marketing-text overlays. Images with complex backgrounds are
discarded because they cannot be regenerated coherently. For each remaining
image, GPT-Image-2 regenerates a clean, coherent scene around the advertised
product while being instructed to preserve the foreground subject and any
marketing text. Real-Ads instead contains real advertising images and enters
the shared pipeline without synthesis.}
\\

\noindent\textbf{Subject clarity curation pipeline.}
\AN{All four collections enter the shared pipeline below. Its purpose is to retain images with a
clearly identifiable, sufficiently detailed foreground subject and construct
supervised pairs in which outpainting interacts meaningfully with that
subject.}

\begin{enumerate}
  \item \AN{\textbf{Raw image screening.} We cap all images at $512$~px and
    discard inputs with either dimension below $384$~px. For Synth-Ads and
    Real-Ads, GPT-5.5 additionally rejects images with no clear subject,
    \ANedit{pure-ambience scenes (generic landscapes, abstract patterns, or
    stock-photo shots with no defensible advertised subject)}, and unsupported
    plain backgrounds. LVIS and Open Images skip semantic screening because
    their instance annotations already define candidate subjects.}

  \item \AN{\textbf{Subject-mask generation.} For maskless sources (Synth-Ads and Real-Ads), we segment
    the VLM-identified primary subject with Sa2VA~\cite{yuan2025sa2va} and
    discard images for which no mask is produced. For instance-masked sources,
    we form the subject mask by taking the union of ground-truth instances
    whose individual area fraction is at least $0.05$, discarding images with
    no qualifying instance.}

  \item \AN{\textbf{Mask-dependent filtering.} \ANedit{We retain images only
    when the subject is sufficiently large (subject area $\geq0.05$) and
    visually structured. We compute edge density as the fraction of edge pixels
    within the subject mask and require it to exceed $0.05$, removing smooth or
    textureless subjects. For instance-masked data, we additionally measure
    edge density over the non-subject background region and require it to
    exceed $0.02$, excluding plain-background examples with little contextual
    structure.} We do not apply the external-edge criterion to Synth-Ads or
    Real-Ads because their semantic screening already removes unsuitable
    ambience scenes and unsupported plain-background cases.}

  \item \AN{\textbf{Subject-adaptive pair generation.} Retained images are
    assigned to train, validation, and test splits in a $70/5/25$ ratio. We
    generate each outpainting pair by sampling a crop direction from the four
    canvas sides and a crop fraction conditioned on subject area, ranging from
    $[0.10,0.15]$ for small subjects to $[0.35,0.50]$ for large subjects.
    This avoids excessive removal of small subjects while retaining
    challenging crops. Each sample contains the cropped input, outpaint mask,
    ground-truth target, and subject mask.}
\end{enumerate}

\noindent\textbf{Filtering statistics.} \AN{\Cref{tab:filtering_summary} summarizes the high-level curation
statistics. ``Input'' denotes images entering the subject clarity curation
pipeline and ``kept'' denotes finalized outpainting pairs. For Synth-Ads, the
input count corresponds to successfully generated images entering screening;
we do not infer the size of the original advertising collection from its
dataset label.}
\\

\begin{table}[hbpt]
  \centering
  \caption{\AN{High-level filtering statistics. Yield is the percentage of
  input images retained as finalized pairs.}}
  \label{tab:filtering_summary}
  \footnotesize
  \setlength{\tabcolsep}{5pt}
  \begin{tabular}{lrrr}
    \toprule
    \textbf{Dataset} & \textbf{Input} & \textbf{Kept} &
      \textbf{Yield} \\
    \midrule
    Synth-Ads   & 5,659   & 3,707 & 65.5\% \\
    Real-Ads    & 1,356   & 969   & 71.5\% \\
    LVIS        & 19,808  & 6,705 & 33.8\% \\
    Open Images & 4,650   & 1,267 & 27.2\% \\
    \bottomrule
  \end{tabular}
\end{table}


\noindent\textbf{Dataset splits.} \ANedit{After filtering, each source's retained
samples are partitioned into a nominal $70/5/25$ train/validation/test split,
with realized sizes reported in \cref{tab:dataset_splits}. This allocation
preserves most samples for training while retaining a substantial held-out test
set for each source. Near-duplicate images are kept within a single partition,
and Synth-Ads and Real-Ads are checked for cross-collection overlap before
splitting, preventing source-level leakage between training and evaluation.}
\\

\begin{table}[t]
  \centering
  \caption{\AN{Realized train/validation/test split sizes after filtering.}}
  \label{tab:dataset_splits}
  \footnotesize
  \setlength{\tabcolsep}{5pt}
  \begin{tabular}{lrrrr}
    \toprule
    \textbf{Dataset} & \textbf{Train} & \textbf{Val.} &
      \textbf{Test} & \textbf{Total} \\
    \midrule
    Synth-Ads   & 2,595 & 178 & 934   & 3,707 \\
    Real-Ads    & 681   & 48  & 240   & 969 \\
    LVIS        & 4,677 & 377 & 1,651 & 6,705 \\
    Open Images & 904   & 65  & 298   & 1,267 \\
    \midrule
    Total       & 8,857 & 668 & 3,123 & 12,648 \\
    \bottomrule
  \end{tabular}
\end{table}
\noindent\textbf{Subject-adaptive cropping.} \ANedit{The sampled crop fraction is conditioned on the fraction of image pixels occupied by the subject, following the policy in \cref{tab:crop_policy}: larger subjects permit a more aggressive crop, while smaller subjects are cropped conservatively to avoid removing them almost entirely. After sampling a fraction, we choose one of the four canvas sides uniformly and place the cut according to cumulative subject-mask mass along that axis.}
\\

\begin{table}[t]
  \centering
  \caption{\AN{Subject-area-dependent crop-fraction sampling policy.}}
  \label{tab:crop_policy}
  \footnotesize
  \setlength{\tabcolsep}{8pt}
  \begin{tabular}{cc}
    \toprule
    \textbf{Subject area fraction} & \textbf{Crop-fraction range} \\
    \midrule
    $[0.05, 0.15)$ & $[0.10, 0.15]$ \\
    $[0.15, 0.30)$ & $[0.15, 0.25]$ \\
    $[0.30, 0.50)$ & $[0.25, 0.35]$ \\
    $[0.50, 1.00]$ & $[0.35, 0.50]$ \\
    \bottomrule
  \end{tabular}
\end{table}

\begin{table}[t]
  \centering
  \caption{\AN{Distribution (\%) of samples by realized masked fraction of the
  full image. Rows may differ from 100\% by rounding.}}
  \label{tab:mask_distribution}
  \scriptsize
  \setlength{\tabcolsep}{2.5pt}
  \resizebox{\columnwidth}{!}{%
  \begin{tabular}{lrrrrrr}
    \toprule
    \textbf{Dataset} & \textbf{$<10\%$} & \textbf{10--20\%} &
      \textbf{20--30\%} & \textbf{30--40\%} & \textbf{40--50\%} &
      \textbf{$\geq50\%$} \\
    \midrule
    Synth-Ads   & 0.6 & 5.7  & 21.4 & 37.7 & 24.8 & 9.7 \\
    Real-Ads    & 4.9 & 11.2 & 18.2 & 28.4 & 20.6 & 16.7 \\
    LVIS        & 6.8 & 12.2 & 18.1 & 24.0 & 20.2 & 18.8 \\
    Open Images & 3.9 & 8.8  & 18.5 & 27.5 & 24.2 & 17.0 \\
    \bottomrule
  \end{tabular}}
\end{table}

\noindent\textbf{Realized mask distribution.} \Cref{tab:mask_distribution} reports the realized percentage of samples in
10-point bins of full-canvas mask coverage, measured from the final preprocessed
binary masks. Although crop fractions are sampled from
\cref{tab:crop_policy}, mask coverage also depends on the subject's location
and spatial extent. Consequently, the final data span a broad range of
outpainting difficulty rather than concentrating at the sampled crop bounds.

\section{Implementation and Evaluation Details}
\label{sec:implementation_details}

\subsection{Training Details}

\ANedit{We train with AdamW at a learning rate of $1\times10^{-4}$ using a
cosine schedule with $5\%$ warmup, an effective batch size of $4$, \texttt{bf16}
precision, and gradient checkpointing. A fixed gray-fill task prefix is
prepended to each VLM-generated scene-guidance prompt.}

\subsection{VLM Prompt Structure}
\label{sec:vlm_prompt_structure}

For each sample, GPT-5.5 receives only the masked input and the side on
which the image is cropped; it does not observe the ground-truth completion.
The VLM identifies whether the subject reaches the outpaint boundary, names
the subject, describes a conservative continuation, and optionally recovers
high-confidence partial text. \Cref{fig:vlm_output_schema} shows the returned
structure.

\begin{figure}[t]
  \centering
  \fbox{\begin{minipage}{0.92\columnwidth}
    \ttfamily\scriptsize
    \{\\
    \quad ``subject\_at\_boundary'': \textless true $\mid$ false\textgreater,\\
    \quad ``subject'': ``\textless short subject phrase\textgreater'',\\
    \quad ``direction'': ``\textless left $\mid$ right $\mid$ top $\mid$ bottom\textgreater'',\\
    \quad ``extension\_detail'': ``\textless plausible continuation\textgreater'',\\
    \quad ``fill\_instruction'': ``\textless short outpainting instruction\textgreater'',\\
    \quad ``text\_completion'': \{\\
    \qquad ``has\_partial\_text'': \textless true $\mid$ false\textgreater,\\
    \qquad ``visible\_fragments'': [``\textless fragment\textgreater''],\\
    \qquad ``predicted\_completions'': [``\textless completion\textgreater''],\\
    \qquad ``confidence'': ``\textless high $\mid$ medium $\mid$ low\textgreater''\\
    \quad \}\\
    \}
  \end{minipage}}
  \caption{VLM output structure. The schema records atomic
  subject-continuation information inferred from the masked image. Partial
  text is used only when its completion confidence is high.}
  \label{fig:vlm_output_schema}
\end{figure}

\begin{table*}[t]
  \centering
  \caption{\AN{Prompts used for the predictions in the main paper qualitative
  comparison. Labels (a)--(d) follow the figure rows.
  Prompt strings are reproduced as used at inference.}}
  \label{tab:qualitative_prompts}
  \scriptsize
  \setlength{\tabcolsep}{5pt}
  \begin{tabular}{p{0.15\textwidth}p{0.79\textwidth}}
    \toprule
    \textbf{Baseline} & \textbf{Prompt(s)} \\
    \midrule
    PowerPaint
      & All rows use the native image-outpainting context token with an empty
        free-form prompt. \\
    \addlinespace
    FLUX.1 Fill
      & \textbf{(a)} \texttt{dog wearing a black coat, with the dog's
        upper body, back, neck, and head continuing from the visible coat.}
        \par
        \textbf{(b)} \texttt{antique handwritten manuscript page and stacked
        books, with manuscript page with cursive writing and leather-bound
        books continuing from the visible boundary.}
        \par
        \textbf{(c)} \texttt{seaplane, with the seaplane's upper
        fuselage, cockpit windows, wings, tail, and propeller continuing above
        the floats.}
        \par
        \textbf{(d)} \texttt{giraffe, with the cropped giraffe's upper
        neck and head, with surrounding dry savanna brush continuing naturally.}
        \par
        \\
    \addlinespace
    fal/FLUX.2 LoRA
      & All rows use the checkpoint-specific instruction
        \texttt{Fill the green spaces according to the image.} \\
    \addlinespace
    FLUX.2 SFT
      & \textbf{(a)} \texttt{Fill the gray region by extending dog wearing a
        black coat, with the dog's upper body, back, neck, and head
        continuing from the visible coat.}
        \par
        \textbf{(b)} \texttt{Fill the gray region by extending antique
        handwritten manuscript page and stacked books, with manuscript page
        with cursive writing and leather-bound books continuing from the
        visible boundary.}
        \par
        \textbf{(c)} \texttt{Fill the gray region by extending seaplane, with
        the seaplane's upper fuselage, cockpit windows, wings, tail, and
        propeller continuing above the floats.}
        \par
        \textbf{(d)} \texttt{Fill the gray region by extending giraffe, with
        the cropped giraffe's upper neck and head, with surrounding dry
        savanna brush continuing naturally.}
        \par
        \\
    \addlinespace
    Ours
      & Same as FLUX.2 SFT. \\
    \bottomrule
  \end{tabular}
\end{table*}

\subsection{Baselines}

\AN{Our baseline set is designed to cover complementary approaches rather than
multiple variants of a single architecture. All methods use identical visible
crops, binary outpaint masks, and target resolutions. Each sample is then
converted using the baseline's official preprocessing and method-specific
masked-canvas convention, including its recommended mask dilation or boundary
handling where applicable; we do not impose our gray-filled conditioning
representation on the baselines. We use official
implementations and released checkpoints with their method-specific inference
settings. Where free-form text conditioning is supported, prompts are derived
from the same per-sample VLM output used by our method, but are reformatted to
match the baseline's pretrained task and native interface rather than forcing
an identical string across incompatible backbones. We evaluate all methods
using their native decoded outputs, without post-hoc pixel compositing or
mask-based replacement of the visible region. Although compositing can enforce
exact preservation of known pixels, it can also conceal preservation and
boundary failures. Hard compositing may introduce visible seams, motivating
Gaussian blending or related boundary processing; however, such blending
introduces tunable post-processing hyperparameters that may affect methods
differently and confound architectural comparisons. Because latent generative
models decode outputs over the complete image canvas, nominally unmasked pixels
may differ from the input due to latent-space generation and VAE
reconstruction.}

\noindent\textbf{BrushNet.}
\AN{BrushNet~\cite{ju2024brushnet} augments a frozen diffusion inpainting
backbone with a dual-branch conditioning network. We use the released
random-mask BrushNet checkpoint with Stable Diffusion XL, its DPM-Solver
scheduler, and 50 inference steps. Because SDXL and BrushNet are trained with
descriptive image captions rather than editing commands, we convert the VLM
output into a caption such as
\texttt{a photo of a \{subject\}, \{extension\_detail\}}.}

\noindent\textbf{PowerPaint.}
\AN{PowerPaint~\cite{zhuang2024task} is a task-conditioned Stable Diffusion
1.5 inpainting model that uses learned task tokens to support object removal,
shape-guided generation, and outpainting. We use the released
\texttt{ppt-v2-1} checkpoint in its native image-outpainting mode with 45 DDIM
steps and guidance scale 10, while disabling its internal canvas expansion so
that the benchmark input and mask are preserved. This mode uses PowerPaint's
context task token and an empty free-form prompt, rather than imposing an
instruction format outside its outpainting training interface. In preliminary
prompt sweeps, subject-only and progressively detailed natural-language prompts
were less reliable than this native configuration. A plausible explanation is
PowerPaint's task-prompt design: the context token $P_{\mathrm{ctxt}}$ is
specialized for context-consistent completion, whereas text-guided prompting
invokes the object token $P_{\mathrm{obj}}$ and can shift generation toward
semantic object synthesis. We therefore retain the native
$P_{\mathrm{ctxt}}$ mode, which gave the strongest preliminary results.}

\noindent\textbf{VIP.}
\AN{VIP~\cite{yang2024vip} is a Stable-Diffusion-based outpainting framework
that uses a multimodal language model and a Center--Total--Surrounding control
mechanism to separate visible-image content from the requested extension.
\ANedit{Since VIP does not provide a released outpainting checkpoint, we train
it on top of its Stable Diffusion backbone using the authors' default training
configuration and the same pooled training split as our model.} At inference,
we use its native pipeline, preserving each benchmark canvas and mask. Following the official inference configuration, we
use guidance scale 7.5 and seed 42, and populate VIP's
\texttt{Center: \{visible content\}; Surrounding: \{extension\}} prompt
structure from the same per-sample VLM description used by the other
text-conditioned methods.}

\noindent\textbf{FLUX.1 Fill.}
\AN{FLUX.1 Fill~\cite{flux2024} is the dedicated inpainting and outpainting
variant of FLUX.1. We evaluate the released \texttt{FLUX.1-Fill-dev} model
with 50 inference steps and guidance scale 30. Its prompt is a
FLUX-compatible scene caption because the model is pretrained with descriptive
text conditioning rather than explicit editing commands. We construct this
caption from the shared VLM fields using
\texttt{\{subject\}, with \{extension\_detail\}}, followed by a predicted
completion of visible text when available with high confidence.}

\noindent\textbf{DING.}
\AN{DING~\cite{moufad2026efficient} is a training-free method that modifies
the denoising trajectory to enforce consistency with the observed image. We
retain its released sampler update and 26-step schedule, while adapting only
the model interface from FLUX.1 to the FLUX.2 Klein Base 4B backbone used in
our benchmark. It receives the same instruction-style VLM information as our
model, reconstructed as
\texttt{Fill the gray region by extending \{subject\} to the \{direction\},
with \{extension\_detail\}}. A predicted text completion is appended only
when the VLM assigns high confidence.}

\noindent\textbf{FlowChef.}
\AN{FlowChef~\cite{patel2024steering} performs training-free steering of
rectified-flow trajectories toward image constraints. We use the same FLUX.2
Klein Base 4B interface, guidance configuration, and per-sample prompts as for
DING, including the same fill-command template and optional high-confidence
text completion, while retaining FlowChef's released update rule and
recommended 51-step schedule. Thus, the comparison isolates the sampling
strategy rather than semantic conditioning or backbone differences.}

\noindent\textbf{fal/FLUX.2 LoRA.}
\AN{The public fal outpainting LoRA~\cite{fal2024outpaint} provides an
adaptation-based comparison on FLUX.2 Klein 4B. We use its released weights
and prescribed four-step inference recipe, including LoRA scale 1.1, the green
masked-canvas representation, and its fixed instruction,
\texttt{Fill the green spaces according to the image}. We retain this
checkpoint-specific prompt instead of substituting our VLM instruction because
the adapter was trained for that interface.}

\noindent\textbf{GPT-Image-2.}
\AN{We evaluate GPT-Image-2~\cite{openai2026gptimage2} through its image-editing
interface at low quality. The input is the same benchmark canvas with the fill
region replaced by gray. We utilize the same VLM prompt structure as our model, and
select the supported output aspect ratio nearest to the benchmark canvas.
\ANedit{Unlike the open baselines, the API returns only a full edited canvas
at its supported resolutions and offers no mask-constrained output. Evaluating
the raw resized output would penalize the model for whole-canvas resampling
drift in the visible region---an artifact of the delivery pipeline rather
than of outpainting quality. We therefore resize the result to the target
resolution and restore the original visible pixels outside the mask,
following common practice for evaluating API-based editors on a fixed
benchmark canvas. This restoration guarantees that the visible region matches
the target, so it can only favor GPT-Image-2 on metrics that include visible
content; the localized subject-region metrics are unaffected.} Because
outputs are available for a subset of the benchmark, GPT-Image-2, matched
SFT, and ours are re-evaluated on the exact same per-dataset intersection.}

\subsection{Computational Cost Analysis}

\AN{\ANedit{Both the matched SFT model and ours use the same frozen FLUX.2
Klein Base 4B backbone and rank-32 LoRA parameterization (39.977M trainable
parameters); the wavelet objective adds no learned parameters. During
training, ours additionally decodes the pseudo-clean prediction and applies a
fixed three-level Haar DWT with masked, per-band weighting, adding
$\mathcal{O}(CHW)$ work per step - for $C$ channels and spatial size
$H{\times}W$ - beyond the matched-SFT cost $X$; this fixed transform requires no optimizer state and
is confined to fine-tuning. At deployment, the pseudo-clean decoder path,
subject masks, DWT, subband weights, and noise-dependent gates are all
removed, so ours has identical transformer evaluations, denoising steps,
adapter storage, and inference cost to matched SFT - consistent with the
13.32\,s/image measured for both in \cref{tab:complexity}.}}

\begin{table}[t]
  \centering
  \caption{\AN{Computational cost relative to matched SFT. Complexity denotes
  per-step training work. Time/image is the sample-weighted wall-clock mean
  over the test set at 30 denoising steps.}}
  \label{tab:complexity}
  \footnotesize
  \setlength{\tabcolsep}{4pt}
  \resizebox{\columnwidth}{!}{%
  \begin{tabular}{lccc}
    \toprule
    \textbf{Method} & \textbf{Trainable params.} &
      \textbf{Train complexity} & \textbf{Time/image} \\
    \midrule
    FLUX.2 SFT & 39.977M & $X$                   & 13.32 s \\
    Ours       & 39.977M & $X+\mathcal{O}(CHW)$ & 13.32 s \\
    \bottomrule
  \end{tabular}}
\end{table}

\subsection{Metric Definitions}

\AN{Overall metrics and subject clarity metrics answer different questions.
FID~\cite{heusel2017gans} compares the distribution of generated and target
images and captures global realism, while full-image
LPIPS~\cite{zhang2018unreasonable} measures paired perceptual fidelity. For generated
features $(\mu_g,\Sigma_g)$ and target features $(\mu_r,\Sigma_r)$, we use the
standard definition
\begin{equation}
  \mathrm{FID}
  = \lVert\mu_g-\mu_r\rVert_2^2
  + \operatorname{Tr}\!\left(
      \Sigma_g+\Sigma_r-2(\Sigma_g\Sigma_r)^{1/2}
    \right).
\end{equation}
For a prediction $\hat{I}$ and target $I^\star$, full-image LPIPS is
$d_{\mathrm{LPIPS}}(\hat{I},I^\star)$. However, most visible pixels are copied
from the input and can dominate a full-image score. We therefore define
$\Omega=M_{\mathrm{outpaint}}\cap M_{\mathrm{subject}}$. LPIPS$_\Omega$ and
DINOv2$_\Omega$ evaluate this generated subject region directly, while
DreamSim$_{\mathrm{ctx}}$ complements them with a contextual assessment of
the complete subject.}

\AN{\textbf{Localized LPIPS~\cite{zhang2018unreasonable}.} LPIPS provides a spatial
perceptual-distance map $\ell_{\mathrm{LPIPS}}(\hat{I},I^\star)$. We process
the full prediction and target so that its receptive fields retain image
context, resize $\Omega$ to the output-map resolution to obtain $\Omega_L$,
and compute
\begin{equation}
  \mathrm{LPIPS}_{\Omega}
  = \frac{1}{|\Omega_L|}
    \sum_{u\in\Omega_L}
    \ell_{\mathrm{LPIPS}}(\hat{I},I^\star)_u .
\end{equation}
This preserves LPIPS's learned perceptual representation without allowing
unchanged pixels to dominate the score.}

\AN{\textbf{Localized DINOv2~\cite{oquab2023dinov2}.} DINOv2 provides spatial
patch features that capture object appearance beyond pixel-level agreement.
We apply identical subject-context crops to the prediction and target. Let
$\mathcal{P}_{\Omega}$ denote spatially aligned patch tokens overlapping
$\Omega$, with prediction and target features $\hat{z}_i$ and $z_i^\star$.
The localized distance is
\begin{equation}
  \mathrm{DINOv2}_{\Omega}
  = \frac{1}{|\mathcal{P}_{\Omega}|}
    \sum_{i\in\mathcal{P}_{\Omega}}
    \left(
      1-\frac{\hat{z}_i^\top z_i^\star}
      {\lVert\hat{z}_i\rVert_2\lVert z_i^\star\rVert_2}
    \right).
\end{equation}
Evaluating the entire subject instead would dilute continuation errors with
features from the unchanged visible portion. DINOv2$_\Omega$ therefore
remains focused on semantic and structural fidelity where new subject content
is generated.}

\AN{\textbf{Contextual DreamSim~\cite{fu2023dreamsim}.} DreamSim produces a
single distance for an entire image rather than spatial tokens or a distance
map. Let $\mathcal{C}(\cdot)$ denote the matched subject-context crop. We
report
\begin{equation}
  \mathrm{DreamSim}_{\mathrm{ctx}}
  = d_{\mathrm{DreamSim}}\!\left(
      \mathcal{C}(\hat{I}),\mathcal{C}(I^\star)
    \right).
\end{equation}
Directly masking the image would introduce artificial boundaries and large
constant regions that change its embedding, causing the metric to partly
measure the masking artifacts. A tight crop around $\Omega$ would avoid those
artifacts but can remove the visible portion needed to judge whether the
generated continuation connects coherently to the original subject. We
therefore evaluate matched, unmasked subject-context crops.
DreamSim$_{\mathrm{ctx}}$ consequently measures holistic completion coherence
and semantic plausibility rather than serving as a strictly localized
outpaint-region metric.}

\section{Additional Results}
\label{sec:additional_results}

\subsection{Metric Behavior on Qualitative Examples}
\label{sec:metric_behavior}

\begin{figure*}[t]
  \centering
  \includegraphics[width=0.90\textwidth]{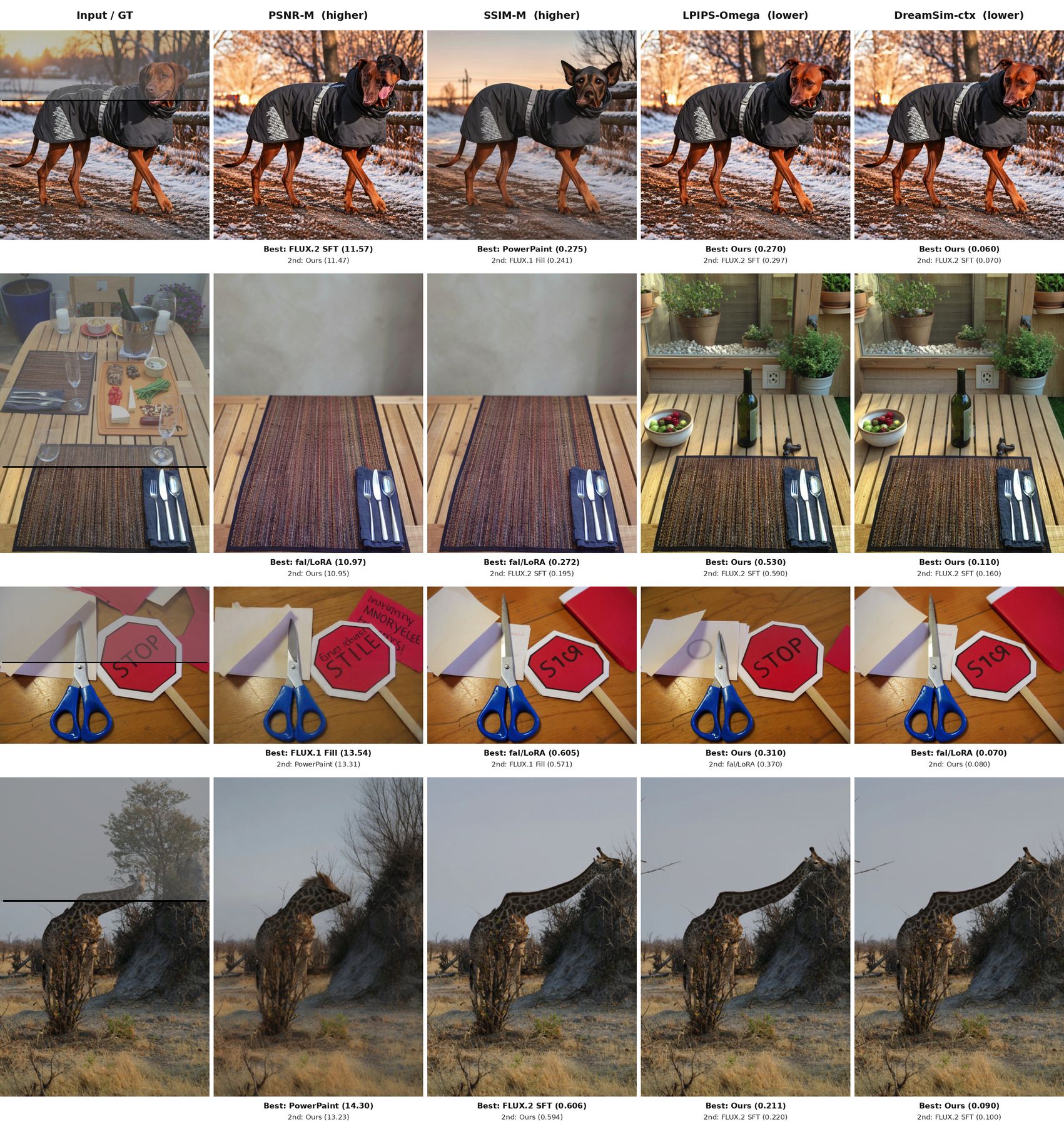}
  \caption{\AN{{Metric-selection behavior.} Each column shows the
  prediction ranked best by the indicated metric among the methods; the best and second-best methods and scores appear
  below the image to expose the selection margin. Higher is better
  for masked PSNR and SSIM, whereas lower is better for LPIPS$_\Omega$ and
  DreamSim$_{\mathrm{ctx}}$.}}
  \label{fig:metric_behavior}
\end{figure*}

\ANedit{Beyond the metrics used in the main paper, we perform a diagnostic
study here using common pixel-level metrics such as PSNR and SSIM, which
reward per-pixel color and structural agreement with the ground truth but do
not account for semantic correctness, to illustrate why such agreement can be
misleading for subject clarity outpainting.} \Cref{fig:metric_behavior} examines four
examples for which pixel-level and subject clarity metrics select different predictions. For each metric, we show
the prediction receiving its best score among the compared methods. PSNR and
SSIM frequently favor smooth, color-matched, or locally similar completions
despite incorrect subject identity, structure, or text. LPIPS$_\Omega$ selects
our prediction in all four cases, while DreamSim$_{\mathrm{ctx}}$ selects ours
in three and fal/FLUX.2 LoRA in the text-completion example. This diagnostic
does not imply that any single perceptual metric perfectly captures visual
quality; rather, it illustrates why pixel-level agreement alone is
insufficient for subject clarity outpainting and motivates our complementary
localized and contextual evaluation.

\subsection{Multimodal-LLM Preference Evaluation}
\label{sec:mllm_judge}

\AN{The reference-based metrics above provide reproducible measurements of
subject fidelity, but the localized metrics depend on subject masks and their
numerical differences can be small even when the predictions exhibit
perceptually meaningful structural or semantic differences. Collecting
criterion-level human judgments over the full test sets is also costly. We
therefore complement the reference-based evaluation with a scalable anonymous
pairwise assessment by GPT-5.5. This evaluation is intended as an additional
semantically grounded diagnostic rather than a replacement for human
evaluation.}

\begin{figure*}[t]
  \centering
  \setlength{\tabcolsep}{2pt}
  \begin{tabular}{@{}cccc@{}}
    \textbf{Masked input} & \textbf{Ground truth} &
    \textbf{Method A (FLUX.2 SFT)} & \textbf{Method B (Ours)} \\[-1pt]
    \includegraphics[width=0.238\textwidth]{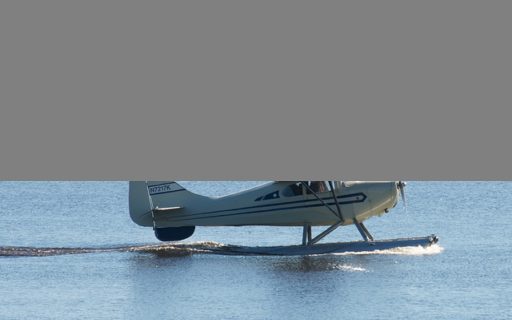} &
    \includegraphics[width=0.238\textwidth]{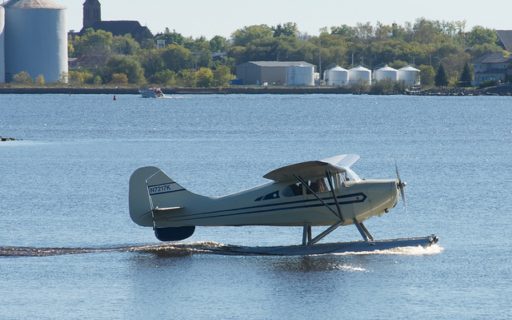} &
    \includegraphics[width=0.238\textwidth]{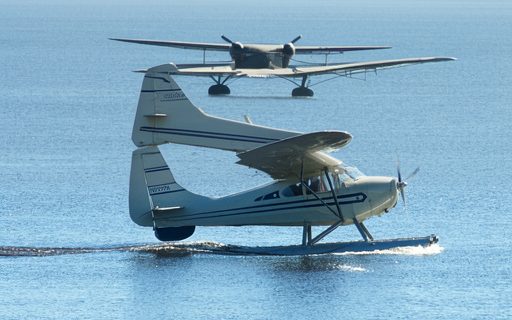} &
    \includegraphics[width=0.238\textwidth]{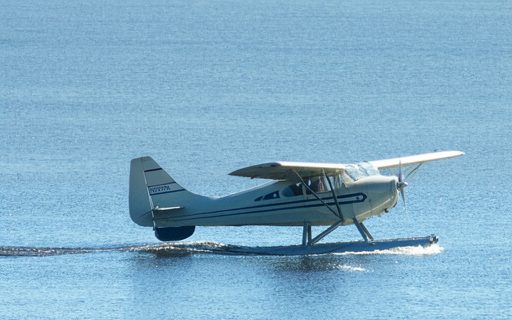}
  \end{tabular}
  \par\medskip
  \fbox{\begin{minipage}{0.96\textwidth}
    \scriptsize
    \textbf{Judge summary:}
    Method B preserves a single plausible seaplane, whereas Method A creates
    duplicated aircraft structures and inconsistent geometry in the
    outpainted region.
    \par\smallskip
    \textbf{Subject completion:}
    Method B plausibly completes the upper wing and cabin; Method A invents an
    additional malformed aircraft above the original.
    \quad
    \textbf{Identity consistency:}
    Method B maintains a single floatplane, whereas Method A changes it into a
    confusing multi-aircraft composition with duplicated parts.
    \par\smallskip
    \textbf{Context alignment:}
    Method B extends the surrounding water and sky consistently; the
    overlapping aircraft structures in Method A are semantically and
    geometrically unsupported.
    \quad
    \textbf{Outpainting realism:}
    Method B is clean and realistic, while Method A contains distorted wings,
    duplicated plane parts, and unnatural intersections.
  \end{minipage}}
  \caption{\AN{Example MLLM judgment on LVIS. Ours is
  preferred over matched SFT for all four reported criteria. In the anonymous
  evaluation, FLUX.2 SFT was presented as Method A and ours as Method B; the
  model identities in parentheses are shown only for the reader. The judge saw
  the masked input, binary mask, and the two predictions, but not the
  ground-truth completion. The text below the images condenses the
  criterion-level rationales returned by GPT-5.5.}}
  \label{fig:mllm_judge_example}
\end{figure*}

\AN{For each sample, the judge receives the masked input, binary outpaint mask,
and the predictions from our model and matched FLUX.2 SFT, labeled only as
Method A and Method B; it does not receive the ground-truth completion or the
method identities. Candidate order is assigned deterministically from a hash
of the sample and method identifiers, producing an approximately balanced A/B
assignment and mitigating positional bias. The judge independently selects A,
B, or a tie for \emph{subject completion}, \emph{subject identity
consistency}, \emph{context alignment}, and \emph{outpainting realism}. For
each criterion, the preference score assigns one point to a win and half a
point to a tie, so $50\%$ denotes parity. We evaluate all 240 Real-Ads test samples and 1,648 valid LVIS comparisons; three additional LVIS samples are
excluded by the API content-safety filter. \ANedit{\cref{fig:mllm_judge_example}
shows a representative judgment under this protocol.}}

\AN{\noindent\textbf{Evaluation criteria.}
\emph{Subject completion} assesses whether missing or cropped subject parts are
completed plausibly. \emph{Subject identity consistency} assesses whether the
completion preserves the subject's identity and defining attributes.
\emph{Context alignment} assesses semantic and geometric consistency with the
original image. \emph{Outpainting realism} assesses whether the generated
region is visually realistic and free of artifacts.}
\ANedit{Original-region preservation was also queried as an auxiliary
diagnostic but is not reported because both methods use the same latent-blending inference procedure, and $96.3\%$ and $97.5\%$ of judgments were ties on
Real-Ads and LVIS, respectively.}
The system and user prompts provided to the GPT-5.5 judge
are shown in \cref{fig:mllm_judge_prompt}.

\begin{figure*}[t]
\centering
\fbox{\begin{minipage}{0.97\textwidth}
\fontsize{6.5}{7.5}\selectfont
\ANedit{\textbf{System prompt.}\par
You are an expert evaluator comparing two image outpainting methods
head-to-head.\par
\smallskip
Your goal is to decide, for each criterion, which method (A or B) produced a
better outpaint. Small visible differences count. Look carefully at the
outpainted region, especially: distorted fingers or hands, malformed faces,
warped text/logos, seam artifacts, unrealistic object geometry, subject
continuation, boundary blending.\par
\smallskip
Be strict and consistent. Return only valid JSON. No markdown.}
\par\medskip
\ANedit{\textbf{User prompt.}\par
You will be given:\par
1. Original input image (with masked region visible as gray/black
placeholder).\par
2. Method A outpainted output.\par
3. Method B outpainted output.\par
4. Binary mask showing the outpainted region.\par
\smallskip
For each criterion, output:\par
\quad winner : ``A'' \textbar{} ``B'' \textbar{} ``tie''\par
\quad margin : integer 0--3\par
\qquad 0 = tie / imperceptible\par
\qquad 1 = slight preference\par
\qquad 2 = clear preference\par
\qquad 3 = large preference\par
\quad reason : short explanation, mention specific artifacts if any}
\par\smallskip
\ANedit{
Criteria:\par
A. \texttt{subject\_completion} --- plausibly completes missing/cropped
subject parts.\par
B. \texttt{subject\_identity\_consistency} --- preserves subject identity and
attributes.\par
C. \texttt{context\_alignment} --- semantic + geometric consistency with the
original.\par
D. \texttt{outpaint\_realism} --- visually realistic, artifact-free in
outpainted region.\par
E. \texttt{original\_region\_preservation} --- visible region unchanged.\par
\smallskip
Focus especially on artifacts inside the outpainted region. Do not reward
overall aesthetic quality if a small region has a clear artifact (e.g.,
malformed finger).\par
\smallskip
Self-check before answering: for each artifact you cite in a reason, name the
image label (``A'' or ``B'') it appears in, and verify that your
\texttt{winner} field matches. Do NOT describe features from Image B while
marking A as the winner (or vice versa).\par
\smallskip
Return only valid JSON:\par
{\ttfamily \{\par
\quad ``subject\_completion'': \{``winner'': ``A'', ``margin'': 1,
``reason'': ``...''\},\par
\quad ``subject\_identity\_consistency'': \{``winner'': ``B'', ``margin'': 2,
``reason'': ``...''\},\par
\quad ``context\_alignment'': \{``winner'': ``tie'', ``margin'': 0,
``reason'': ``...''\},\par
\quad ``outpaint\_realism'': \{``winner'': ``A'', ``margin'': 1,
``reason'': ``...''\},\par
\quad ``original\_region\_preservation'': \{``winner'': ``tie'',
``margin'': 0, ``reason'': ``...''\},\par
\quad ``overall\_winner'': ``A \textbar{} B \textbar{} tie'',\par
\quad ``one\_sentence\_summary'': ``...''\par
\}}}
\end{minipage}}
\caption{\ANedit{System and user prompts used for the GPT-5.5 pairwise
evaluation.}}
\label{fig:mllm_judge_prompt}
\end{figure*}

\subsection{Auxiliary Loss Weight Sensitivity}
\label{sec:hf_weight_sensitivity}

\AN{We examine the sensitivity of the auxiliary high-frequency objective to
its global coefficient $\lambda_{\mathrm{hf}}$. \ANedit{All variants use the
same configuration, varying only $\lambda_{\mathrm{hf}}$.} We compare fixed
coefficients $\lambda_{\mathrm{hf}}\in\{0.1,0.3,0.5\}$.}

\begin{figure}[t]
  \centering
  \includegraphics[width=\columnwidth]{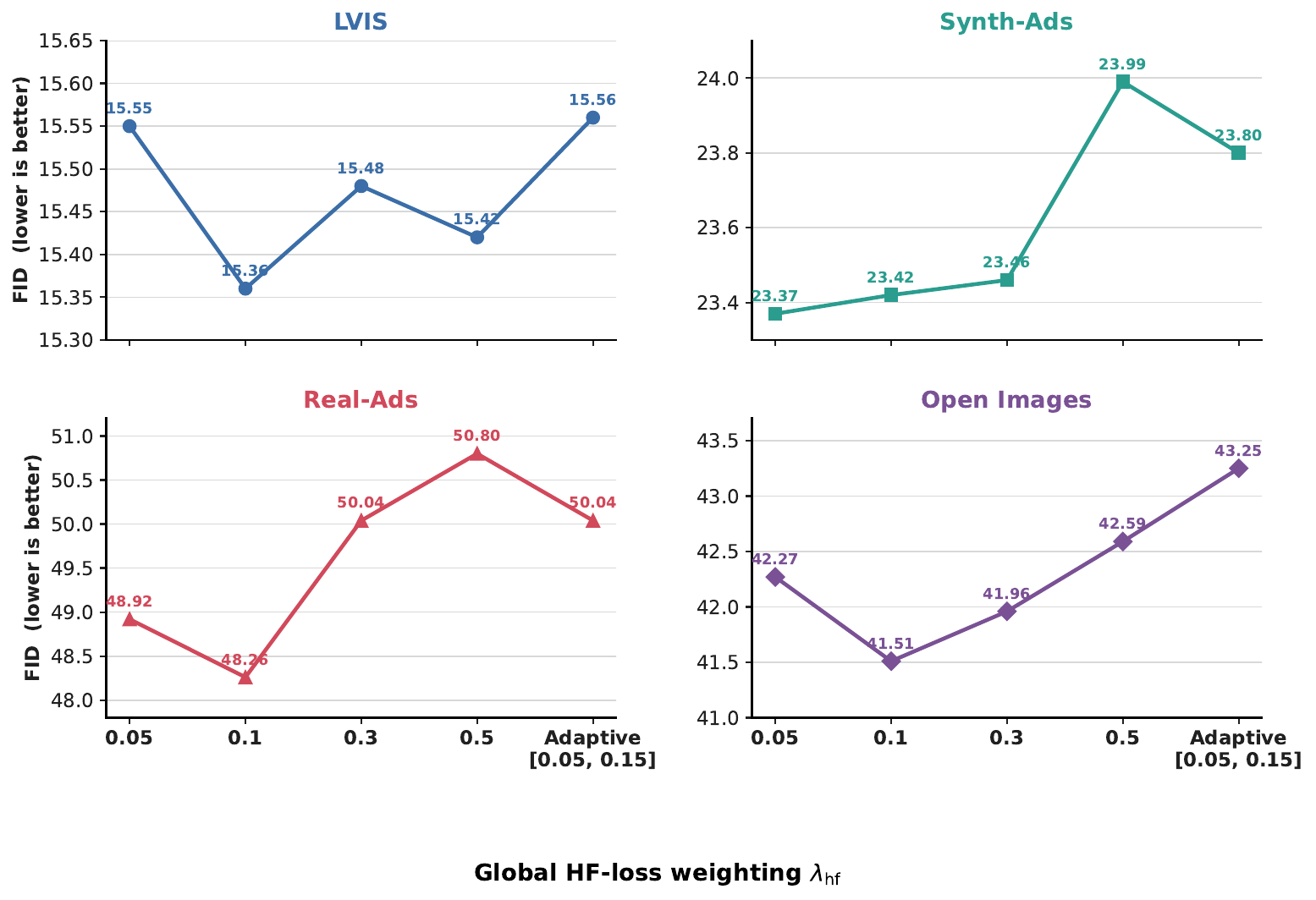}
  \caption{\AN{Auxiliary high-frequency loss-weight sensitivity.
  FID for fixed global coefficients \ANedit{and a sample-adaptive
  coefficient}. Lower is better.}}
  \label{fig:hf_weight_sensitivity}
\end{figure}

\AN{As shown in \cref{fig:hf_weight_sensitivity}, the fixed coefficient
$0.1$ gives the lowest FID on all four datasets. Increasing the coefficient
to $0.3$ or $0.5$ places progressively more emphasis on matching sparse
wavelet responses relative to the flow objective. This can over-penalize
small coefficient misalignments and favor local texture or edge energy at the
expense of globally coherent, distributionally realistic completions, which
FID measures. The effect is clearest on Real-Ads and Open Images, while the
smaller variation on LVIS and Synth-Ads indicates that the method is not
acutely sensitive near the selected operating point.}

\AN{\ANedit{We also tried a sample-adaptive coefficient
$\lambda_{\mathrm{hf}}^{(i)} = \operatorname{clip}\bigl(0.1\,(e_i/e_{\mathrm{ref}})^{1/2},\,0.05,\,0.15\bigr)$,
where $e_i$ is the mean absolute ground-truth wavelet coefficient over the
strictly visible subject region and $e_{\mathrm{ref}}$ is its training-set
median, but it gave no consistent improvement: visible subject energy
reflects texture strength rather than the difficulty of completing the hidden
subject, and the loss already adapts its subband weights and activation
across diffusion time, making a further adaptive global coefficient largely
redundant. We therefore use the stable fixed value
$\lambda_{\mathrm{hf}}=0.1$ throughout the main experiments.}}

\subsection{Effect of Subject-HF-Aware Training}
\label{sec:attention_case_study}

\AN{\ANedit{Continuing from the main paper, where
we show that subject-HF-aware training concentrates visual and textual
attention more selectively within the true subject continuation, we provide
step-level results and additional probes for the underlying attention
diagnostic case study.} We inspect direct joint attention in the final
double-stream block at denoising steps $5$, $10$, $20$, and $30$, averaged
over all 24 heads, on a representative failure case where the matched FLUX.2
SFT baseline duplicates an elephant foreleg while our subject-HF-aware model
completes the subject correctly. The maps are descriptive probes of one
network pathway rather than a causal decomposition.}

\begin{center}
  \setlength{\fboxsep}{7pt}
  \fbox{\begin{minipage}{0.92\linewidth}
    \small
    \textbf{VLM-generated scene guidance (shared by both models)}
    \par\medskip
    \ttfamily
    Fill the gray region by extending elephant to the left, with the missing
    left half of its head, ear, shoulder, and foreleg.
  \end{minipage}}
\end{center}

\AN{Both models receive the boxed VLM-generated content verbatim. The word-level probes below use the exact Qwen
subwords corresponding to the boxed scene guidance.}

\noindent\textbf{Attention definitions.}
\AN{Let $A_h^{(l,t)}$ be the direct joint-attention matrix for head $h$, layer
$l$, and denoising step $t$. We average over $H$ heads without attention
rollout:}
\begin{equation}
  A_h^{(l,t)}
  = \operatorname{softmax}\!\left(
      \frac{Q_h^{(l,t)}K_h^{(l,t)\top}}{\sqrt{d_h}}
    \right),
  \quad
  \bar A^{(l,t)}=\frac{1}{H}\sum_{h=1}^{H}A_h^{(l,t)}.
  \label{eq:attention-direct}
\end{equation}
\AN{For each image query $i$, we measure attention to visible-subject image
keys $V$ and to all Qwen subwords $T_w$ corresponding to prompt word $w$:}
\begin{equation}
  M_{\mathrm{vis}}(i)
  = \sum_{j\in V}\bar A^{(l,t)}_{ij},
  \qquad
  M_w(i)
  = \sum_{j\in T_w}\bar A^{(l,t)}_{ij}.
  \label{eq:attention-routing}
\end{equation}
\AN{Let $H_{\mathrm{subj}}$ denote outpaint queries overlapping the hidden
ground-truth subject and let $O$ denote the remaining outpaint queries. For
either routing map $M$, we compute}
\begin{equation}
  \begin{aligned}
  \mu_H(M) &= \frac{1}{|H_{\mathrm{subj}}|}
    \sum_{i\in H_{\mathrm{subj}}}M(i),&
  \mu_O(M) &= \frac{1}{|O|}\sum_{i\in O}M(i),\\
  R(M) &= \frac{\mu_H(M)}{\mu_O(M)}.&&
  \end{aligned}
  \label{eq:attention-metrics}
\end{equation}
\AN{Here $\mu_H$ measures routing within the correct subject continuation,
$\mu_O$ measures \emph{unsupported-region routing}, i.e., attention assigned
to locations that should remain background, and $R$ measures spatial
localization. The subject mask is used only for this post-hoc analysis and is
never supplied to either model.}

\begin{figure}[htbp]
  \centering
  \begin{subfigure}[t]{\linewidth}
    \centering
    \includegraphics[width=\linewidth]{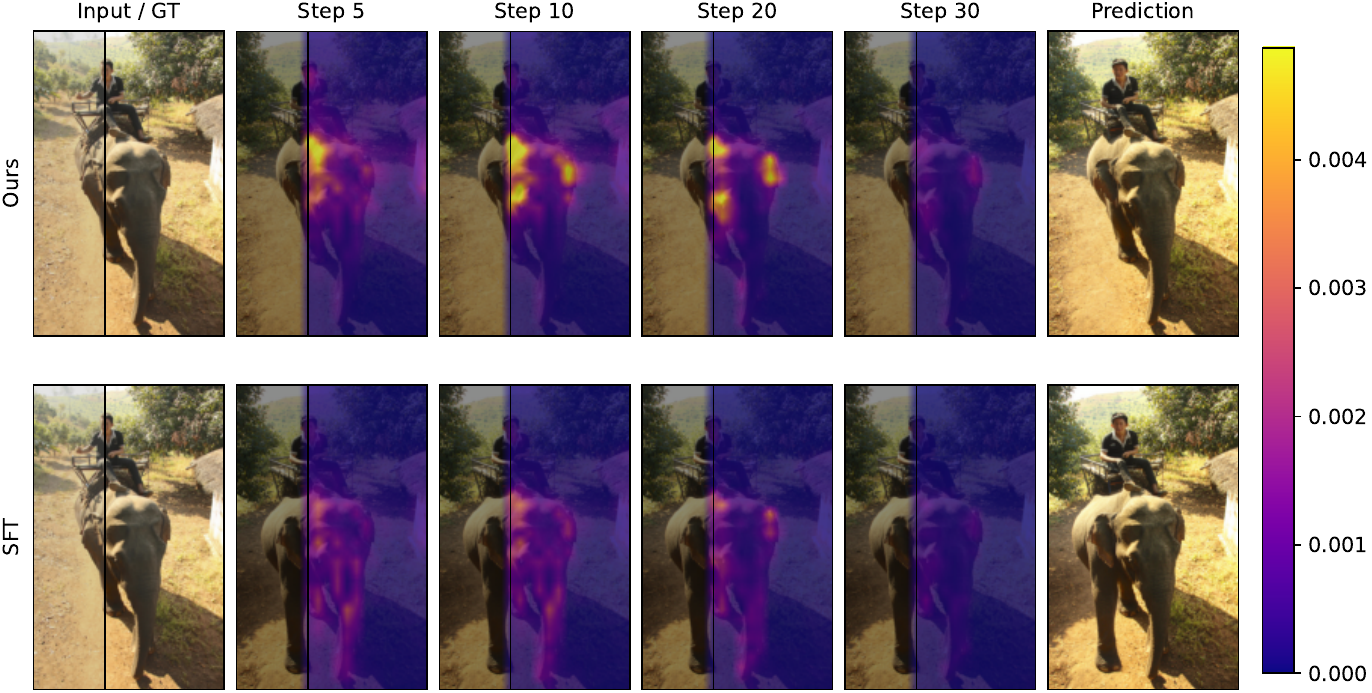}
    \caption{Visible-source retrieval: where generated-subject queries obtain
    evidence in the visible conditioning image.}
    \label{fig:visible_source_retrieval}
  \end{subfigure}
  \vspace{5pt}
  \begin{subfigure}[t]{\linewidth}
    \centering
    \includegraphics[width=\linewidth]{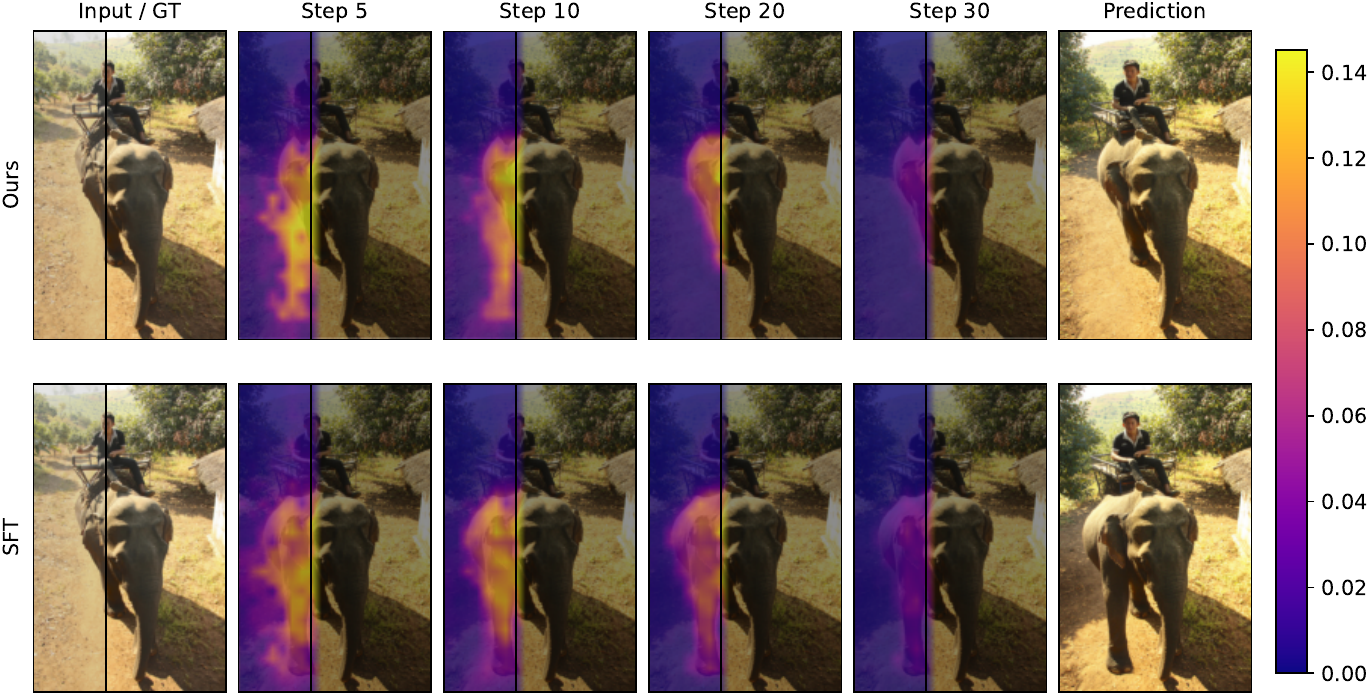}
    \caption{Generated-region routing: where visible-subject evidence is
    deployed across outpaint queries.}
    \label{fig:generated_region_routing}
  \end{subfigure}
  \caption{\AN{\textbf{Two complementary directions of visual attention.}
  Rows compare ours with matched SFT; columns show denoising steps. Both models
  retrieve evidence from the correct visible subject in \textbf{(a)}, but
  differ in where they deploy that evidence in \textbf{(b)}.}}
  \label{fig:attention_diagnostic_pair}
\end{figure}

\noindent\textbf{Visible-source retrieval.}
\AN{We first ask where generated-subject queries retrieve visual evidence from.
For generated-subject queries $G_m$ from model $m$ and a visible conditioning
location $j$, the \emph{visible-source retrieval map} is}
\begin{equation}
  S_m(j)=\frac{1}{H|G_m|}\sum_{h=1}^{H}\sum_{i\in G_m}A_{hij}.
  \label{eq:visible_source_retrieval}
\end{equation}
\AN{We obtain $G_m$ post hoc from each prediction using
GroundingDINO~\cite{liu2024grounding} and SAM2~\cite{ravi2025sam}, restricted to the outpaint region. \Cref{fig:visible_source_retrieval}
shows that both models retrieve evidence primarily from the visible elephant;
SFT exhibits no conspicuous high-attention source in the background. A
shared-query control using the same ground-truth hidden-subject locations for
both models reaches the same conclusion: SFT is more concentrated on the
visible subject in 11 of 12 layer - step comparisons. The failure therefore
cannot be explained by SFT consulting an obviously incorrect visible source.}

\noindent\textbf{Generated-region routing.}
\AN{We next reverse the question. For every generated query $i$ in the
outpaint region, $M_{\mathrm{vis}}(i)$ from \cref{eq:attention-routing} sums
its attention to all visible-subject keys. This \emph{generated-region routing
map} reveals where retrieved subject evidence is deployed. Unlike the source
maps, \cref{fig:generated_region_routing} separates the models: SFT spreads
visible-elephant evidence into a larger unsupported region, while ours
concentrates it around the plausible continuation.}

\AN{The step-resolved measurements in \cref{tab:attention_case_metrics}
quantify this distinction. Ours has lower unsupported-region routing of
visible-subject evidence and a higher localization ratio at every measured
step. The gap grows late in denoising: at step 20, ours reaches $R=13.54$
versus $4.91$ for SFT. Thus, ours does not merely reduce visual conditioning
uniformly; it deploys it more selectively.}

\begin{table}[t]
  \centering
  \caption{\AN{Step-resolved routing metrics in the final double-stream block.
  Lower $\mu_O$ indicates less unsupported-region routing, while higher $R$
  indicates better localization. Best values within each probe and step are
  bold.}}
  \label{tab:attention_case_metrics}
  \footnotesize
  \setlength{\tabcolsep}{3.5pt}
  \resizebox{\columnwidth}{!}{%
  \begin{tabular}{c cc cc cc cc}
    \toprule
    & \multicolumn{4}{c}{\textbf{Visible-subject routing}} &
      \multicolumn{4}{c}{\textbf{``Foreleg'' routing}} \\
    \cmidrule(lr){2-5}\cmidrule(lr){6-9}
    \textbf{Step} & \multicolumn{2}{c}{$\mu_O$} & \multicolumn{2}{c}{$R$} &
      \multicolumn{2}{c}{$\mu_O$} & \multicolumn{2}{c}{$R$} \\
    & SFT & Ours & SFT & Ours & SFT & Ours & SFT & Ours \\
    \midrule
    5  & 0.0341 & \textbf{0.0277} & 2.56 & \textbf{3.23} & 0.00341 & \textbf{0.00056} & \textbf{0.66} & 0.31 \\
    10 & 0.0251 & \textbf{0.0207} & 4.03 & \textbf{5.15} & 0.00238 & \textbf{0.00102} & \textbf{0.67} & 0.59 \\
    20 & 0.0162 & \textbf{0.0068} & 4.91 & \textbf{13.54} & 0.00209 & \textbf{0.00024} & 0.90 & \textbf{5.39} \\
    30 & 0.0063 & \textbf{0.0026} & 5.13 & \textbf{11.63} & 0.00087 & \textbf{0.00022} & 1.47 & \textbf{4.91} \\
    \bottomrule
  \end{tabular}}
\end{table}

\begin{figure*}[p]
  \centering
  \begin{subfigure}[t]{0.49\textwidth}
    \centering
    \includegraphics[width=\linewidth]{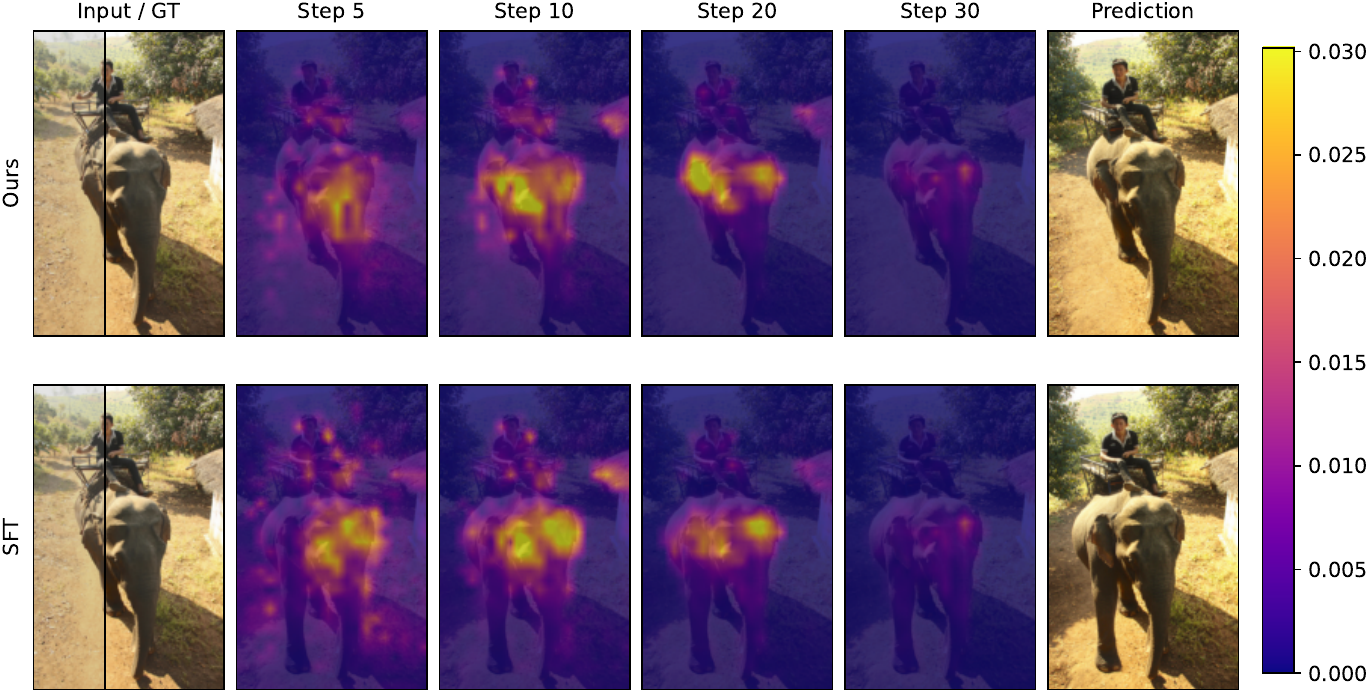}
    \caption{``Elephant''}
  \end{subfigure}\hfill
  \begin{subfigure}[t]{0.49\textwidth}
    \centering
    \includegraphics[width=\linewidth]{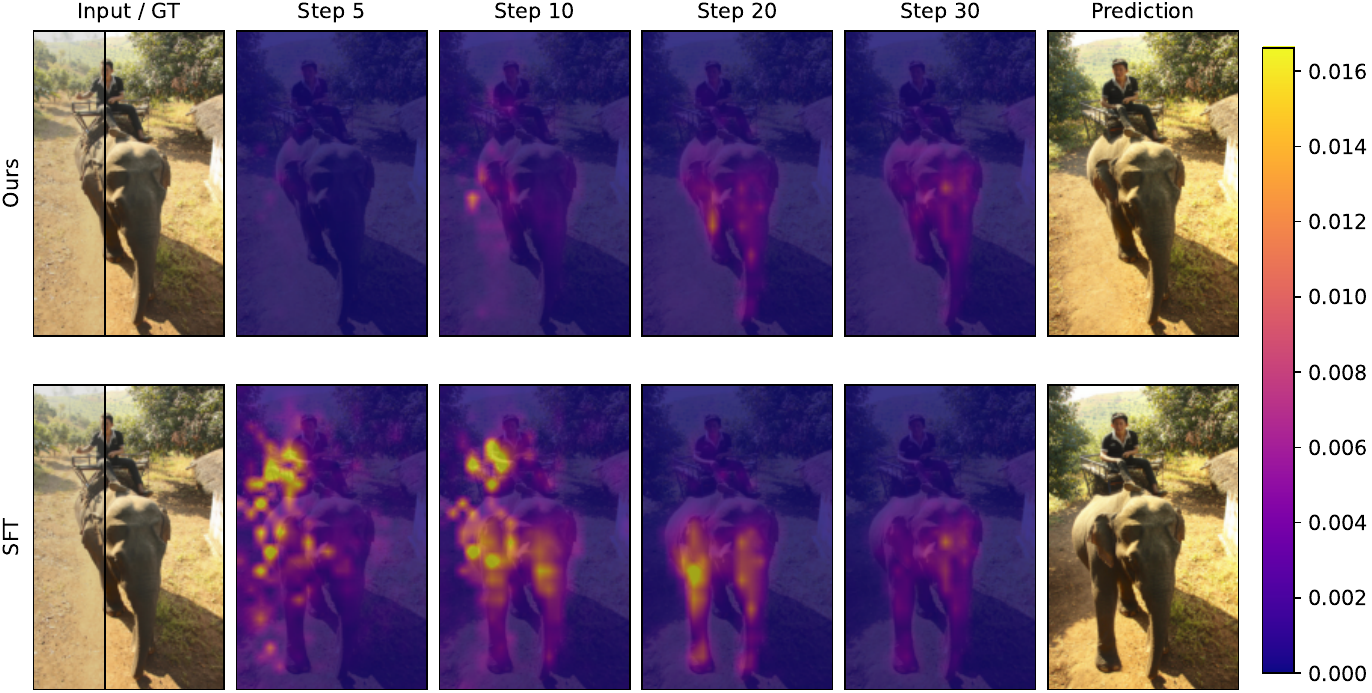}
    \caption{``Head''}
  \end{subfigure}
  \par\vspace{4pt}
  \begin{subfigure}[t]{0.49\textwidth}
    \centering
    \includegraphics[width=\linewidth]{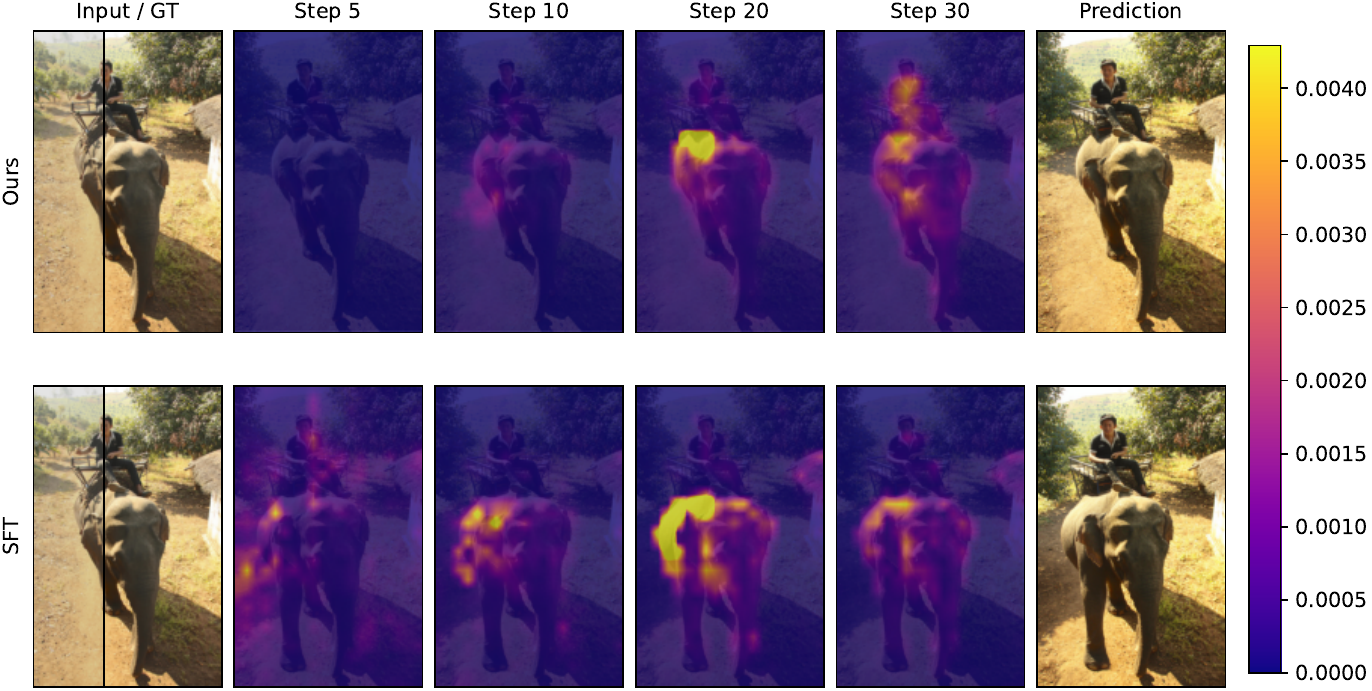}
    \caption{``Shoulder''}
  \end{subfigure}\hfill
  \begin{subfigure}[t]{0.49\textwidth}
    \centering
    \includegraphics[width=\linewidth]{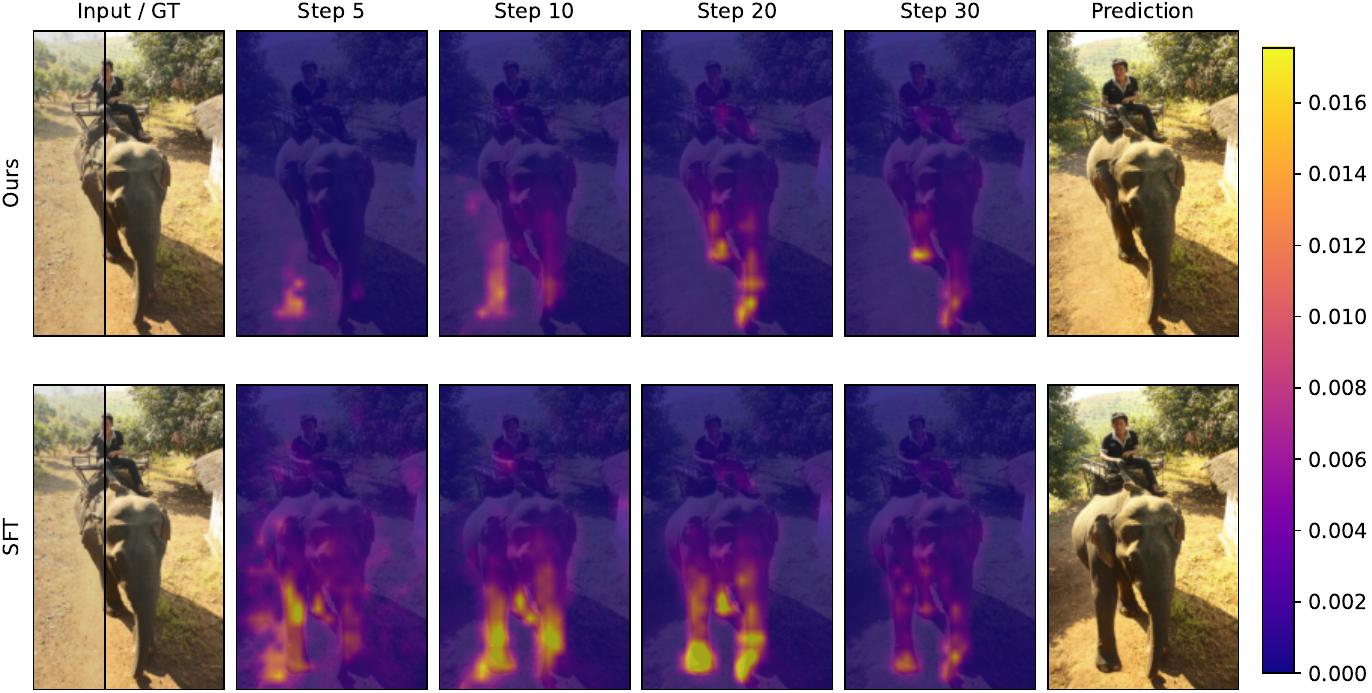}
    \caption{``Foreleg''}
    \label{fig:attention_word_foreleg}
  \end{subfigure}
  \caption{\AN{\textbf{Word-conditioned routing.} Direct attention from
  image queries to selected prompt subwords. The ``foreleg'' cue is the
  clearest discriminator: SFT spreads it into unsupported locations,
  consistent with the duplicated limb, whereas ours increasingly confines it
  to the valid subject continuation}}
  \label{fig:attention_word_routing}
\end{figure*}

\noindent\textbf{Word-conditioned routing.}
\AN{Because FLUX.2 performs joint image--text attention, we finally replace
the visible-subject key set with the Qwen subwords for individual prompt words.
The prompt explicitly requests the missing ``head, ear, shoulder, and
foreleg.'' \Cref{fig:attention_word_routing} shows that not every word produces
the same spatial response. Most notably, SFT routes substantially more
``foreleg'' attention into non-subject outpaint locations at all four steps.
Ours also attends to this cue, but its routing becomes sharply localized late
in denoising: its ``foreleg'' ratio rises from $0.59$ at step 10 to $5.39$ at
step 20, while SFT remains below $1$ at step 20. This transition coincides
with the structurally correct completion. In contrast, SFT's strong response
to ``foreleg'' outside the supported subject region can reinforce the wrong
subject continuation: the semantic cue is relevant, but it is applied at
locations where the emerging image structure does not support another part.}

\AN{Together, these probes support a selective-grounding interpretation of
subject-HF-aware training. The objective does not make the model insensitive
to text, nor does it simply increase attention to the visible subject. Rather,
both models retrieve the appropriate visual source, but ours increasingly
gates visual and anatomy-word evidence according to where fine subject
structure is supported. This behavior is consistent with the intended effect
of multiscale high-frequency supervision: preserving coherent boundaries and
anatomical detail constrains how semantic cues are spatially realized. Since
attention omits value vectors, output projections, MLP paths, and later
nonlinear interactions, we treat this as evidence consistent with the
mechanism, not proof that the observed maps cause the final prediction.}

\subsection{Qualitative: Ablations}
\label{sec:qualitative_ablations}

\begin{figure*}[p]
  \centering
  \begin{subfigure}[t]{0.85\textwidth}
    \scriptsize
    \makebox[\linewidth][c]{%
      \makebox[0.20\linewidth][c]{\textbf{Input / GT}}%
      \makebox[0.20\linewidth][c]{\textbf{w/o subj. localization}}%
      \makebox[0.20\linewidth][c]{\textbf{FLUX.2 SFT}}%
      \makebox[0.20\linewidth][c]{%
        \shortstack{\textbf{w/o N.S.}\\
                    \textbf{\& S.A.}}}%
      \makebox[0.20\linewidth][c]{\textbf{Ours}}}
  \end{subfigure}\par\smallskip
  \begin{subfigure}[t]{0.85\textwidth}
    \includegraphics[width=\linewidth]{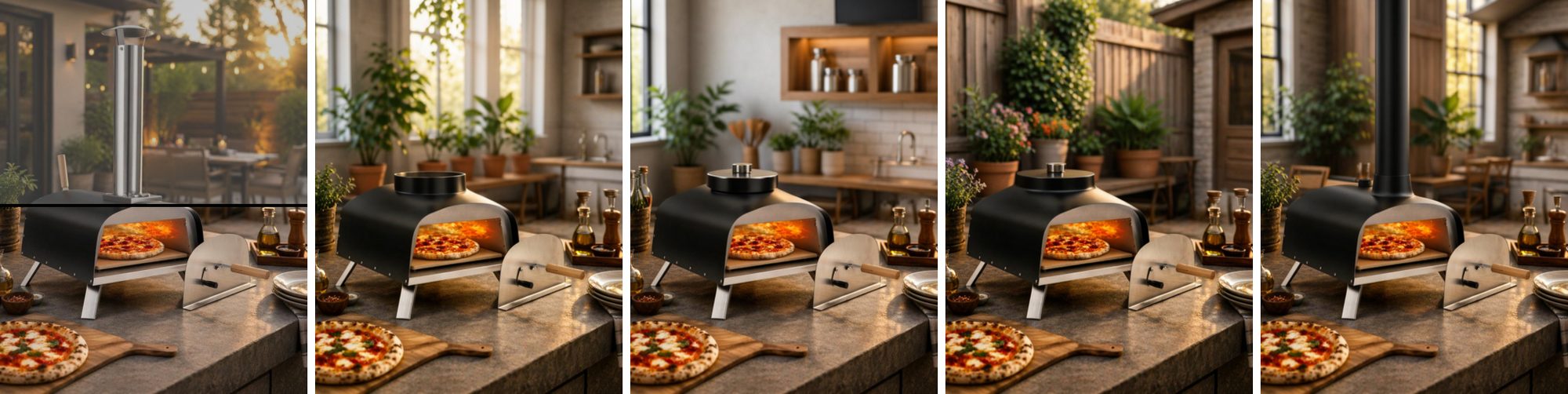}
  \end{subfigure}\par\smallskip
  \begin{subfigure}[t]{0.85\textwidth}
    \includegraphics[width=\linewidth]{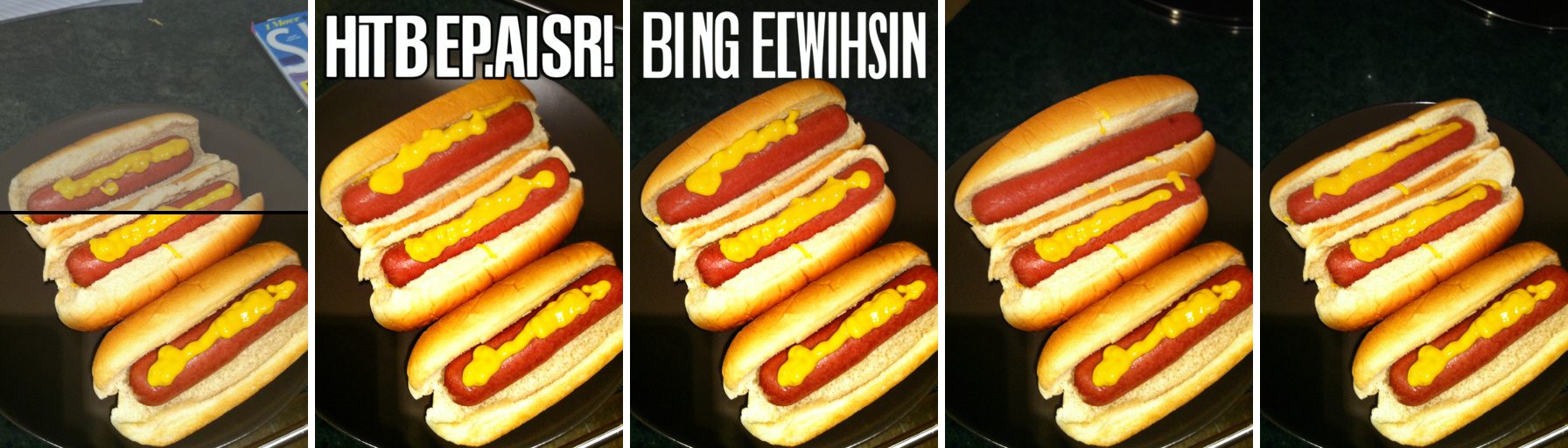}
  \end{subfigure}\par\smallskip
  \begin{subfigure}[t]{0.85\textwidth}
    \includegraphics[width=\linewidth]{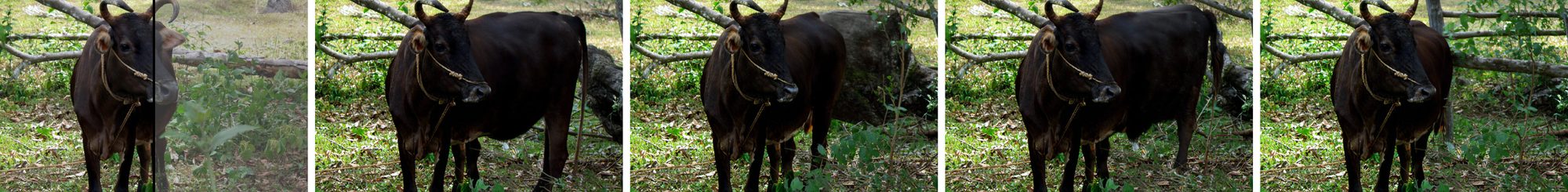}
  \end{subfigure}\par\smallskip
  \begin{subfigure}[t]{0.85\textwidth}
    \includegraphics[width=\linewidth]{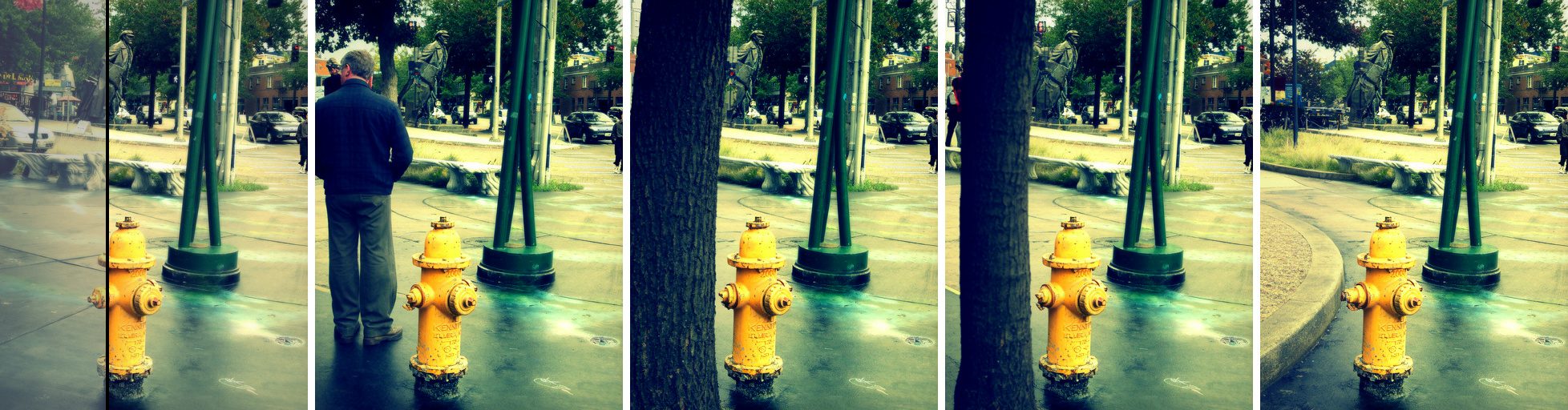}
  \end{subfigure}\par\smallskip
  \begin{subfigure}[t]{0.85\textwidth}
    \includegraphics[width=\linewidth]{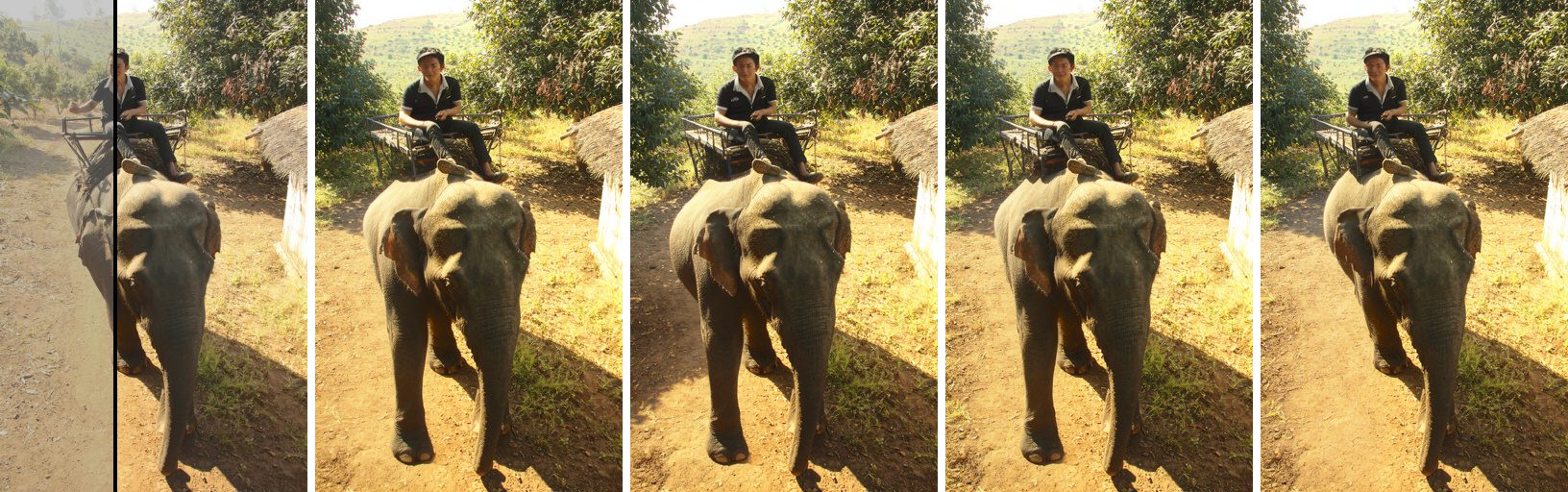}
  \end{subfigure}\par\smallskip
  \begin{subfigure}[t]{0.85\textwidth}
    \includegraphics[width=\linewidth]{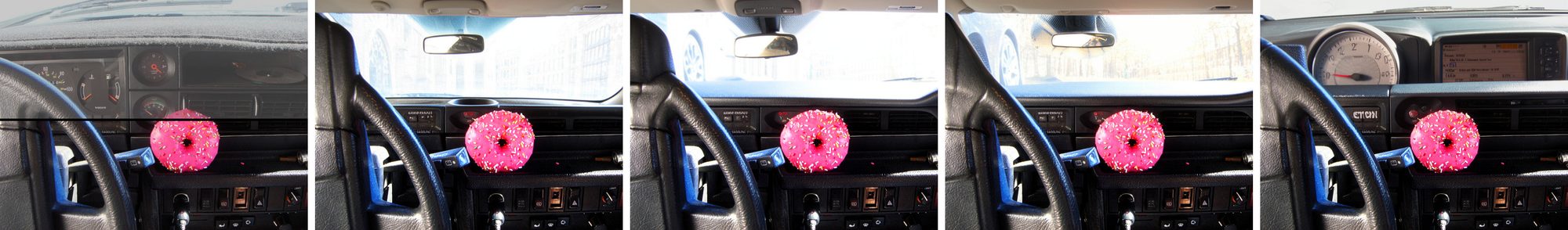}
  \end{subfigure}
  \caption{\AN{\textbf{Additional qualitative ablations.} \ANedit{N.S. \& S.A. denote noise
  schedule and subband adaptation, respectively}}}
  \label{fig:additional_qualitative_ablations}
\end{figure*}

\AN{\Cref{fig:additional_qualitative_ablations}
shows how the quantitative ablation trends 
manifest visually. Without subject localization, the wavelet objective can reward
high-frequency content anywhere in the outpaint region, encouraging
additional text, accessories, or scene detail that competes with the intended
subject continuation. FLUX.2 SFT often produces a plausible extension but can
alter identity-defining appearance or introduce unsupported fine structure
because it lacks direct subject-detail supervision. Removing both the noise
schedule and subband adaptation treats scales and subjects less selectively,
leading to weaker shape continuity and less faithful local texture. The full
model more consistently extends the visible subject without introducing
distracting detail.}

\subsection{Qualitative: Main Results}

\begin{figure*}[p]
  \centering
  \scriptsize
  \setlength{\tabcolsep}{4pt}
  \renewcommand{\arraystretch}{1.02}
  \setlength{\extrarowheight}{0.5pt}
  \newcommand{\PartOneCell}[2]{%
    \includegraphics[height=0.09\textheight]{figures/qualitative/ads/#1/#2.jpg}}
  \newcommand{\PartOneInput}[1]{%
    \includegraphics[height=0.09\textheight]{figures/qualitative/part1_cells/#1_input_gt.jpg}}
  \newcommand{\PartOneVIP}[1]{%
    \includegraphics[height=0.09\textheight]{figures/qualitative/part1_cells/#1_vip.jpg}}
  \newcommand{\PartOneLabel}[1]{%
    \raisebox{\dimexpr0.045\textheight-0.5\height\relax}{%
      \rotatebox[origin=c]{90}{\fontsize{4.5}{5}\selectfont\textbf{#1}}}}
  \begin{tabular}{@{}c@{\hspace{2pt}}cccccc@{}}
    \PartOneLabel{Input / GT}
      & \PartOneInput{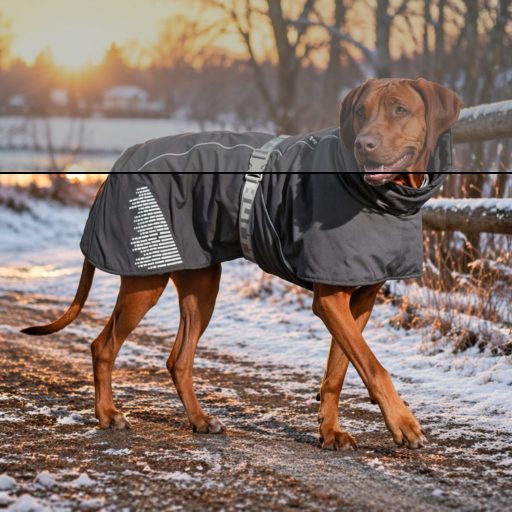}
      & \PartOneInput{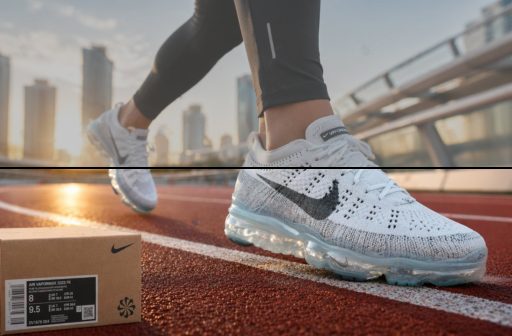}
      & \PartOneInput{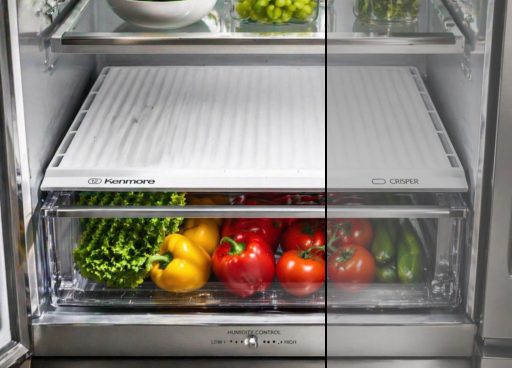}
      & \PartOneInput{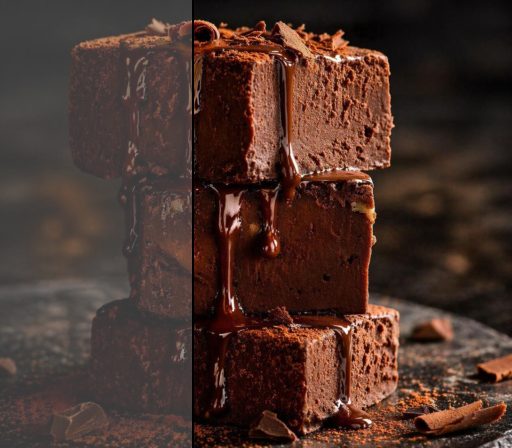}
      & \PartOneInput{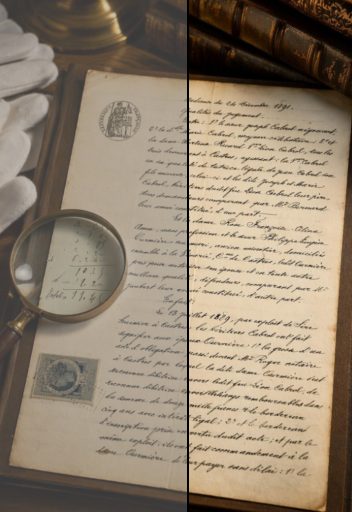}
      & \PartOneInput{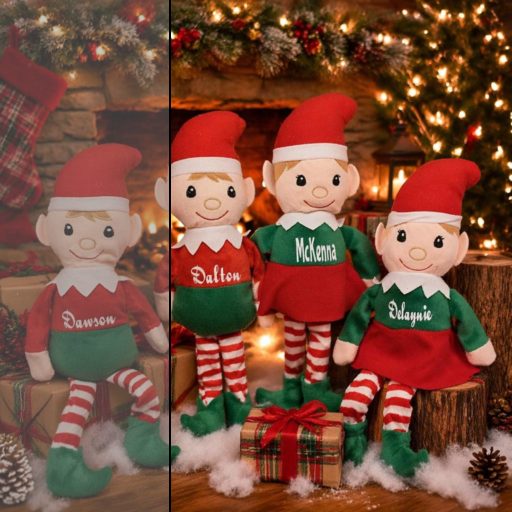} \\
    \PartOneLabel{BrushNet}
      & \PartOneCell{121812439288}{brushnet}
      & \PartOneCell{164449117403}{brushnet}
      & \PartOneCell{165485290655}{brushnet}
      & \PartOneCell{167776202638}{brushnet}
      & \PartOneCell{170519176871}{brushnet}
      & \PartOneCell{172262803061}{brushnet} \\
    \PartOneLabel{PowerPaint}
      & \PartOneCell{121812439288}{powerpaint}
      & \PartOneCell{164449117403}{powerpaint}
      & \PartOneCell{165485290655}{powerpaint}
      & \PartOneCell{167776202638}{powerpaint}
      & \PartOneCell{170519176871}{powerpaint}
      & \PartOneCell{172262803061}{powerpaint} \\
    \PartOneLabel{VIP}
      & \PartOneVIP{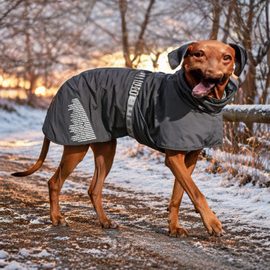}
      & \PartOneVIP{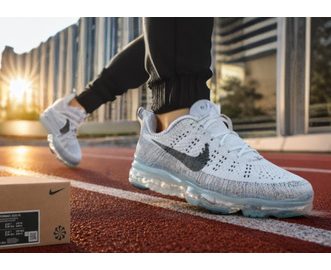}
      & \PartOneVIP{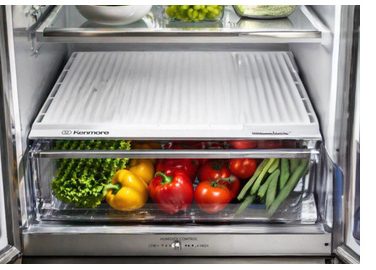}
      & \PartOneVIP{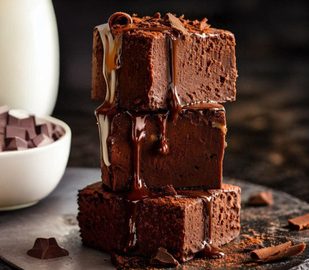}
      & \PartOneVIP{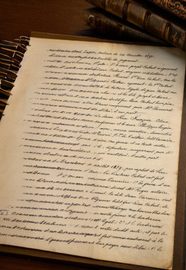}
      & \PartOneVIP{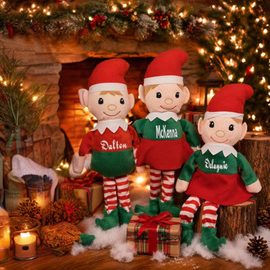} \\
    \PartOneLabel{FLUX.1 Fill}
      & \PartOneCell{121812439288}{flux1_fill}
      & \PartOneCell{164449117403}{flux1_fill}
      & \PartOneCell{165485290655}{flux1_fill}
      & \PartOneCell{167776202638}{flux1_fill}
      & \PartOneCell{170519176871}{flux1_fill}
      & \PartOneCell{172262803061}{flux1_fill} \\
    \PartOneLabel{DING}
      & \PartOneCell{121812439288}{ding}
      & \PartOneCell{164449117403}{ding}
      & \PartOneCell{165485290655}{ding}
      & \PartOneCell{167776202638}{ding}
      & \PartOneCell{170519176871}{ding}
      & \PartOneCell{172262803061}{ding} \\
    \PartOneLabel{FlowChef}
      & \PartOneCell{121812439288}{flowchef}
      & \PartOneCell{164449117403}{flowchef}
      & \PartOneCell{165485290655}{flowchef}
      & \PartOneCell{167776202638}{flowchef}
      & \PartOneCell{170519176871}{flowchef}
      & \PartOneCell{172262803061}{flowchef} \\
    \PartOneLabel{fal/lora-outpaint}
      & \PartOneCell{121812439288}{fal_flux2}
      & \PartOneCell{164449117403}{fal_flux2}
      & \PartOneCell{165485290655}{fal_flux2}
      & \PartOneCell{167776202638}{fal_flux2}
      & \PartOneCell{170519176871}{fal_flux2}
      & \PartOneCell{172262803061}{fal_flux2} \\
    \PartOneLabel{FLUX.2 SFT}
      & \PartOneCell{121812439288}{flux2_sft}
      & \PartOneCell{164449117403}{flux2_sft}
      & \PartOneCell{165485290655}{flux2_sft}
      & \PartOneCell{167776202638}{flux2_sft}
      & \PartOneCell{170519176871}{flux2_sft}
      & \PartOneCell{172262803061}{flux2_sft} \\
    \PartOneLabel{Ours}
      & \PartOneCell{121812439288}{ours}
      & \PartOneCell{164449117403}{ours}
      & \PartOneCell{165485290655}{ours}
      & \PartOneCell{167776202638}{ours}
      & \PartOneCell{170519176871}{ours}
      & \PartOneCell{172262803061}{ours}
  \end{tabular}
  \caption{\AN{Additional qualitative comparisons (Part 1). Columns denote
  samples and rows denote methods.}}
  \label{fig:additional_qualitative_1}
\end{figure*}

\begin{figure*}[p]
  \centering
  \scriptsize
  \setlength{\tabcolsep}{3pt}
  \renewcommand{\arraystretch}{1.02}
  \setlength{\extrarowheight}{0.5pt}
  \newcommand{\PartTwoCell}[3]{%
    \includegraphics[height=0.091\textheight]{figures/qualitative/#1/#2/#3.jpg}}
  \newcommand{\PartTwoInput}[1]{%
    \includegraphics[height=0.091\textheight]{figures/qualitative/part2_cells/#1_input_gt.jpg}}
  \newcommand{\PartTwoVIP}[1]{%
    \includegraphics[height=0.091\textheight]{figures/qualitative/part2_cells/#1_vip.jpg}}
  \newcommand{\PartTwoLabel}[1]{%
    \raisebox{\dimexpr0.0455\textheight-0.5\height\relax}{%
      \rotatebox[origin=c]{90}{\fontsize{4.5}{5}\selectfont\textbf{#1}}}}
  \begin{tabular}{@{}c@{\hspace{2pt}}cccccc@{}}
    \PartTwoLabel{Input / GT}
      & \PartTwoInput{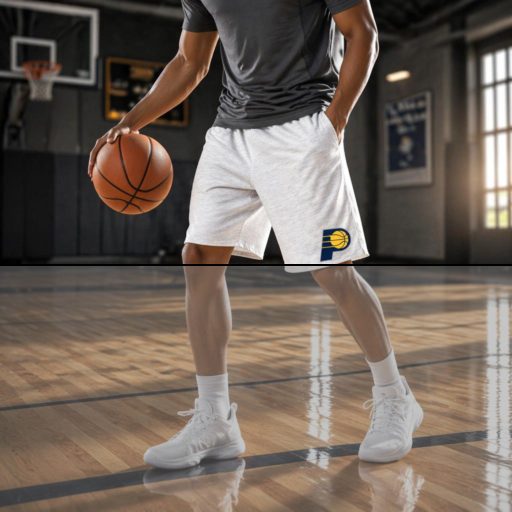}
      & \PartTwoInput{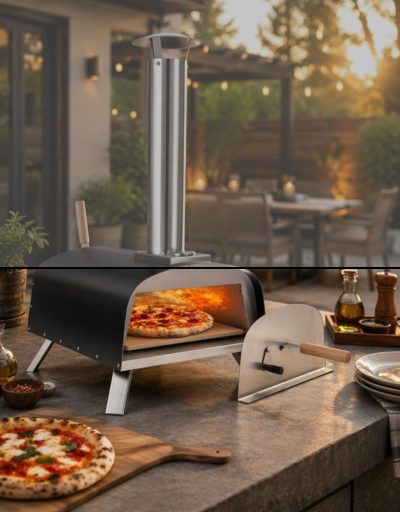}
      & \PartTwoInput{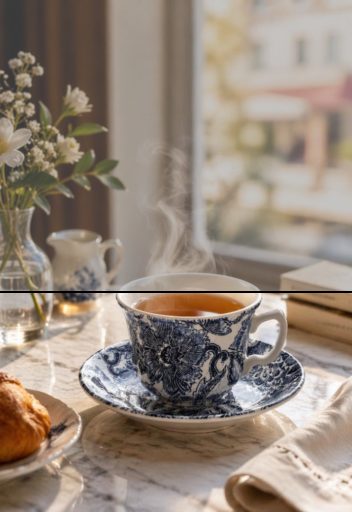}
      & \PartTwoInput{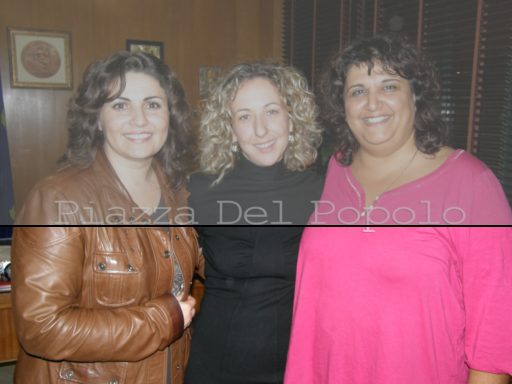}
      & \PartTwoInput{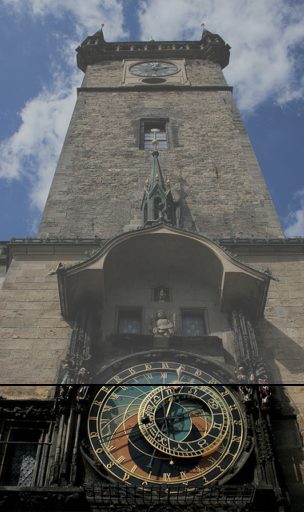}
      & \PartTwoInput{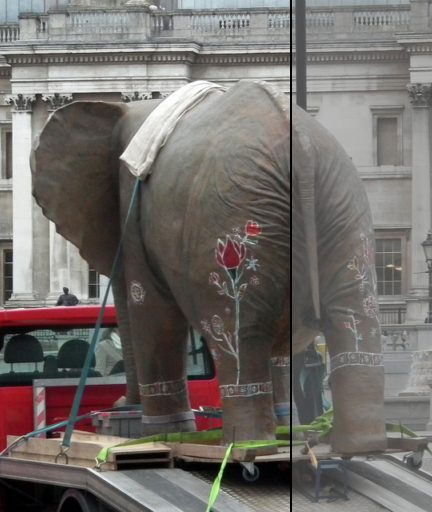} \\
    \PartTwoLabel{BrushNet}
      & \PartTwoCell{ads}{172529431199}{brushnet}
      & \PartTwoCell{ads}{172557213348}{brushnet}
      & \PartTwoCell{ads}{87016611605}{brushnet}
      & \PartTwoCell{oi}{f44a796ccefadfed}{brushnet}
      & \PartTwoCell{lvis}{000000024582}{brushnet}
      & \PartTwoCell{lvis}{000000027989}{brushnet} \\
    \PartTwoLabel{PowerPaint}
      & \PartTwoCell{ads}{172529431199}{powerpaint}
      & \PartTwoCell{ads}{172557213348}{powerpaint}
      & \PartTwoCell{ads}{87016611605}{powerpaint}
      & \PartTwoCell{oi}{f44a796ccefadfed}{powerpaint}
      & \PartTwoCell{lvis}{000000024582}{powerpaint}
      & \PartTwoCell{lvis}{000000027989}{powerpaint} \\
    \PartTwoLabel{VIP}
      & \PartTwoVIP{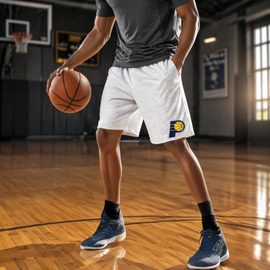}
      & \PartTwoVIP{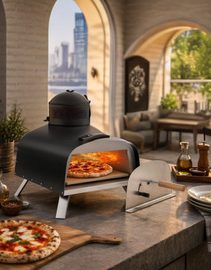}
      & \PartTwoVIP{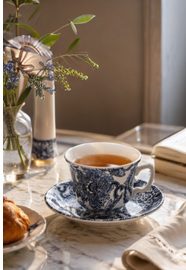}
      & \PartTwoVIP{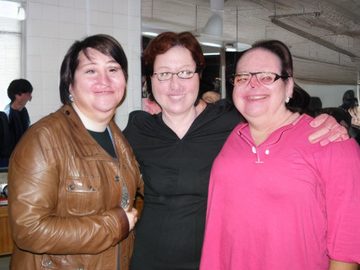}
      & \PartTwoVIP{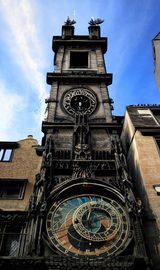}
      & \PartTwoVIP{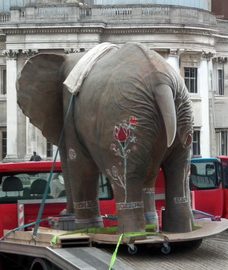} \\
    \PartTwoLabel{FLUX.1 Fill}
      & \PartTwoCell{ads}{172529431199}{flux1_fill}
      & \PartTwoCell{ads}{172557213348}{flux1_fill}
      & \PartTwoCell{ads}{87016611605}{flux1_fill}
      & \PartTwoCell{oi}{f44a796ccefadfed}{flux1_fill}
      & \PartTwoCell{lvis}{000000024582}{flux1_fill}
      & \PartTwoCell{lvis}{000000027989}{flux1_fill} \\
    \PartTwoLabel{DING}
      & \PartTwoCell{ads}{172529431199}{ding}
      & \PartTwoCell{ads}{172557213348}{ding}
      & \PartTwoCell{ads}{87016611605}{ding}
      & \PartTwoCell{oi}{f44a796ccefadfed}{ding}
      & \PartTwoCell{lvis}{000000024582}{ding}
      & \PartTwoCell{lvis}{000000027989}{ding} \\
    \PartTwoLabel{FlowChef}
      & \PartTwoCell{ads}{172529431199}{flowchef}
      & \PartTwoCell{ads}{172557213348}{flowchef}
      & \PartTwoCell{ads}{87016611605}{flowchef}
      & \PartTwoCell{oi}{f44a796ccefadfed}{flowchef}
      & \PartTwoCell{lvis}{000000024582}{flowchef}
      & \PartTwoCell{lvis}{000000027989}{flowchef} \\
    \PartTwoLabel{fal/lora-outpaint}
      & \PartTwoCell{ads}{172529431199}{fal_flux2}
      & \PartTwoCell{ads}{172557213348}{fal_flux2}
      & \PartTwoCell{ads}{87016611605}{fal_flux2}
      & \PartTwoCell{oi}{f44a796ccefadfed}{fal_flux2}
      & \PartTwoCell{lvis}{000000024582}{fal_flux2}
      & \PartTwoCell{lvis}{000000027989}{fal_flux2} \\
    \PartTwoLabel{FLUX.2 SFT}
      & \PartTwoCell{ads}{172529431199}{flux2_sft}
      & \PartTwoCell{ads}{172557213348}{flux2_sft}
      & \PartTwoCell{ads}{87016611605}{flux2_sft}
      & \PartTwoCell{oi}{f44a796ccefadfed}{flux2_sft}
      & \PartTwoCell{lvis}{000000024582}{flux2_sft}
      & \PartTwoCell{lvis}{000000027989}{flux2_sft} \\
    \PartTwoLabel{Ours}
      & \PartTwoCell{ads}{172529431199}{ours}
      & \PartTwoCell{ads}{172557213348}{ours}
      & \PartTwoCell{ads}{87016611605}{ours}
      & \PartTwoCell{oi}{f44a796ccefadfed}{ours}
      & \PartTwoCell{lvis}{000000024582}{ours}
      & \PartTwoCell{lvis}{000000027989}{ours}
  \end{tabular}
  \caption{\AN{Additional qualitative comparisons (Part 2). Columns denote
  samples and rows denote methods.}}
  \label{fig:additional_qualitative_2}
\end{figure*}

\begin{figure*}[p]
  \centering
  \scriptsize
  \setlength{\tabcolsep}{4pt}
  \renewcommand{\arraystretch}{1.02}
  \setlength{\extrarowheight}{0.5pt}
  \newcommand{\PartThreeCell}[2]{%
    \includegraphics[height=0.091\textheight]{figures/qualitative/lvis/#1/#2.jpg}}
  \newcommand{\PartThreeOICell}[2]{%
    \includegraphics[height=0.091\textheight]{figures/qualitative/oi/#1/#2.jpg}}
  \newcommand{\PartThreeInput}[1]{%
    \includegraphics[height=0.091\textheight]{figures/qualitative/part3_cells/#1_input_gt.jpg}}
  \newcommand{\PartThreeVIP}[1]{%
    \includegraphics[height=0.091\textheight]{figures/qualitative/part3_cells/#1_vip.jpg}}
  \newcommand{\PartThreeLabel}[1]{%
    \raisebox{\dimexpr0.0455\textheight-0.5\height\relax}{%
      \rotatebox[origin=c]{90}{\fontsize{4.5}{5}\selectfont\textbf{#1}}}}
  \begin{tabular}{@{}c@{\hspace{2pt}}cccccc@{}}
    \PartThreeLabel{Input / GT}
      & \PartThreeInput{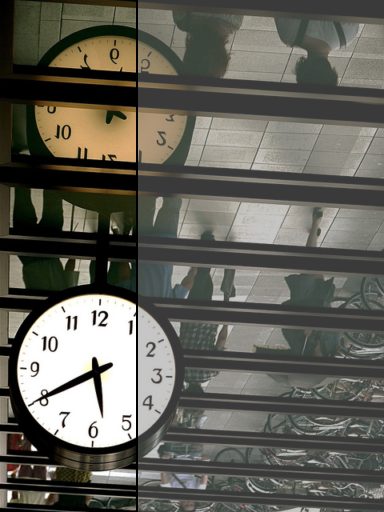}
      & \PartThreeInput{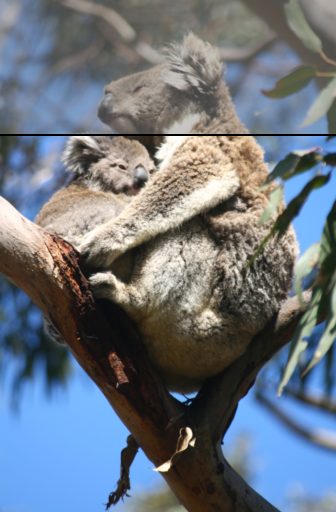}
      & \PartThreeInput{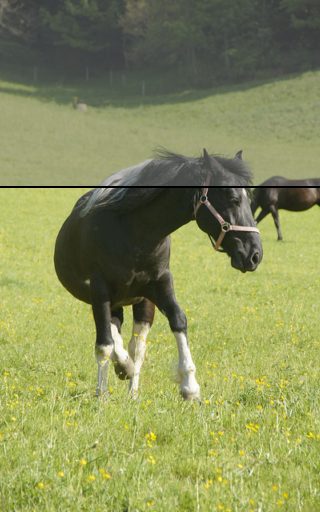}
      & \PartThreeInput{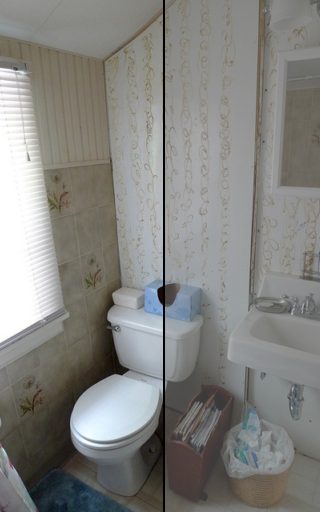}
      & \PartThreeInput{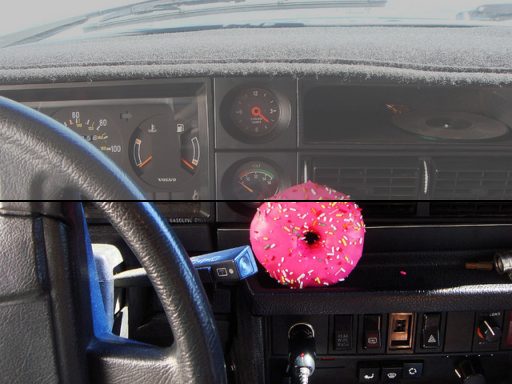}
      & \PartThreeInput{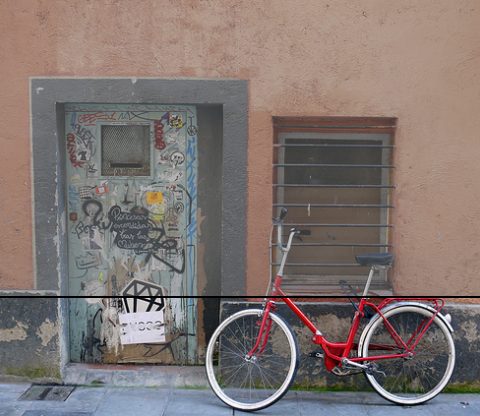} \\
    \PartThreeLabel{BrushNet}
      & \PartThreeCell{000000303408}{brushnet}
      & \PartThreeOICell{9d5252360090f591}{brushnet}
      & \PartThreeCell{000000095988}{brushnet}
      & \PartThreeCell{000000151832}{brushnet}
      & \PartThreeCell{000000196002}{brushnet}
      & \PartThreeCell{000000203317}{brushnet} \\
    \PartThreeLabel{PowerPaint}
      & \PartThreeCell{000000303408}{powerpaint}
      & \PartThreeOICell{9d5252360090f591}{powerpaint}
      & \PartThreeCell{000000095988}{powerpaint}
      & \PartThreeCell{000000151832}{powerpaint}
      & \PartThreeCell{000000196002}{powerpaint}
      & \PartThreeCell{000000203317}{powerpaint} \\
    \PartThreeLabel{VIP}
      & \PartThreeVIP{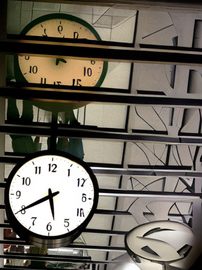}
      & \PartThreeVIP{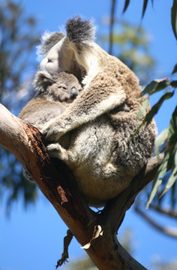}
      & \PartThreeVIP{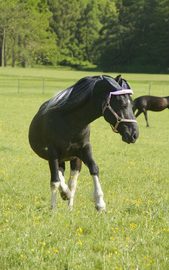}
      & \PartThreeVIP{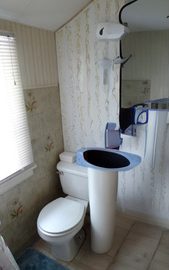}
      & \PartThreeVIP{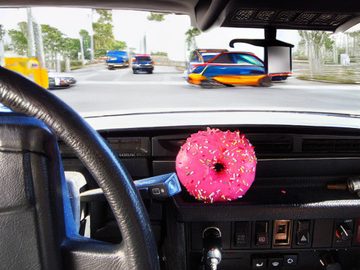}
      & \PartThreeVIP{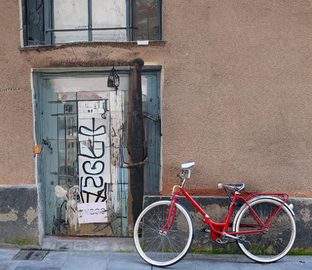} \\
    \PartThreeLabel{FLUX.1 Fill}
      & \PartThreeCell{000000303408}{flux1_fill}
      & \PartThreeOICell{9d5252360090f591}{flux1_fill}
      & \PartThreeCell{000000095988}{flux1_fill}
      & \PartThreeCell{000000151832}{flux1_fill}
      & \PartThreeCell{000000196002}{flux1_fill}
      & \PartThreeCell{000000203317}{flux1_fill} \\
    \PartThreeLabel{DING}
      & \PartThreeCell{000000303408}{ding}
      & \PartThreeOICell{9d5252360090f591}{ding}
      & \PartThreeCell{000000095988}{ding}
      & \PartThreeCell{000000151832}{ding}
      & \PartThreeCell{000000196002}{ding}
      & \PartThreeCell{000000203317}{ding} \\
    \PartThreeLabel{FlowChef}
      & \PartThreeCell{000000303408}{flowchef}
      & \PartThreeOICell{9d5252360090f591}{flowchef}
      & \PartThreeCell{000000095988}{flowchef}
      & \PartThreeCell{000000151832}{flowchef}
      & \PartThreeCell{000000196002}{flowchef}
      & \PartThreeCell{000000203317}{flowchef} \\
    \PartThreeLabel{fal/lora-outpaint}
      & \PartThreeCell{000000303408}{fal_flux2}
      & \PartThreeOICell{9d5252360090f591}{fal_flux2}
      & \PartThreeCell{000000095988}{fal_flux2}
      & \PartThreeCell{000000151832}{fal_flux2}
      & \PartThreeCell{000000196002}{fal_flux2}
      & \PartThreeCell{000000203317}{fal_flux2} \\
    \PartThreeLabel{FLUX.2 SFT}
      & \PartThreeCell{000000303408}{flux2_sft}
      & \PartThreeOICell{9d5252360090f591}{flux2_sft}
      & \PartThreeCell{000000095988}{flux2_sft}
      & \PartThreeCell{000000151832}{flux2_sft}
      & \PartThreeCell{000000196002}{flux2_sft}
      & \PartThreeCell{000000203317}{flux2_sft} \\
    \PartThreeLabel{Ours}
      & \PartThreeCell{000000303408}{ours}
      & \PartThreeOICell{9d5252360090f591}{ours}
      & \PartThreeCell{000000095988}{ours}
      & \PartThreeCell{000000151832}{ours}
      & \PartThreeCell{000000196002}{ours}
      & \PartThreeCell{000000203317}{ours}
  \end{tabular}
  \caption{\AN{Additional qualitative comparisons (Part 3). Columns denote
  samples and rows denote methods.}}
  \label{fig:additional_qualitative_3}
\end{figure*}

\begin{figure*}[p]
  \centering
  \scriptsize
  \setlength{\tabcolsep}{4pt}
  \renewcommand{\arraystretch}{1.02}
  \setlength{\extrarowheight}{0.5pt}
  \newcommand{\PartFourCell}[2]{%
    \includegraphics[height=0.091\textheight]{figures/qualitative/lvis/#1/#2.jpg}}
  \newcommand{\PartFourOICell}[2]{%
    \includegraphics[height=0.091\textheight]{figures/qualitative/oi/#1/#2.jpg}}
  \newcommand{\PartFourInput}[1]{%
    \includegraphics[height=0.091\textheight]{figures/qualitative/part4_cells/#1_input_gt.jpg}}
  \newcommand{\PartFourVIP}[1]{%
    \includegraphics[height=0.091\textheight]{figures/qualitative/part4_cells/#1_vip.jpg}}
  \newcommand{\PartFourLabel}[1]{%
    \raisebox{\dimexpr0.0455\textheight-0.5\height\relax}{%
      \rotatebox[origin=c]{90}{\fontsize{4.5}{5}\selectfont\textbf{#1}}}}
  \begin{tabular}{@{}c@{\hspace{2pt}}ccccc@{}}
    \PartFourLabel{Input / GT}
      & \PartFourInput{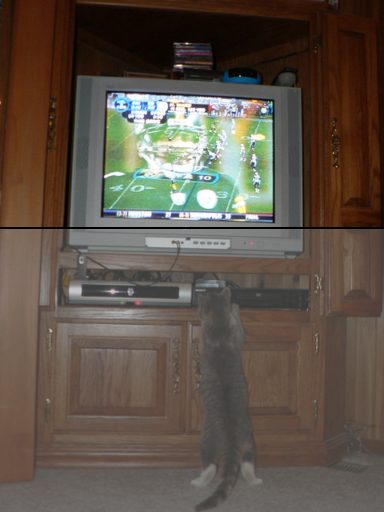}
      & \PartFourInput{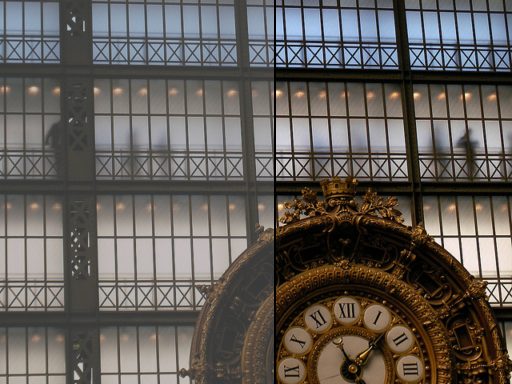}
      & \PartFourInput{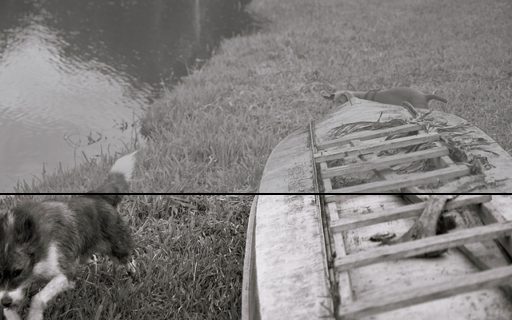}
      & \PartFourInput{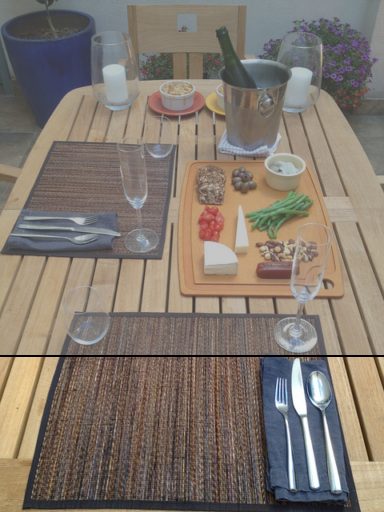}
      & \PartFourInput{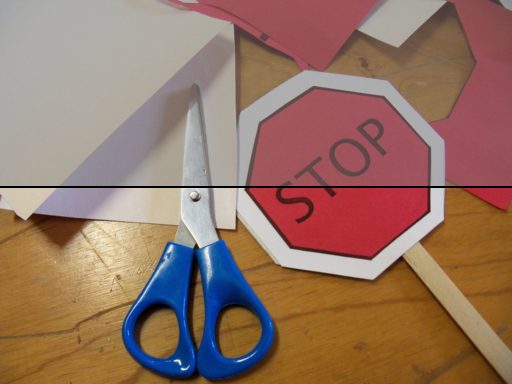} \\
    \PartFourLabel{BrushNet}
      & \PartFourCell{000000307649}{brushnet}
      & \PartFourCell{000000331317}{brushnet}
      & \PartFourCell{000000341363}{brushnet}
      & \PartFourCell{000000519558}{brushnet}
      & \PartFourOICell{0665cd6767d4c327}{brushnet} \\
    \PartFourLabel{PowerPaint}
      & \PartFourCell{000000307649}{powerpaint}
      & \PartFourCell{000000331317}{powerpaint}
      & \PartFourCell{000000341363}{powerpaint}
      & \PartFourCell{000000519558}{powerpaint}
      & \PartFourOICell{0665cd6767d4c327}{powerpaint} \\
    \PartFourLabel{VIP}
      & \PartFourVIP{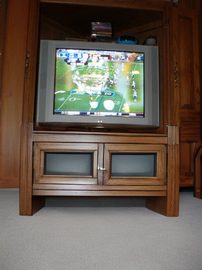}
      & \PartFourVIP{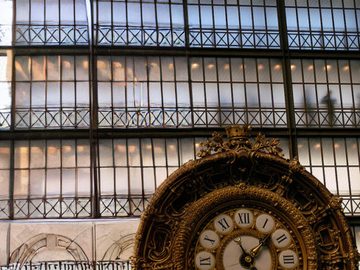}
      & \PartFourVIP{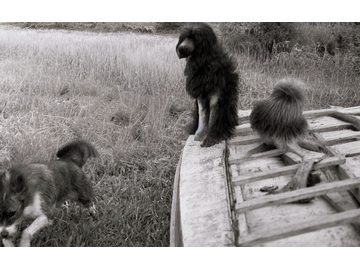}
      & \PartFourVIP{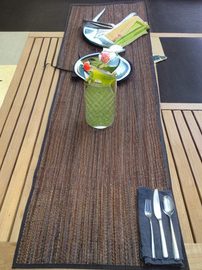}
      & \PartFourVIP{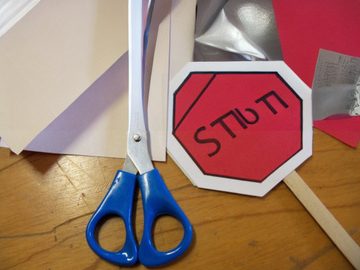} \\
    \PartFourLabel{FLUX.1 Fill}
      & \PartFourCell{000000307649}{flux1_fill}
      & \PartFourCell{000000331317}{flux1_fill}
      & \PartFourCell{000000341363}{flux1_fill}
      & \PartFourCell{000000519558}{flux1_fill}
      & \PartFourOICell{0665cd6767d4c327}{flux1_fill} \\
    \PartFourLabel{DING}
      & \PartFourCell{000000307649}{ding}
      & \PartFourCell{000000331317}{ding}
      & \PartFourCell{000000341363}{ding}
      & \PartFourCell{000000519558}{ding}
      & \PartFourOICell{0665cd6767d4c327}{ding} \\
    \PartFourLabel{FlowChef}
      & \PartFourCell{000000307649}{flowchef}
      & \PartFourCell{000000331317}{flowchef}
      & \PartFourCell{000000341363}{flowchef}
      & \PartFourCell{000000519558}{flowchef}
      & \PartFourOICell{0665cd6767d4c327}{flowchef} \\
    \PartFourLabel{fal/lora-outpaint}
      & \PartFourCell{000000307649}{fal_flux2}
      & \PartFourCell{000000331317}{fal_flux2}
      & \PartFourCell{000000341363}{fal_flux2}
      & \PartFourCell{000000519558}{fal_flux2}
      & \PartFourOICell{0665cd6767d4c327}{fal_flux2} \\
    \PartFourLabel{FLUX.2 SFT}
      & \PartFourCell{000000307649}{flux2_sft}
      & \PartFourCell{000000331317}{flux2_sft}
      & \PartFourCell{000000341363}{flux2_sft}
      & \PartFourCell{000000519558}{flux2_sft}
      & \PartFourOICell{0665cd6767d4c327}{flux2_sft} \\
    \PartFourLabel{Ours}
      & \PartFourCell{000000307649}{ours}
      & \PartFourCell{000000331317}{ours}
      & \PartFourCell{000000341363}{ours}
      & \PartFourCell{000000519558}{ours}
      & \PartFourOICell{0665cd6767d4c327}{ours}
  \end{tabular}
  \caption{\AN{Additional qualitative comparisons (Part 4). Columns denote
  samples and rows denote methods.}}
  \label{fig:additional_qualitative_4}
\end{figure*}

\begin{figure*}[p]
  \centering
  \scriptsize
  \setlength{\tabcolsep}{4pt}
  \renewcommand{\arraystretch}{1.02}
  \setlength{\extrarowheight}{0.5pt}
  \newcommand{\PartFiveCell}[2]{%
    \includegraphics[height=0.091\textheight]{figures/qualitative/oi/#1/#2.jpg}}
  \newcommand{\PartFiveInput}[1]{%
    \includegraphics[height=0.091\textheight]{figures/qualitative/part5_cells/#1_input_gt.jpg}}
  \newcommand{\PartFiveVIP}[1]{%
    \includegraphics[height=0.091\textheight]{figures/qualitative/part5_cells/#1_vip.jpg}}
  \newcommand{\PartFiveLabel}[1]{%
    \raisebox{\dimexpr0.0455\textheight-0.5\height\relax}{%
      \rotatebox[origin=c]{90}{\fontsize{4.5}{5}\selectfont\textbf{#1}}}}
  \begin{tabular}{@{}c@{\hspace{2pt}}ccccc@{}}
    \PartFiveLabel{Input / GT}
      & \PartFiveInput{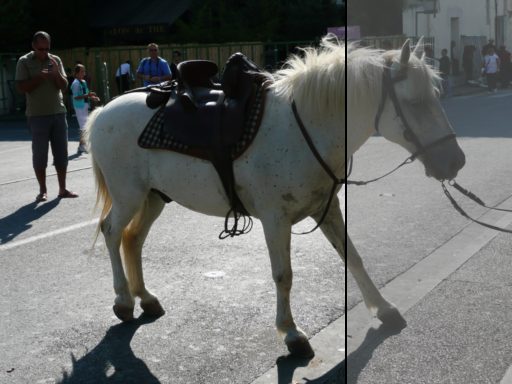}
      & \PartFiveInput{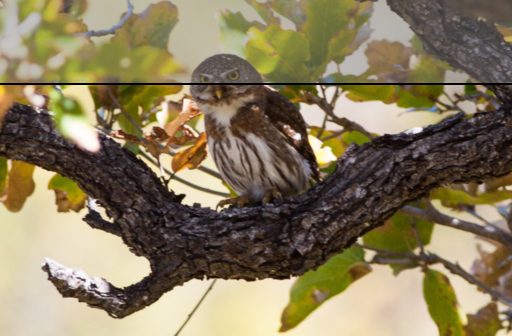}
      & \PartFiveInput{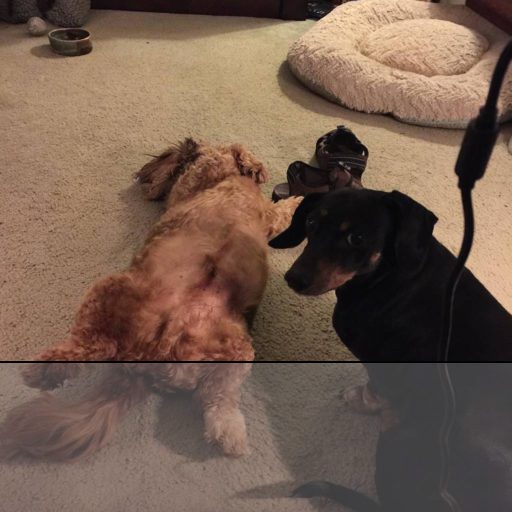}
      & \PartFiveInput{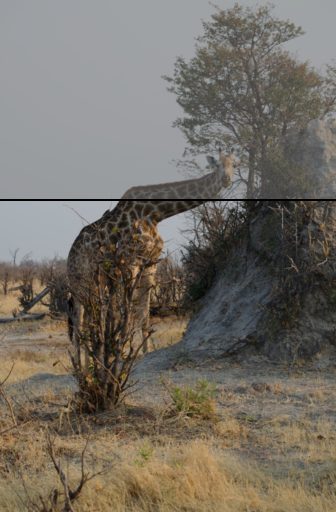}
      & \PartFiveInput{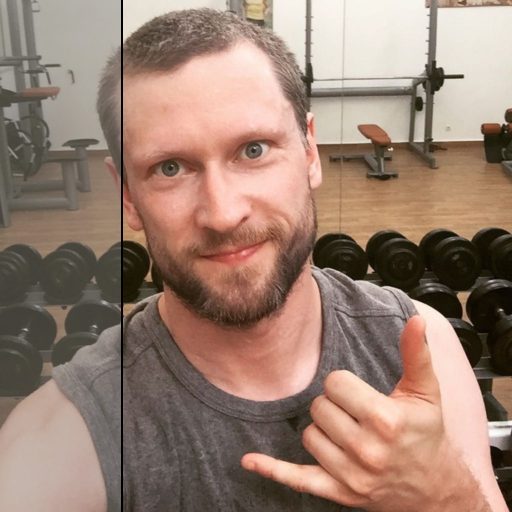} \\
    \PartFiveLabel{BrushNet}
      & \PartFiveCell{000a12b69b0de9f2}{brushnet}
      & \PartFiveCell{1170e81084bd523e}{brushnet}
      & \PartFiveCell{0c35a387e8a80f36}{brushnet}
      & \PartFiveCell{686e51475c13d00a}{brushnet}
      & \PartFiveCell{7258e9223b1753f9}{brushnet} \\
    \PartFiveLabel{PowerPaint}
      & \PartFiveCell{000a12b69b0de9f2}{powerpaint}
      & \PartFiveCell{1170e81084bd523e}{powerpaint}
      & \PartFiveCell{0c35a387e8a80f36}{powerpaint}
      & \PartFiveCell{686e51475c13d00a}{powerpaint}
      & \PartFiveCell{7258e9223b1753f9}{powerpaint} \\
    \PartFiveLabel{VIP}
      & \PartFiveVIP{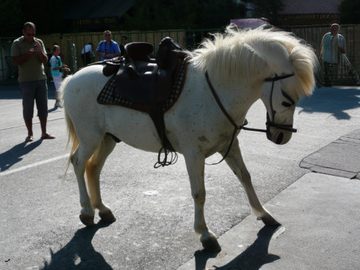}
      & \PartFiveVIP{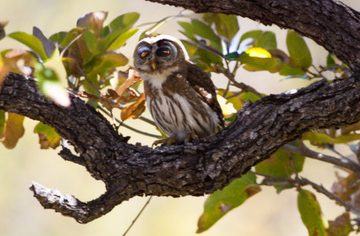}
      & \PartFiveVIP{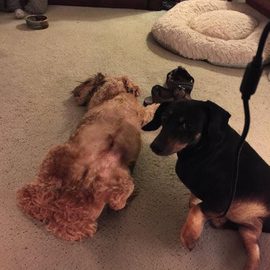}
      & \PartFiveVIP{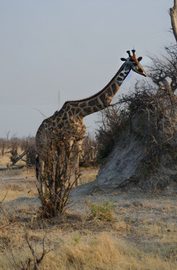}
      & \PartFiveVIP{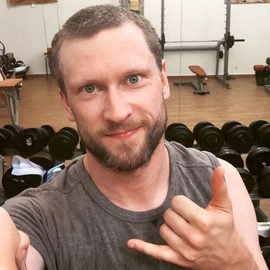} \\
    \PartFiveLabel{FLUX.1 Fill}
      & \PartFiveCell{000a12b69b0de9f2}{flux1_fill}
      & \PartFiveCell{1170e81084bd523e}{flux1_fill}
      & \PartFiveCell{0c35a387e8a80f36}{flux1_fill}
      & \PartFiveCell{686e51475c13d00a}{flux1_fill}
      & \PartFiveCell{7258e9223b1753f9}{flux1_fill} \\
    \PartFiveLabel{DING}
      & \PartFiveCell{000a12b69b0de9f2}{ding}
      & \PartFiveCell{1170e81084bd523e}{ding}
      & \PartFiveCell{0c35a387e8a80f36}{ding}
      & \PartFiveCell{686e51475c13d00a}{ding}
      & \PartFiveCell{7258e9223b1753f9}{ding} \\
    \PartFiveLabel{FlowChef}
      & \PartFiveCell{000a12b69b0de9f2}{flowchef}
      & \PartFiveCell{1170e81084bd523e}{flowchef}
      & \PartFiveCell{0c35a387e8a80f36}{flowchef}
      & \PartFiveCell{686e51475c13d00a}{flowchef}
      & \PartFiveCell{7258e9223b1753f9}{flowchef} \\
    \PartFiveLabel{fal/lora-outpaint}
      & \PartFiveCell{000a12b69b0de9f2}{fal_flux2}
      & \PartFiveCell{1170e81084bd523e}{fal_flux2}
      & \PartFiveCell{0c35a387e8a80f36}{fal_flux2}
      & \PartFiveCell{686e51475c13d00a}{fal_flux2}
      & \PartFiveCell{7258e9223b1753f9}{fal_flux2} \\
    \PartFiveLabel{FLUX.2 SFT}
      & \PartFiveCell{000a12b69b0de9f2}{flux2_sft}
      & \PartFiveCell{1170e81084bd523e}{flux2_sft}
      & \PartFiveCell{0c35a387e8a80f36}{flux2_sft}
      & \PartFiveCell{686e51475c13d00a}{flux2_sft}
      & \PartFiveCell{7258e9223b1753f9}{flux2_sft} \\
    \PartFiveLabel{Ours}
      & \PartFiveCell{000a12b69b0de9f2}{ours}
      & \PartFiveCell{1170e81084bd523e}{ours}
      & \PartFiveCell{0c35a387e8a80f36}{ours}
      & \PartFiveCell{686e51475c13d00a}{ours}
      & \PartFiveCell{7258e9223b1753f9}{ours}
  \end{tabular}
  \caption{\AN{Additional qualitative comparisons (Part 5). Columns denote
  samples and rows denote methods.}}
  \label{fig:additional_qualitative_5}
\end{figure*}

\begin{figure*}[p]
  \centering
  \begin{subfigure}[t]{0.82\textwidth}
    \makebox[\linewidth][c]{%
      \makebox[0.333\linewidth][c]{\textbf{Input / GT}}%
      \makebox[0.333\linewidth][c]{\textbf{GPT-Image-2}}%
      \makebox[0.333\linewidth][c]{\textbf{Ours}}}
  \end{subfigure}\par\smallskip
  \begin{subfigure}[t]{0.82\textwidth}
    \includegraphics[width=\linewidth]{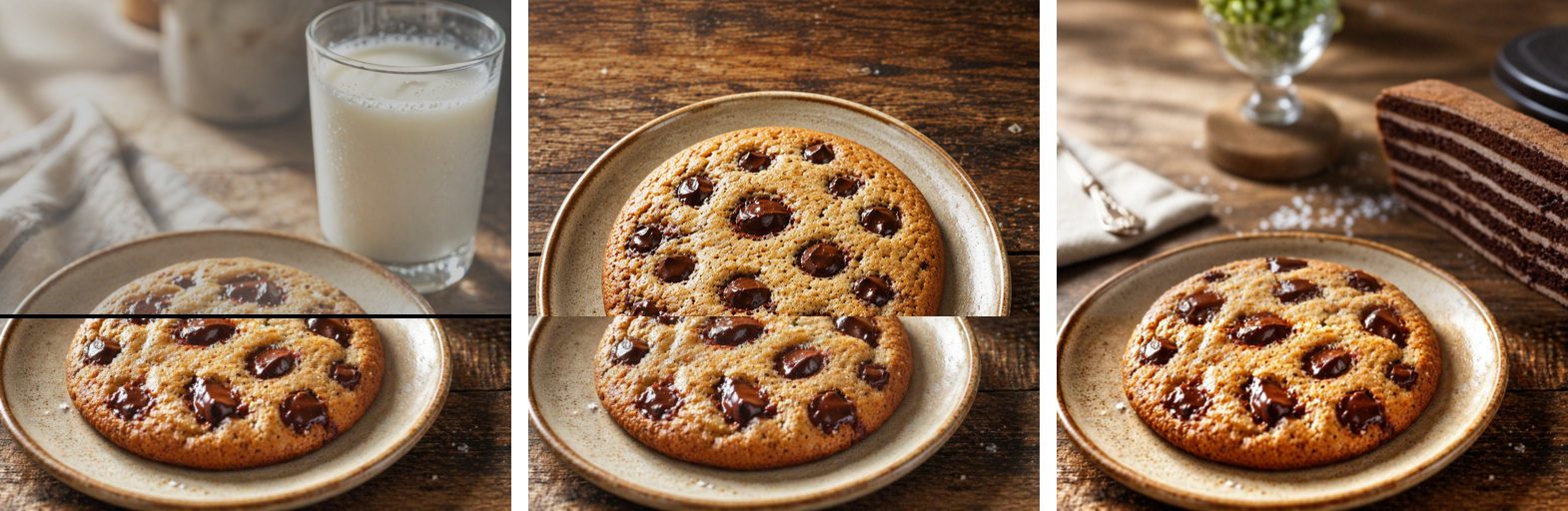}
  \end{subfigure}\par\smallskip
  \begin{subfigure}[t]{0.82\textwidth}
    \includegraphics[width=\linewidth]{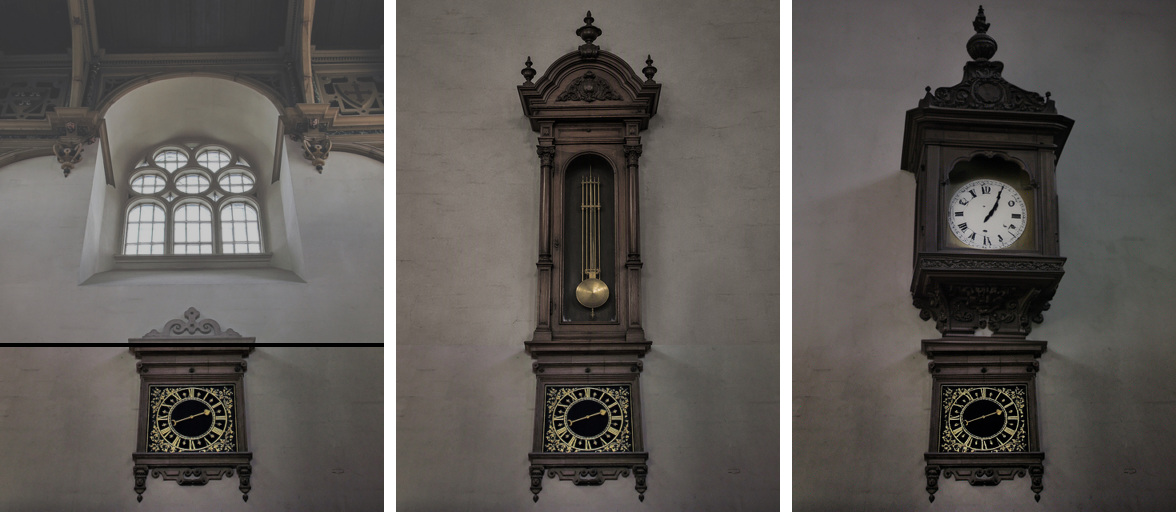}
  \end{subfigure}\par\smallskip
  \begin{subfigure}[t]{0.82\textwidth}
    \includegraphics[width=\linewidth]{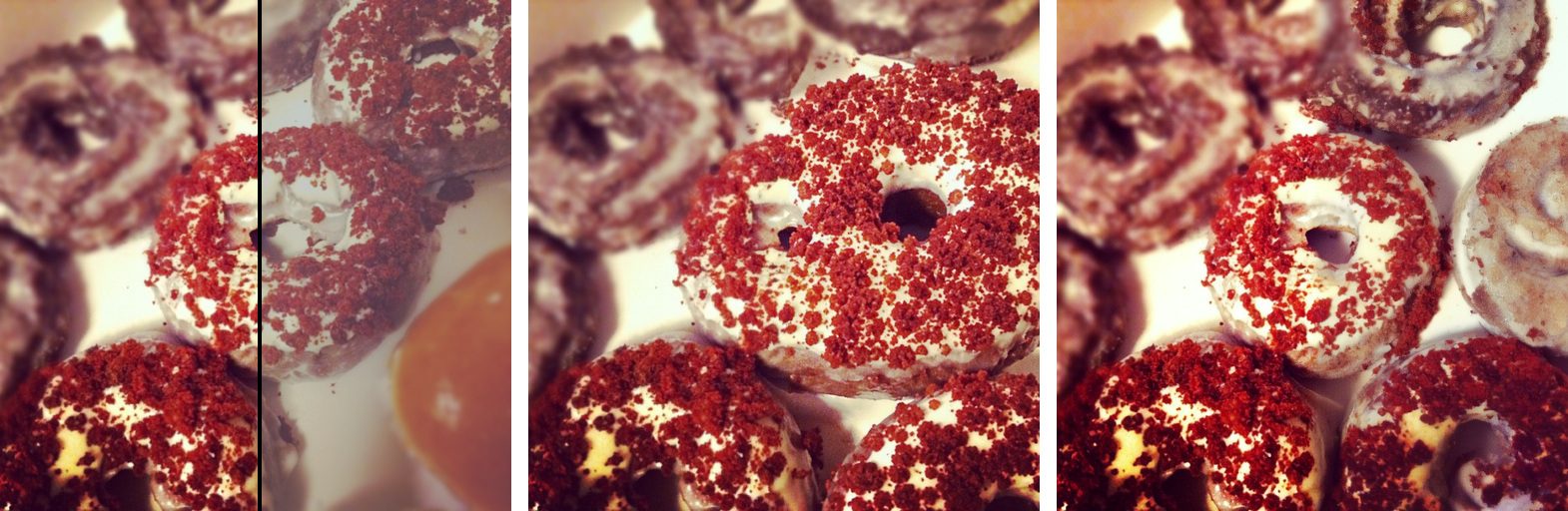}
  \end{subfigure}\par\smallskip
  \begin{subfigure}[t]{0.82\textwidth}
    \includegraphics[width=\linewidth]{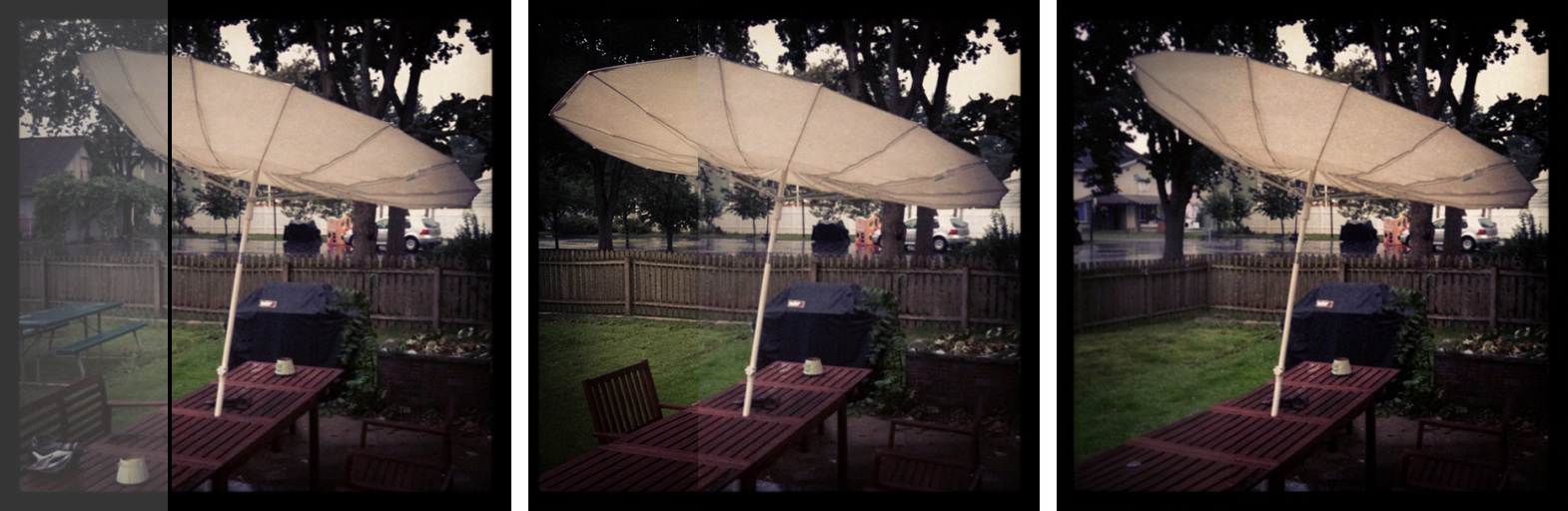}
  \end{subfigure}
  \caption{\AN{\textbf{Additional comparisons with GPT-Image-2 (Part 1).}
  Each row is a separate comparison panel; columns show the Input / GT visualization,
  GPT-Image-2, and our model.}}
  \label{fig:additional_gpt2_1}
\end{figure*}

\begin{figure*}[p]
  \centering
  \begin{subfigure}[t]{0.82\textwidth}
    \makebox[\linewidth][c]{%
      \makebox[0.333\linewidth][c]{\textbf{Input / GT}}%
      \makebox[0.333\linewidth][c]{\textbf{GPT-Image-2}}%
      \makebox[0.333\linewidth][c]{\textbf{Ours}}}
  \end{subfigure}\par\smallskip
  \begin{subfigure}[t]{0.82\textwidth}
    \includegraphics[width=\linewidth]{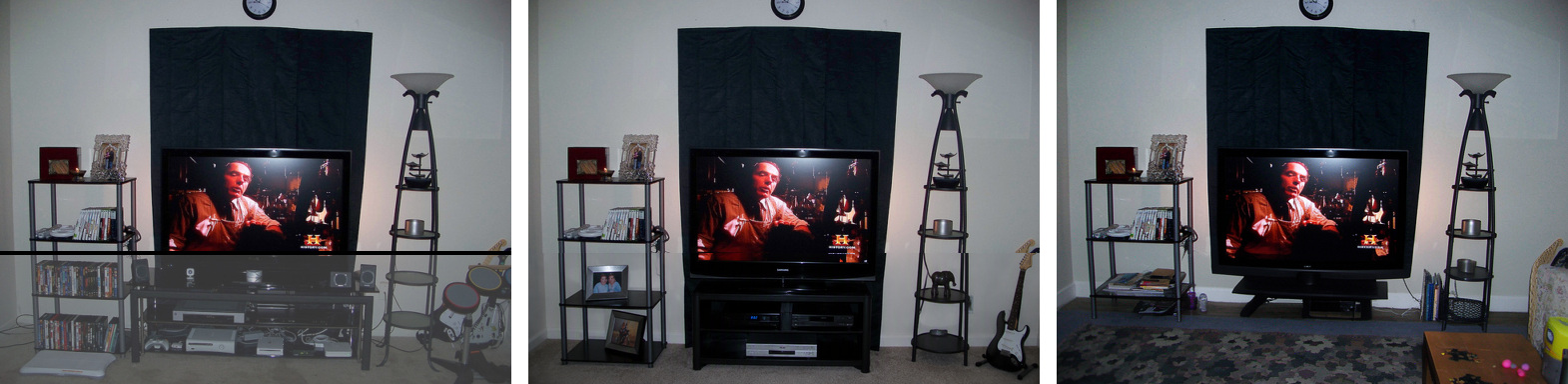}
  \end{subfigure}\par\smallskip
  \begin{subfigure}[t]{0.82\textwidth}
    \includegraphics[width=\linewidth]{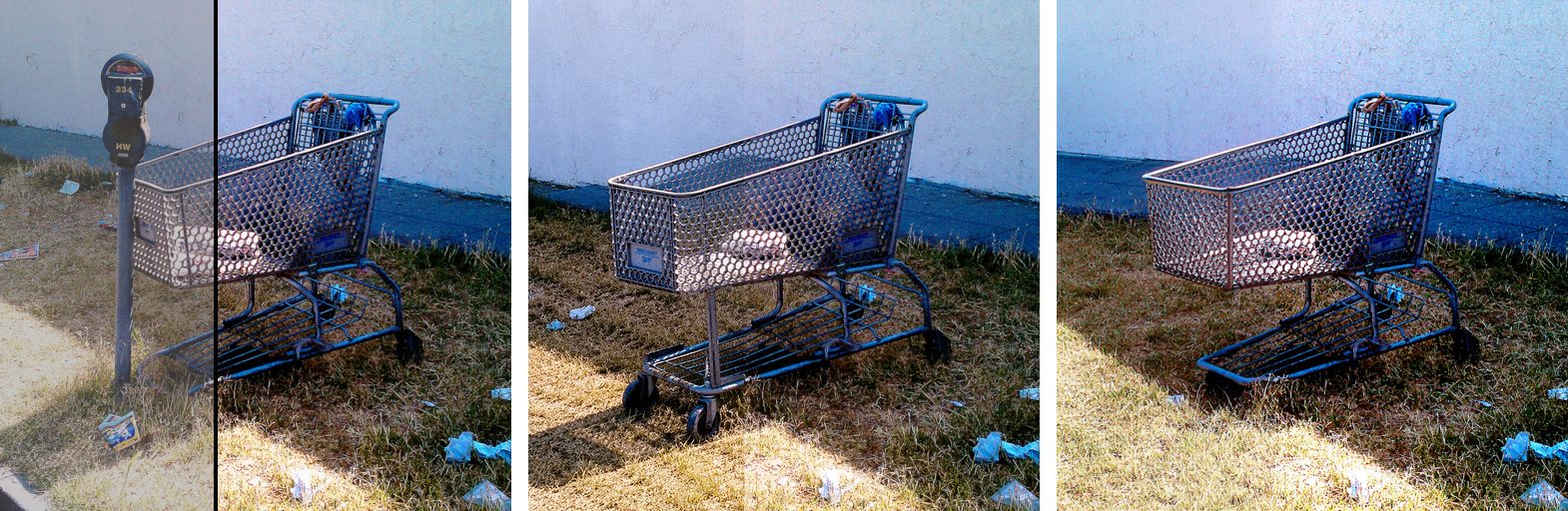}
  \end{subfigure}\par\smallskip
  \begin{subfigure}[t]{0.82\textwidth}
    \includegraphics[width=\linewidth]{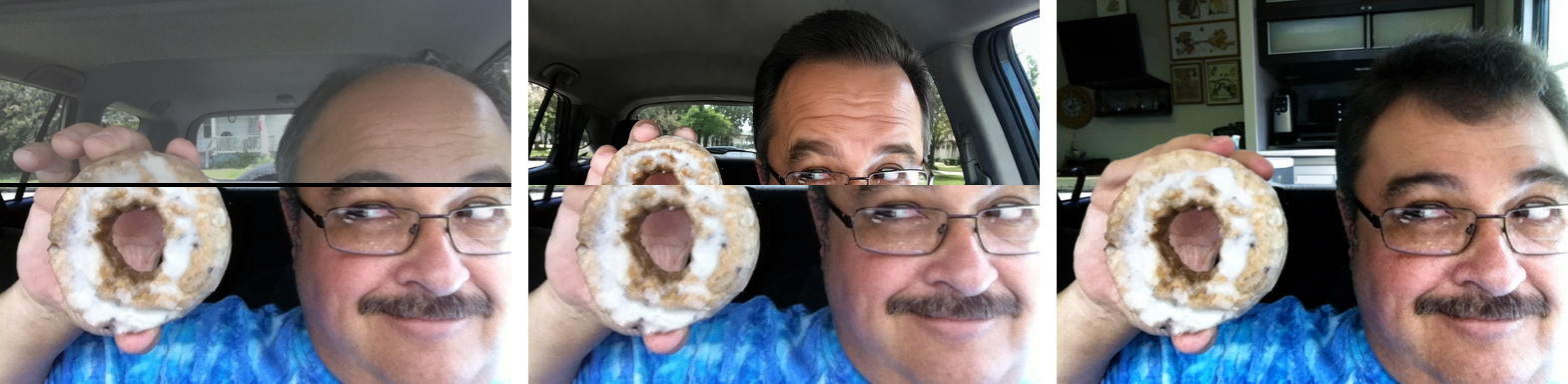}
  \end{subfigure}\par\smallskip
  \begin{subfigure}[t]{0.82\textwidth}
    \includegraphics[width=\linewidth]{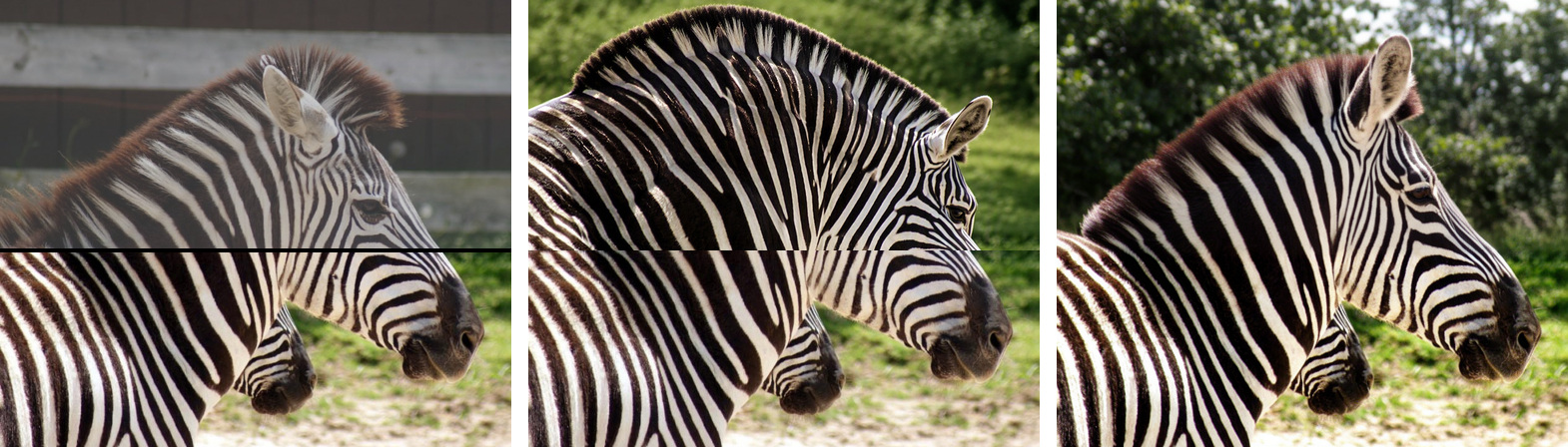}
  \end{subfigure}
  \caption{\AN{\textbf{Additional comparisons with GPT-Image-2 (Part 2).}
  Each row is a separate comparison panel; columns show the Input / GT visualization,
  GPT-Image-2, and our model.}}
  \label{fig:additional_gpt2_2}
\end{figure*}

\begin{figure*}[p]
  \centering
  \begin{subfigure}[t]{0.82\textwidth}
    \makebox[\linewidth][c]{%
      \makebox[0.333\linewidth][c]{\textbf{Input / GT}}%
      \makebox[0.333\linewidth][c]{\textbf{GPT-Image-2}}%
      \makebox[0.333\linewidth][c]{\textbf{Ours}}}
  \end{subfigure}\par\smallskip
  \begin{subfigure}[t]{0.82\textwidth}
    \includegraphics[width=\linewidth]{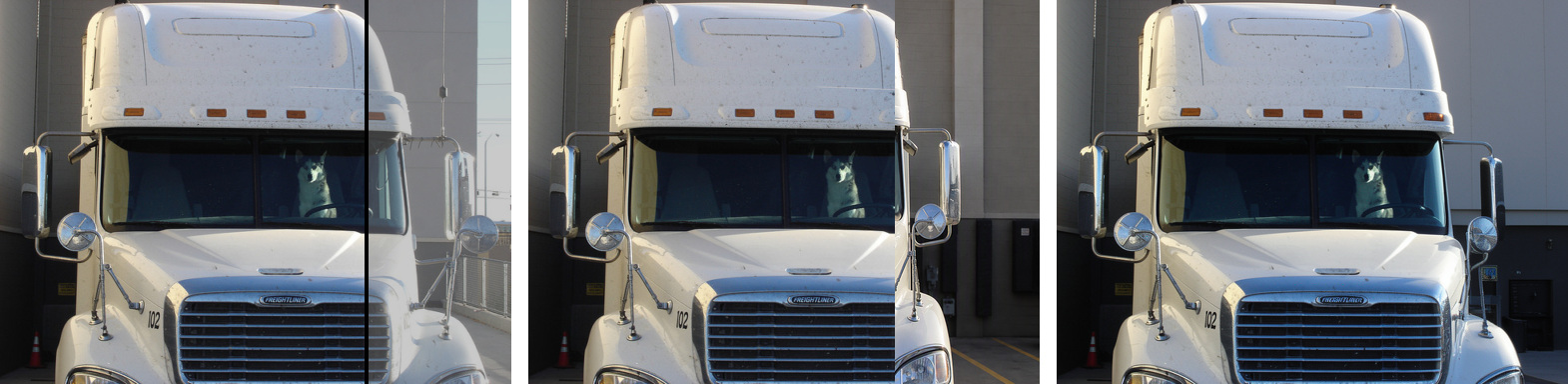}
  \end{subfigure}\par\smallskip
  \begin{subfigure}[t]{0.82\textwidth}
    \includegraphics[width=\linewidth]{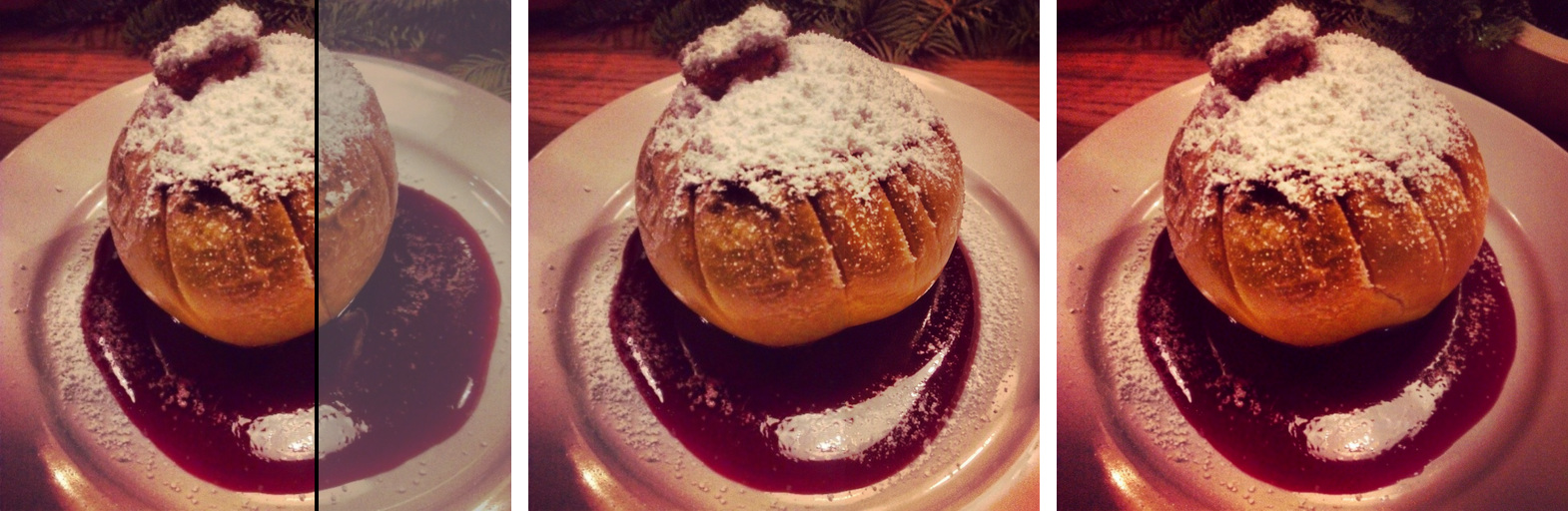}
  \end{subfigure}\par\smallskip
  \begin{subfigure}[t]{0.82\textwidth}
    \includegraphics[width=\linewidth]{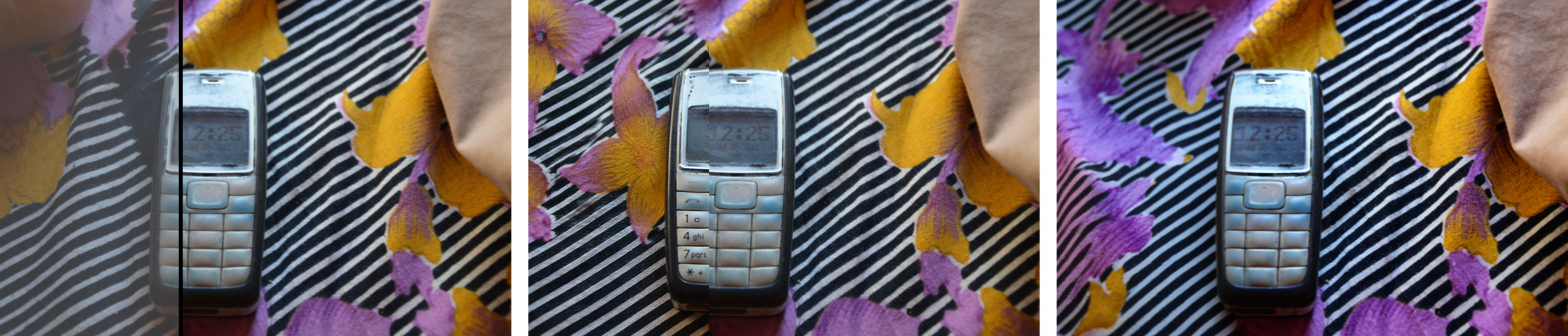}
  \end{subfigure}
  \caption{\AN{\textbf{Additional comparisons with GPT-Image-2 (Part 3).}
  Each row is a separate comparison panel; columns show the Input / GT visualization,
  GPT-Image-2, and our model.}}
  \label{fig:additional_gpt2_3}
\end{figure*}

\AN{\Cref{fig:additional_qualitative_1,fig:additional_qualitative_2,fig:additional_qualitative_3,fig:additional_qualitative_4,fig:additional_qualitative_5}
shows the complete baseline-comparison sample pool, including the examples
selected for the main paper. Samples are arranged as columns and methods as
rows, including all the baselines, and our model. The
Input / GT row uses the same visualization as the main paper: the ground-truth
outpaint region is gray-tinted and separated from the visible input by a black
boundary. \Cref{fig:additional_gpt2_1,fig:additional_gpt2_2,fig:additional_gpt2_3}
provides further comparisons with GPT-Image-2.}


\end{document}